\RequirePackage{booktabs}
\documentclass{article}

\usepackage{arxiv}

\usepackage{graphicx}
\usepackage{mathrsfs}%
\usepackage[title]{appendix}%
\usepackage{xcolor}%
\usepackage{textcomp}%
\usepackage{manyfoot}%
\usepackage{booktabs}%
\usepackage{listings}%
\usepackage{soul}
\usepackage{slashbox}

\usepackage{amsthm,amsmath,amssymb,amsfonts,mathtools,cuted}
\usepackage{algorithmic}
\usepackage{algorithm}
\usepackage{MnSymbol}
\usepackage{subcaption}

\DeclarePairedDelimiter{\nint}\lfloor\rceil

\DeclareMathOperator{\atantwo}{atan2}

\usepackage{flushend}
\usepackage{scalerel}
\usepackage{tabularx}
\usepackage{multirow}
\usepackage{color,soul}

\usepackage[hyperfootnotes=false]{hyperref} 
\hypersetup{
		colorlinks,
		urlcolor={blue!80!black}
	}

\title{  SKYSURF: A Self-learning Framework for Persistent Surveillance using Cooperative Aerial Gliders }

	\author{ \href{https://orcid.org/0000-0003-1547-1895}{\includegraphics[scale=0.06]{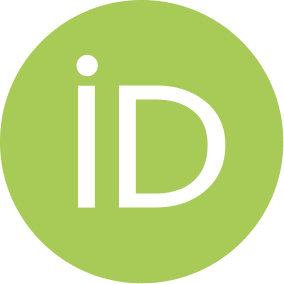}\hspace{1mm}Mohamadi Houssem Eddine} \\
	Department of Software and IT Engineering\\
	\'Ecole de Technologie Sup\'erieure\\
	Montreal, Quebec, Canada \\
	\texttt{houssem-eddine.mohamadi@etsmtl.ca} \\
	\And
	\href{https://orcid.org/0000-0002-0061-3283}{\includegraphics[scale=0.06]{orcid}\hspace{1mm}Nadjia Kara} \\
	Department of Software and IT Engineering\\
	\'Ecole de Technologie Sup\'erieure\\
	Montreal, Quebec, Canada \\
	\texttt{nadjia.kara@etsmtl.ca} \\
}

\renewcommand{\shorttitle}{}

\date{}

\begin{document}
	
\maketitle


	
	\begin{abstract}
		The success of surveillance applications involving small unmanned aerial vehicles (UAVs) depends on how long the limited on-board power would persist. To cope with this challenge, alternative renewable sources of lift are sought. One promising solution is to extract energy from rising masses of buoyant air. This paper proposes a local-global behavioral management and decision-making approach for the autonomous deployment of soaring-capable UAVs. The cooperative UAVs are modeled as non-deterministic finite state-based rational agents. In addition to a mission planning module for assigning tasks and issuing dynamic navigation waypoints for a new path planning scheme, in which the concepts of visibility and prediction are applied to avoid the collisions. Moreover, a delayed learning and tuning strategy is employed optimize the gains of the path tracking controller. Rigorous comparative analyses carried out with three benchmarking baselines and 15 evolutionary algorithms highlight the adequacy of the proposed approach for maintaining the surveillance persistency (staying aloft for longer periods without landing) and maximizing the detection of targets (two times better than non-cooperative and semi-cooperative approaches) with less power consumption (almost 6\% of battery consumed in six hours).
	\end{abstract}

	
	

\section{Introduction}

Unmanned Aerial Systems (UAS) are gaining momentum as they increasingly witnessing a raising uptake for missions such as surveillance, reconnaissance, data harvesting and communication relaying. With further advanced missions often shows up a greater necessity for performance requirements much like being capable of staying aloft or to fly farther for longer duration. A general prerequisite to fulfill such performance issues is to carry more fuel. However, this would call for bigger aircrafts and eventually making the deployment system more complicated to handle and more expensive \cite{andersson2009cooperating}.

One of the potential solutions for this challenge involves exploiting alternative sources of renewable energy from the environment (solar and wind power) throughout the course of the mission. The objective is to extract enough energy for metaphorically boosting the endurance of UAVs and help sustaining their flights as longer as possible.
A solar-driven solution requires mounting solar panel on UAVs and charging their batteries with the induced electricity. However, such a solution may not be cost-effective as the expenses of deployment and maintenance are pricey along with increasing the payload of UAVs.\\

Nevertheless, although not obvious, the UAVs can exploit solar-induced atmospheric energy, which takes many forms. Among others, rising masses of buoyant air a.k.a. thermals are formed when the air close to the earth surface is unevenly heated by solar radiation. It then becomes less dense compared with the surrounding air and little plumes of warm air ascend and line up to from a thermal \cite{tom2000meteorololgy,federal2015glider}.\\
It is of substantial importance to find methods for localizing and exploiting thermals. By mimicking the behaviors of soaring birds, UAVs being equipped with intelligent flight controllers, can integrate autonomous capabilities allowing them to cover large distances and stay aloft for longer periods.

During such flights, two major flight phases can be distinguished: searching and climbing. Whenever a thermal is abandoned or the UAV is in need of lift, a searching phase is carried out to timely localize a new lift zone before running out of altitude. And upon encountering a suitably strong thermal, the climbing phase is engaged by usually following circular paths around the core of the updraft to gain a maximum amount of energy. The UAV should harbor autonomous algorithms enabling it to appropriately balance between soaring flight behaviors and mission objectives.\\

In this work, we address the problem of prolonging the endurance of a flock formed by powered gliding UAVs. The aim is to ensure a persistent surveillance over a delimited region where a set of targets should be continuously monitored. The sought objective is to let the UAVs cooperate with each other and take advantage of thermal updrafts to stay aloft for longer periods without landing in order to keep monitoring the targets.\\
We propose a modular and scalable framework where the UAVs are modeled as rational agents with sufficient level of intelligence and autonomy, which reduce the intervention of Ground Control Station (GCS) operators since the UAVs are continuously making-decisions.
In the proposed self learning-based cooperative approach, the UAVs coordinate and manage their behaviors in such a way to maximize their own benefits (immediate goals) via learning how to interact with each other and with their environment while meeting all the mission requirements (future goals). By interacting with the environment, the UAVs will learn the pattern and conditions according to which updrafts and targets may appear or where they generally flock.
In addition to a dynamic path and mission planning module for safely navigating between waypoints and enabling the UAVs to smoothly switch between surveillance and exploration sub-missions.\\
The principal contributions of the paper involve the following points:
\begin{itemize}
	\itemsep0em
	\item A non-deterministic finite state and modular approach with distinct built-in behaviors and adaptive criteria for autonomously switching between them according to knowledge acquired through interactions with the dynamic environment. 
	\item A local and global management and decision-making mechanisms; the former (at UAV level) enables the UAVs to learn and make decisions for their benefit, and the latter (at flock level) manages and coordinates the behaviors of UAVs.
	\item An efficient mission planning method for assigning tasks and generating navigation waypoints in accordance with the UAV's ongoing sub-mission, which could be either surveillance or exploration.
	\item A new path planning scheme $H{{_{\scaleto{w}{4pt}}}}H$ is employed to generate collision-and-conflict-free trajectories using UAVs' perception (cameras and sensors) and their environmental awareness to discern present obstacles or predict future threats.
	\item A DLnT (Delayed learning and tuning) control strategy using the concepts of local searching, prediction and relation learning to reduce the tracking errors of 6-DOF fixed-wing UAV equipped with a PID controller.
\end{itemize}

The rest of the paper is structured as follows: Section \ref{sec2} summaries the recent and relevant related works classified into three subsections. In Section \ref{sec3}, the mathematical background of the problem with assumptions and objective functions are provided. Section \ref{sec4} for its part, it overviews the conceptual principle of the proposed solutions and describe its components. Section \ref{sec5} follows with concrete results and validating analyses. Section \ref{sec6} serves as a conclusion for the paper.

\section{Related works}
\label{sec2}

Multiple studies have investigated the theme of extracting energy from the atmosphere in various ways to boost and sustain the endurance of UAVs. The forthcoming subsections outline the relevant prior works that have addressed a similar research track. It is noteworthy to mention that only the works revolving around soaring-based surveillance and autonomous static soaring have been considered.

\subsection{Autonomous soaring algorithms}

The authors in \cite{liu2015uav} presented an atmospheric energy-harvesting scheme via an NMPC (Nonlinear Model Predictive Control) strategy. The objective was to determine the optimal trajectory of a UAV that maximizes its kinetic and potential energy over a finite receding horizon. Similarly, other researchers have adopted MPC-based soaring control schemes for 3-DOF UAVs in their works as a means to extract energy from updrafts \cite{liu2013nonlinear,pogorzelski2019autonomous}.\\
Using a nonlinear model with long prediction horizons often tend to worsen the behavior of thermal centering while linear models for prediction lead to mediocre results let alone the computational complexity when the number of UAVs grows.

Other researchers like \cite{hattenberger2024experimental,stolle2016vision} chose to depend on experimental vision-based approaches and real-world cues, namely cumulus-clouds to estimate the positions of thermals.
However, such an approach is imperfect since clouds may not always be formed where lift is present if there is not enough moisture in the rising air. Also, pursuing distant promising clouds may be deceptive because it is very likely that the lift has vanished by the time of arrival.\\

On the other hand, a detailed autonomous soaring algorithm is developed in \cite{depenbusch2018autosoar}, where a set of components are combined such as using a finite state machine modeling and heuristic switching criteria to manage the behaviors of the UAV and a dynamic mapping of environmental lift sources.\\
Unlike the self learning rules employed in our proposed approach, usually with heuristic task assignment like the one in \cite{depenbusch2018autosoar}, the eventual control policy tends to be myopic and to overlook potential global objectives. Moreover, the current heuristic rules used in decision-making are only suitable for missions where the goal is to remain aloft and are inadequate for managing a cooperative flock of UAVs.

Continuing with the context of mapping, an artificial lumbered flight algorithm that estimates updrafts and predicts multiple candidate paths for a UAV is developed in \cite{powers2020artificial}. Likewise, the authors in \cite{lawrance2011autonomous} proposed an approach for a model-free simulation of wind mapping with GPR (Gaussian process regression) using in-situ observations from an unpowered UAV.\\
The practicality of a global map with fast time varying atmosphere is questionable using one UAV. There may be no benefit from leaving a sufficiently strong lift to go visit a relatively distant zone anticipating to return if no better lift is found, since the previous thermal may have already dissipated.\\

Moreover, many AI-based strategies, namely reinforcement learning and its derivatives have been widely applied in the field. For instance, in \cite{notter2023deep} a UAV is trained by a deep RL approach to track and exploit thermals. A Twin Delayed Deep Deterministic policy gradient is employed for an unpowered glider to harvest energy from updrafts in \cite{zhao2023energy}. The authors in \cite{kim2021neuroevolutionary} utilized a NEAT (Neuro-Evolution of Augmenting Topologies) algorithm to train simulated soaring neuro-controllers. Their approach only applies penalties for unwanted behaviors and considers constant bank angles during thermaling. More similar works can be found in \cite{schimpf2021multi,notter2022integrated,flato2024revealing,wang2024learning}.

A prevailing issue with AI-based methods concerns the training process of UAVs. In contrast to supervised learning, agents in RL solutions collect data through interactions with the environment. This dependence can give rise to a no-win endless loop if the agent keeps collecting poor quality data and so several runs are required to get quantitative results. Another issue is about the choice of RL algorithms and reward functions as they greatly influence the outcomes. In our off-policy model-free approach, rational agents learn as they fly and interact with their environment without any prior training before the mission.

\subsection{Persistent surveillance}

As for deploying soaring-capable UAVs to maintain a continuous surveillance, the authors in \cite{makovkin2014optimal} presented a coordinated soaring strategy involving small UAVs reiterating round-trip flights between an updraft of known location and a monitoring static target. The objective was to minimize the number of UAVs required to ensure a persistent surveillance by optimizing the cruise speeds.\\
An EKF (Extended Kalman Filter) prediction based on ordinary least squares (OLS) is proposed in \cite{an2023autonomous}. Here, the UAVs are operated in two ways; either following a lawnmower-based path to systemically cover a mission area or continuously tracking the predicted thermal centers. An akin search strategy is presented in \cite{rosales2024evaluation}, where the authors proposed a simulated tree architecture-based soaring algorithm for a powered UAV to navigate between thermal updrafts.\\
A single gliding UAV is programmed to perform incremental surveillance by visiting concentric expanding regions \cite{gao2015autonomous}, where a GPR-based algorithm is utilized for estimating the updrafts and a Dubins-based path planning is used for smoothly transiting between tasks.

\subsection{Cooperative autonomous soaring}

Fewer related works investigated the advantages of deploying multiple collaborative UAVs.\\
In \cite{andersson2021improving}, the authors proposed an algorithm for deploying two autonomous gliders and enabling them to cooperate to augment the probability of finding thermal lifts.\\
A distributed mapping of convective activities is adopted in \cite{depenbusch2011coordinated}. The map was coupled with a behavior-based controller to facilitate the coordination within a flock of UAVs and maximize their endurance. An analogous approach was employed in \cite{cheng2014guided} albeit a guidance exploration priority due to wind speed uncertainties was integrated into the discretized maps. The authors in \cite{cobano2013multiple} proposed a BRHS (Bounded Recursive Heuristic Search) algorithm to prolong the flight duration of cooperative explorations with multiple gliding UAVs.\\
Nevertheless, the cooperation between UAVs in these works is limited to only sharing a map of where lift sources have been encountered or exploited without discussing how to manage the behaviors of the flock, excluding the work of \cite{depenbusch2011coordinated}, which also considered coordinated task assignment to some extent in contrast to our approach as will be discussed in section \ref{sec4}.

\section{Problem formulation}
\label{sec3}

Given a geographical region-of-interest (RoI) where meteorological conditions (e.g., moisture and solar radiation) are favorable for forming rising masses of air, the objective is to deploy a flock of powered gliding UAVs. These gliders (denoted hereinafter as agents) are capable of exploiting updrafts to ensure a prolonged and continuous surveillance over a set of targets randomly scattered across the region.
Before modeling the system and defining the objective functions, the first step consists in outlining the problem's assumptions and which are the following:\\

\textbf{1)} The UAVs are modeled as rational agents with acceptable level of intelligence and autonomy. \textbf{2)} The UAVs are homogeneous in terms of physical features and computational capabilities. \textbf{3)} The fields-of-view of UAVs' cameras are modeled as circles of variable radii during flight. \textbf{4)} Updrafts can be exploited by more than one UAV if safety measures are met. \textbf{5)} Atmospheric conditions last indefinitely. \textbf{6)} Losses in the U2X communication links are neglected. \textbf{7)} Targets are modeled as mass points that can be detected by UAVs passing over them. \textbf{8)} Each updraft/target is unknown in advance and it appears at a random time/location then it persists for some duration and once it fades out, a new one will randomly pop up later. \textbf{9)} Some UAV-related flight phases and metrics such as take-off /landing and avionic power consumption are ignored.\\

\begin{figure}
	\centering
	\includegraphics[scale=0.2]{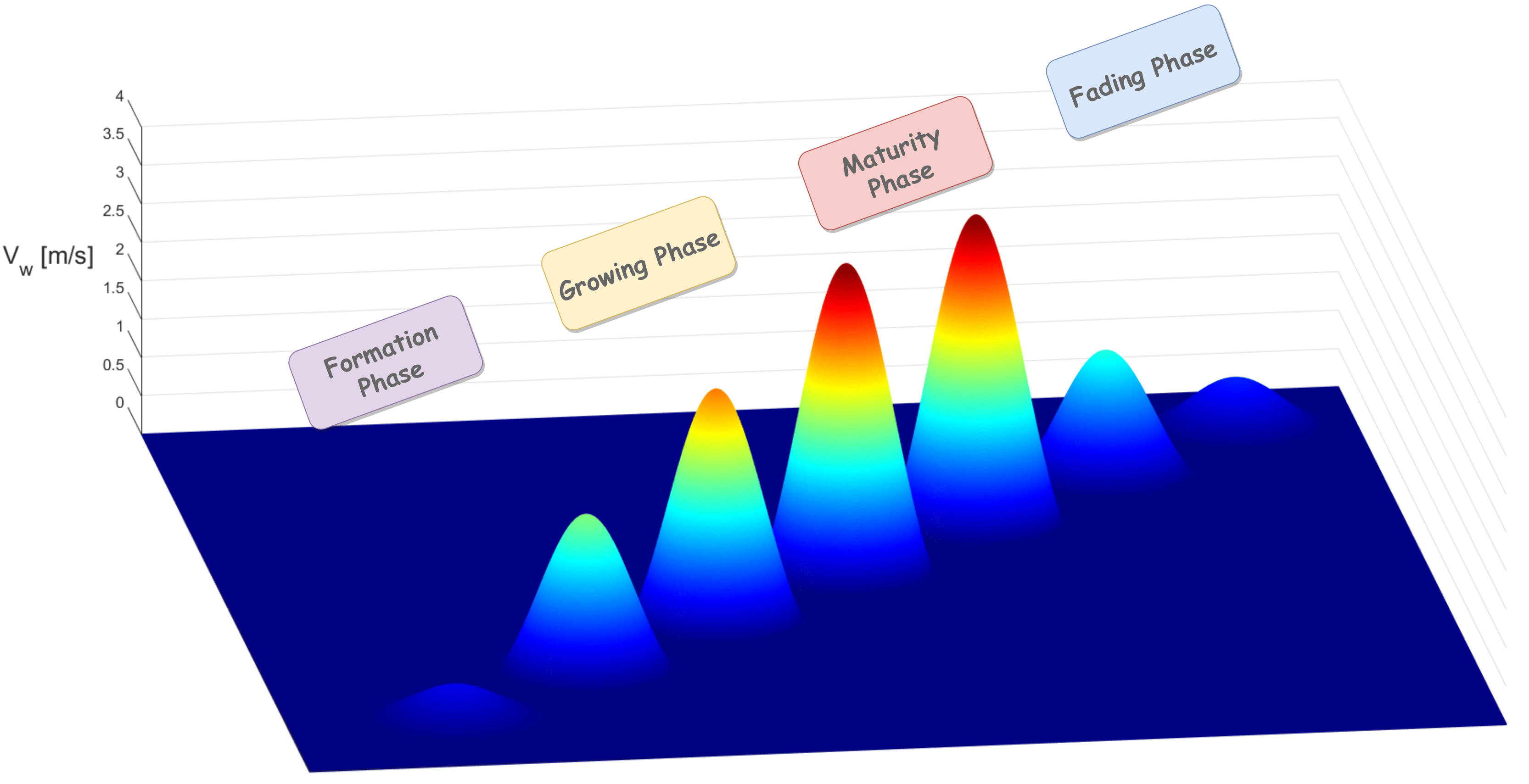}
	\caption{ Velocity profile of a sample updraft in different development phases }
	\label{fig_updraft}
\end{figure}

The deployed agents are assumed to be subject to ascending air masses, whose vertical velocity $V_w$ adheres to the Gedeon updraft model as given by Eq. \ref{eq_updraft} with respect to each coordinate $(x,y)$ in the Region-of-Interest.

\begin{equation}
		V_w(x,y,t) = \tau(t)V_{w_0}\exp^{-\left(\frac{\left(x-\left(x_w\right)\right)^2-\left(y-\left(y_w\right)\right)^2}{r_w^2}\right)} \times  \left(1-\left(\frac{\left(x-\left(x_w\right)\right)^2-\left(y-\left(y_w\right)\right)^2}{r_w^2}\right)\right)
	\label{eq_updraft}
\end{equation}

where $r_w,V_{w_0},x_w,y_w$ are respectively the updraft's radius, vertical wind speed at the center and the current center coordinates. $\tau(t) \in [0, 1]$ is a function inspired from \cite{lecarpentier2017empirical} of trapezoidal form reflecting the development of an updraft with time. Four phases are considered: formation, growing, maturity and fade-off, which represent accordingly 1\%, 2\%, 5\% and 2\% of the updraft's lifecycle. An example is shown in Fig. \ref{fig_updraft}.

It should be noted that the maximum number of updrafts that can be found in the region at an instant $t$ depends on the average radii and height of updrafts that could be formed in the convective boundary layer.

Now after having set up the simulation environment, an appropriate fixed-wing UAV model should be developed in order to validate the adopted guidance and control strategy and namely during the powered-climb flight phases.

Referring to several textbooks on aeronautics and flight dynamics (e.g., \cite{beard2012small,gao2015autonomous}), the 6-DOF model for the fixed-wing UAV dynamics and kinematics are given by the nonlinear equations \ref{eq_uav_model}(a-i) and are mathematically expressed with respect to Newton's laws as follows:

\begin{subequations}
	\footnotesize
	\begin{align}
		&  \dot{p} = \Gamma_1pq - \Gamma_2qr + \frac{1}{2}\rho {V_a}^2Sb \left(\frac{J_zC_l+J_{xz}C_n}{\Gamma} \right) \label{eq_p_dot} \\
		&  \dot{q} = \frac{J_z-J_x}{J_y}pr - \frac{J_{xz}}{J_y}(p^2-r^2) + \frac{1}{2J_y}\rho {V_a}^2S\bar{c}C_m  \label{eq_q_dot} \\
		&	\dot{r} = \Gamma_3pq -\Gamma_1qr + \frac{1}{2}\rho {V_a}^2Sb \left(\frac{J_{xz}C_l + J_xC_n}{\Gamma} \right)  \label{eq_r_dot}\\
		& \dot{V_a} = \frac{1}{2m}\rho {V_a}^2S\left(C_{{_{\scaleto{Y}{4pt}}}}\sin\beta - C_{{_{\scaleto{D}{4pt}}}} \right) - g\sin\gamma  \\
		& \dot{\gamma} = \frac{1}{2m}\rho {V_a}S\left( C_{{_{\scaleto{L}{4pt}}}}\cos\mu - C_{{_{\scaleto{Y}{4pt}}}}\sin\mu\cos\beta \right) - \frac{g\cos\gamma}{V_a}  \\
		& \dot{\psi} = \frac{1}{2m\cos\gamma}\rho {V_a}S\left( C_{{_{\scaleto{L}{4pt}}}}\sin\mu + C_{{_{\scaleto{Y}{4pt}}}}\cos\mu\cos\beta \right)  \\
		& \dot{\alpha} = q -\frac{\rho {V_a}SC_{{_{\scaleto{L}{4pt}}}}}{2m\cos\beta} + \frac{g\cos\gamma\cos\mu}{V_a\cos\beta} - \tan\beta\left(p\cos\alpha + r\sin\alpha\right)  \\
		& \dot{\beta} = \frac{\rho {V_a}SC_{{_{\scaleto{Y}{4pt}}}}\cos\beta}{2mV_a} + \frac{g\cos\gamma\sin\mu}{V_a} + p\sin\alpha - r\sin\alpha   \\
		\begin{split}
			&	\dot{\mu} = \frac{p\cos\alpha + r\sin\alpha}{\cos\beta} + \frac{\rho {V_a}S}{2mV_a}   \Big[C_{{_{\scaleto{L}{4pt}}}}\left(\tan\gamma\sin\mu + \tan\beta\right)   \nonumber
		\end{split}  \\
		& {\quad}  + C_{{_{\scaleto{Y}{4pt}}}}\left(\tan\gamma\sin\mu\cos\beta\right) \Big] - \frac{g\cos\gamma\cos\mu\tan\beta}{V_a}
	\end{align}
	\label{eq_uav_model}
\end{subequations}

where $p, q, r$ are respectively the roll, pitch and yaw rates, $J_x, J_y, J_z$ are the moments of inertia, $J_{xz}$ is the product of inertia, $ \Gamma = J_xJ_z-{J_{xz}}^2$, $\Gamma_1 = \frac{J_{xz}(J_x-J_y+J_z)}{\Gamma}$, $\Gamma_2 = \frac{J_z(J_z-J_y)+{J_{xz}}^2}{\Gamma}$ and $\Gamma_3 = \frac{J_{x}(J_x-J_y)+{J_{xz}}^2}{\Gamma}$.
$V_a,\gamma,\psi,\alpha,\beta,\mu$ are the UAV's airspeed, flight path angle, heading angle, angle of attack, side-slip attack and bank angle.\\
$S, b, \bar{c}, m, g, \rho$ are the UAV's wing area, wing span, chord length, mass, gravity acceleration and atmospheric density.\\
Moreover, roll, pitch and yaw moments are given according to \cite{beard2012small} by Eqs. \ref{eq_uav_coeff}(a-c), while side force, drag and lift coefficients are expressed in the order shown in Eqs. \ref{eq_uav_coeff}(d-f).

\begin{subequations}
	\small
	\begin{align}
		& C_l =  C_{l_\beta}\beta + \frac{1}{2V}\left(C_{l_p}pb + C_{l_r}rb\right) + C_{l_{\delta_r}}\delta_r + C_{l_{\delta_a}}\delta_a  \label{eq_cl}\\
		& C_m =  C_{m_0} + C_{m_{\alpha}}\alpha + \frac{1}{2V}C_{m_q}q\bar{c} + C_{m_{\delta_e}}\delta_e \label{eq_cm}\\
		& C_n = C_{n_\beta}\beta + \frac{1}{2V}\left(C_{n_p}pb + C_{n_r}rb\right) + C_{n_{\delta_r}}\delta_r + C_{n_{\delta_a}}\delta_a  \label{eq_cn} \\
		& C_{{_{\scaleto{Y}{4pt}}}} = C_{{_{\scaleto{Y_\beta}{5pt}}}} + C_{{_{\scaleto{Y_{\delta_r}}{5pt}}}}\delta_r \label{eq_cY} \\
		& C_{{_{\scaleto{D}{4pt}}}} = C_{{_{\scaleto{D_0}{5pt}}}} + \frac{ C_{{_{\scaleto{L}{4pt}}}}^2 }{ \pi e\frac{b^2}{S}} \label{eq_cD} \\
		& C_{{_{\scaleto{L}{4pt}}}} = C_{{_{\scaleto{L_0}{4pt}}}} + C_{{_{\scaleto{L_\alpha}{4pt}}}}\alpha + C_{{_{\scaleto{L_{\delta_e}}{5pt}}}}\delta_e \label{eq_cL}
	\end{align}
	\label{eq_uav_coeff}
\end{subequations}

By virtue of the state-dependent Riccati equation approach (SDRE) \cite{cimen2012survey}, The actuator inputs $U = {\left[\delta_e, \delta_a, \delta_r\right]}^T$, which are the elevator, aileron and rudder deflections can be formulated by rewriting the rotational dynamics (Eqs. \ref{eq_p_dot}, \ref{eq_q_dot}, \ref{eq_r_dot}) in the affine-in-control form, bearing in mind that $X = {\left[p, q, r\right]}^T$ as indicated below:
\begin{equation}
	\small
	U = {G(X)}^{-1}\left( \dot{X} - F(X) \right)
\end{equation}

where
\begin{subequations}
	\small
	\begin{align}
		&  \begin{aligned}
			F(X) &=
			\left[\begin{matrix}
				\Gamma_1pq - \Gamma_2qr + \frac{\rho V_a Sb }{4}\big[ \frac{J_z}{\Gamma}\left( 2V_aC_{l_\beta}\beta + b( C_{l_p}p + C_{l_r}r) \right)  \\
				\frac{\rho V_a S{\bar{c}}^2 }{4J_y}\left[ C_{m_q}q + 2\left( C_{m_0} + C_{m_\alpha}\alpha \right)\right] \\
				\Gamma_3pq -\Gamma_1qr + \frac{\rho V_a Sb }{4}\big[ \frac{J_{xz}}{\Gamma}\left( 2V_aC_{l_\beta}\beta + b(C_{l_p}p + C_{l_r}r) \right) 
			\end{matrix}\right. \\
			&\qquad\qquad
			\left.\begin{matrix}
				+ \frac{J_{xz}}{\Gamma}\left( 2V_aC_{n_\beta}\beta +  b(C_{n_p}p + C_{n_r}r) \right) \big] \\
				+\frac{J_z-J_x}{J_y}pr + \frac{J_{xz}}{J_y}(r^2 - p^2) \\
				+ \frac{J_{x}}{\Gamma}\left( 2V_aC_{n_\beta}\beta +  b(C_{n_p}p + C_{n_r}r) \right) \big]
			\end{matrix}\right]
		\end{aligned} \\
		& G(X) =  \frac{\rho V_a^2 S}{2} \begin{bmatrix}
			0  &  \frac{bJ_zC_{l_{\delta_r}} + bJ_{xz}C_{n_{\delta_r}}}{\Gamma}  &  \frac{bJ_zC_{l_{\delta_a}} + bJ_{xz}C_{n_{\delta_a}}}{\Gamma} \\
			\frac{\bar{c}C_{m_{\delta_e}}}{J_y}  &  0  &  0\\
			0  &  \frac{bJ_{xz}C_{l_{\delta_r}} + bJ_{x}C_{n_{\delta_r}}}{\Gamma}  &  \frac{bJ_{xz}C_{l_{\delta_a}} + bJ_{x}C_{n_{\delta_a}}}{\Gamma}
		\end{bmatrix}
	\end{align}
	\label{eq_crowding}
\end{subequations}

Recapping once more the addressed problem in which a flock of $N_u$ gliding UAVs are to be deployed for a mission lasting $T_\mathcal{M}$ hours. The objective is to stay aloft as longer as possible by exploiting updrafts and to monitor a set of $N_t(t)$ targets scattered across a region-of-interest. Each target requires to be monitored for a $\tau_t$ period to be labeled as decidedly observed. \\
In this context, the objective function \ref{eq_F1} aims at maximizing the number of recorded targets that have been appropriately detected at each instant $t$. The functions \ref{eq_F2} and \ref{eq_F3} are meant for optimizing the flight duration and potential energy extraction with less battery consumption. Implicitly, this should be enough to force the UAVs to seek and maximize the exploitation of updrafts.

\begin{subequations}
	\footnotesize
	\begin{align}
		&	F_1 = \quad \sum_{t = 1}^{T_s} \frac{1}{\tau_tN_t(t)}\sum_{i = 1}^{N_t(t)} \sum_{j = 1}^{N_u}p_{tij}d_{tij}\max(t_{m_{i}}, \tau_t)  \label{eq_F1}  \\
		&   F_2 = \quad \frac{1}{N_u}\sum_{i = 1}^{N_u} \sum_{t = 1}^{T_s} \frac{1}{3600}k_{ti}  \label{eq_F2} \\
		&   F_3 = \quad \frac{1}{N_u} \sum_{i = 1}^{N_u} \sum_{t = 1}^{T_s} c_{ti} \left( (1 - q_{ti} )\frac{mgz_{ti}}{3600} - q_{ti}\lambda b_{ti} \right) \label{eq_F3}
	\end{align}
\end{subequations}

Here, the binary variables are expressed as follows:

\begin{subequations}
	\footnotesize
	\begin{align}
		&   p_{tij} = \begin{cases}
			1  &  \text{if} \quad (\mathcal{ID}_{ti} \notin \mathcal{L}_t) \; \lor \; (\mathcal{ID}_{ti} \in \mathcal{L}_t \; \land \; t_{m_{ti}} < \tau_t )\\
			0  & \text{otherwise}
		\end{cases}  \\
		&	d_{tij} = \begin{cases}
			1  &  \text{if} \quad DIST_{ij} \leq z_{tj}\tan(\frac{\vartheta}{2}) \\
			0  & \text{otherwise}
		\end{cases}  \\
		&   k_{ti} = \begin{cases}
			1  &  \text{if} \quad \frac{B_{c_0} - r_{ti}}{B_{c_0}} \geq \mathcal{B}_d \\
			0  & \text{otherwise}
		\end{cases}  \\
		&  	c_{ti} = \begin{cases}
			1  &  \text{if} \quad z_{ti} > z_{\widehat{t}i}; \quad \widehat{t} = t-1 \\
			0  & \text{otherwise}
		\end{cases}  \\
		&  	q_{ti} = \begin{cases}
			1  &  \text{if} \quad b_{ti} < b_{\widehat{t}i}; \quad \widehat{t} = t-1  \\
			0  & \text{otherwise}
		\end{cases} 
	\end{align}
\end{subequations}

where $T_s = 3600\times T_{\mathcal{M}}$ and $\mathcal{ID}_{ti}, \mathcal{L}_t, t_{m_{ti}},\vartheta$ are respectively the ID of a target, the list of previously detected targets, the accumulated monitoring time over a target and the UAV's angle of view. $B_{c_0},\mathcal{B}_d, r_{ti}, r_{ti}$ are the UAV's initial and desired battery capacities as well as the remaining battery capacity and power consumption at the instant $t$. $\lor,\land$ are the logical or/and operators and $\lambda$ is a penalty coefficient.\\
These objective functions come with the following set of constraints that should be congruently satisfied:

\begin{subequations}
	\footnotesize
	\begin{align}
		&	\forall\; t \in [1, T_s]; \; i \in [1, N_u]; \; (x_{ti}, y_{ti}, z_{ti}) \in {\mathcal{S}_{\mathcal{P}}}_{\textit{RoI}} \label{eq_const1} \\
		& \forall\; i \in [1, N_u]; \; \sum_{t = t_{0_i}}^{t_{l_i}} {b_c}_{ti} \geq \mathcal{B}_d  \label{eq_const2}\\
		& \sum_{i=1}^{N_u}\underset{j \neq i}{\sum_{j = 1}^{N_u}} l_{ij} \geq N_u-1; \; l_{ij} = \begin{cases}
			1 &  \text{if} \quad d_{ij} \leq T_r \\
			0 &  \text{otherwise}			
		\end{cases} \label{eq_const3} \\
		&   \sum_{i=1}^{N_u-1} \sum_{j=i+1}^{N_u} \left[ \left( z_i^2 + z_j^2\right)\tan^2\left(\frac{\vartheta}{2}\right) \atantwo\left(Y, X\right) - d_{ij}Y \right] = 0  \label{eq_const4}\\
		& X = \frac{\left( z_i^2 - z_j^2\right)\tan^2\left(\frac{\vartheta}{2}\right) + d_{ij}^2}{2d_{ij}}; \; Y = \sqrt{\left(z_i\tan\left(\frac{\vartheta}{2}\right)\right)^2 - X^2} \nonumber \\
		& \forall\; i \in [1, N_u]; \; \frac{1}{R_{\min}}\sqrt{\frac{1}{\cos\mu_i}}\sqrt{\frac{2mg}{\rho SC_{{_{\scaleto{L_i}{4.5pt}}}}}} \label{eq_const5} \leq \dot{\psi}_{\max}
	\end{align}
\end{subequations}

The constraint \ref{eq_const1} forces the UAVs to remain within the operational search space ${\mathcal{S}_{\mathcal{P}}}_{\textit{RoI}}$; while constraints \ref{eq_const2} and \ref{eq_const3} intend to maximize the survivability of the flock. They also minimize the risks on the mission accomplishment by ensuring that: 1) the UAVs have enough energy for urgent maneuvers from taking-off at $t_0$ until landing at $t_l$; and 2) the airborne network is always connected (i.e., there exist a communication route connecting all UAVs together where mutual distances are within the transmission range $T_r$).\\
In addition to the constraints \ref{eq_const4} and \ref{eq_const5} which respectively ensure that the fields-of-view of UAVs should not overlap so that to augment the probability of detecting more targets and by implication avoid any conflicts or collisions. And with \ref{eq_const5}, the UAVs should refrain from making abrupt turns that do not compel with their aerodynamic limitations (i.e., the turning radius is less than $R_{\min}$ or the turning rate is superior to $\dot{\psi}_{\max}$).

\section{Proposed approach}
\label{sec4}

\begin{figure*}
	\centering
	\hspace*{-0.2in}
	\includegraphics[scale=0.35]{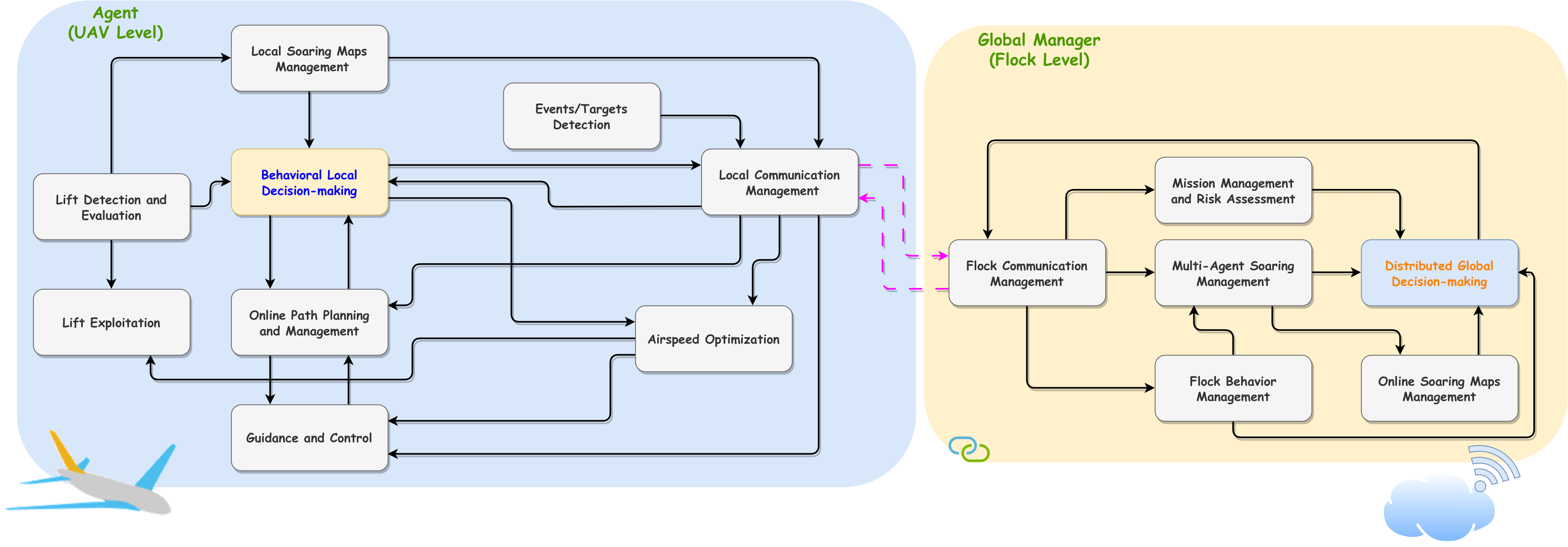}
	\caption{ High-level depiction of the components in the framework (at UAV and flock levels) }
	\label{fig_components}
\end{figure*}

This section gives a comprehensive explanation of the proposed cooperative soaring framework and further describes its underlying components.\\
As portrayed in Fig. \ref{fig_components}, each agent incorporates a set of sub-modules dedicated for a definite role. The most crucial one enables the agent to make decisions according to a local behavioral logic aimed at maximizing the collection of short-term rewards while interacting with the environment. It progressively aids the agent to learn choosing the actions that build up its profit. Such locally-made decisions should render the agent more autonomous and less dependent on external commands and namely in situations where the agent need to make an immediate maneuver.\\
The other sub-modules act as sensors and actuators to provide the core component with the necessary data and feedback. This data may be interpreted as relay waypoints, guidance commands that lead to a steady and safe flight with maximum gliding range, or MacCready speed-to-fly yielding minimum sink rate. In addition to defining the state conditions for triggering explicit behaviors like latching onto a lift or engaging in a powered climb.\\

And since the agents move as per their own benefit (short-term reward functions), a global manager with forward-looking action policy is needed to coordinate the behavior transitions of agents. By means of a distributed rule-based decision-making sub-module, the global manager probes into the future cost of actions prioritizing the overall profit of the mission (long-term reward functions). This could be for instance, the future potentiality of energy gain, targets detection and risks.
The associated action policy can be established either online using the communicated data of all agent states or offline using a model of behaviors switching rationale.\\
To maintain the system's resiliency and autonomy, the global manager level can be integrated either:
\begin{itemize}
	\itemsep0em
	\item [-] Within all agents but activated in only one agent acting as a leader and in case any failure occurs a new leader can be elected.
	\item [-] Within an accompanying backup UAV whose sole role is to coordinate between the agents.
	\item [-] Within a mobile edge computing node deployed in the proximity of the region-of-interest. \\
\end{itemize}

It is noteworthy to affirm that the decisions of the global manager are not absolute for cases where an agent is or soon-to-be in a critical state. In such cases, the decisions issued by the local decision-making module are prioritized for the safety of the agent. More details on how the agents operate during the surveillance mission and how the underlying components of the system work are provided in the next subsections.

\subsection{Cooperative surveillance and soaring}
\label{sec4.1}

\begin{figure*}[!h]
	\centering
	\hspace*{-0.25in}
	\includegraphics[scale=0.35]{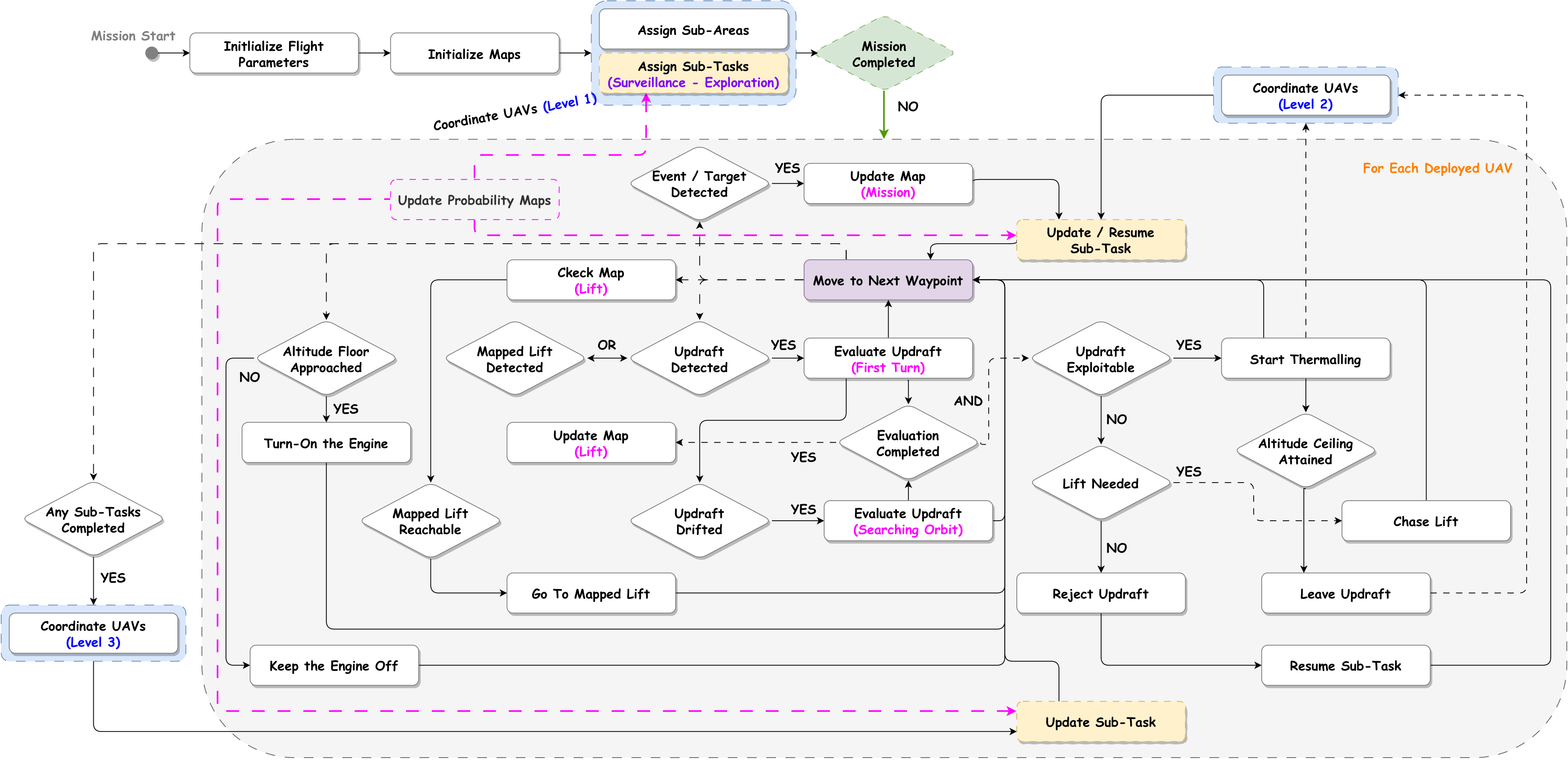}
	\caption{ Illustration of the discrete behaviors (states /transitions) that make up the soaring surveillance system }
	\label{fig_states}
\end{figure*}

During their mission of prolonging the endurance and surveillance, the actions of agents are encoded as discrete states. Each state comes with a distinct role and a finite set of heuristic-prompted transitions between other states. An abridged illustration is given in Fig. \ref{fig_states} to show the behaviors that may arise as a result of combining a set of actions.\\
Furthermore, Algorithm \ref{alg_soaring} outlines the different steps and flight operational modes. As can be observed, the $N_u$ agents kick-off the mission by initializing the necessary parameters like $ \textit{FLT}_{\textit{data}}$, which stores the real-time flight variables (e.g., battery level, bank angle, path angle, airspeed, ground speed, etc.). In addition to a set of feature-based ($\mathcal{M}_{\textit{lift}}, \mathcal{M}_{\textit{mission}}$) and grid-based ($\mathcal{M}_{\textit{lift\_prob}}$) maps, where the environment is divided into cells; each holding an estimate of location, a state and other relevant information.\\
A first and basic level of coordination is invoked at the beginning of the mission. Each agent is tasked to carry out a sub-mission $\mathcal{T}$ in a specific sub-area $\mathcal{A}$ depending on the parameter \textit{RoI}$_\rho$, indicating whether the UAVs should operate on the entire area (\textit{RoI}$_\rho=0$) or on a divided sub-areas (\textit{RoI}$_\rho=1$).
Here, the agents can choose one of the two allowable sub-missions that seek to gather more knowledge about the environment. Both sub-missions: $\langle \texttt{surveillance}\rangle$ and $\langle \texttt{exploration}\rangle$ come with two operational modes under which the agents trigger the "\textit{follow coverage planner}" and "\textit{explore}" states portrayed in Fig. \ref{fig_states}.

The former state includes a $\langle \texttt{sweep}\rangle $ mode, where the agent follows definite lawnmower-based patterns and an $\langle \texttt{expand\_sweep}\rangle $ mode, in which an expanding concentric pattern is tracked. Likewise for the latter, two modes: $\langle \texttt{global}\rangle $ and $\langle \texttt{local}\rangle $ are involved with a difference in the exploration range. More details regarding these sub-mission modes and how waypoints are generated will be given in subsection \ref{sec4.3}\\
At this level, the global manager will make sure that:
\begin{itemize}
	\itemsep0em
	\item [-] At least one agent is performing a surveillance sub-mission.
	\item [-] If multiple agents are with a same surveillance mode above a same sub-area, they should not follow the same pattern in order to efficiently gather more information. \\
\end{itemize}

The surveillance mission persists as long as there are still agents with enough battery capacity. Any target found at the instant $t$ and failing to satisfy any of these three conditions will be ignored. 1) the distance towards an agent should fall within the coverage field-of-view of that agent. 2) the time interval between two consecutive detections of the same target $|\Delta^t_{\textit{target}}-t| $ should be greater than a threshold $\delta_t $ to avoid needlessly saturating the $\mathcal{M}_{\textit{mission}}$ map, where $\Delta^t_{\textit{target}}$ is the detection time of the target. 3) the monitoring time of the target does not surpass a threshold $\tau_t$ (i.e., $|\Delta^{t_0}_{\textit{target}}-t| \leq \tau_t $, here $\Delta^{t_0}_{\textit{target}}$ is the time at which the target is detected for the first time). Otherwise, new data entries will be inserted in the corresponding map.\\
Moreover, as agents continuously look for a lift source to stay aloft, they may inadvertently encounter an unmapped updraft $\eta_{\textit{lift}}$ or by intentionally moving towards to a mapped lift if it has not faded yet. In the latter case, the "\textit{go to mapped lift}" state can be activated by one of three criteria. 1) the distance to reach the mapped lift is inferior to a threshold $\Delta_{\textit{map}}$ (in meters) defining the proximity to the lift. 2) the mapped lift has not been evaluated or exploited since a period $\delta_l$ (i.e., $|\nabla^t_{\textit{lift}}-t| \geq \delta_l$, where $|\nabla^t_{\textit{lift}}$ is the last time at which the updraft has been entered). 3) the agent is losing altitude and approaching the minimum MSL (Mean Sea Level) altitude $\mathcal{Z}$.\\ Supplementary metrics are used to judge the decision of going to a mapped lift (e.g., expected altitude gain, flight time, altitude clearance violation on arrival at the destination, battery consumption in the worst case etc.)\\

Now, once an updraft is found, the agent routinely engages in a "\textit{first turn}" state. It performs one circular orbit with a turn direction freely chosen on entry and a constant radius around the estimated location of the updraft's core. This orbit should be completed in order to sufficiently evaluate the newly detected updraft unless the "\textit{maintain safety}" state aborts the action. Another likely case occurs when an agent reaches the mapped lift location but could not find it due to being drifted or dissipated. Here, the "\textit{searching orbit}" state will be entered to compensate for such inaccuracies by pursuing a large circular orbit around the mapped location.

Upon completing the evaluation of an updraft, the challenge of staying within it and efficiently exploiting it to gain altitude must be addressed. One potential solution is to approximate the expected climb rate $w_c$, which in turn will be used to determine the optimal airspeed and bank angle. To this end, if an agent deems that the updraft is strong enough, it should adjust the center of its circular flight path to coincide with the updraft's core. An updraft is assumed to be exploitable if the conditions to reject the lift are not met, while the energy rate $\dot{E}$ and acceleration $\ddot{E}$ (Eq. \ref{eq_energy}) attain definite threshold values.

\begin{subequations}
	\footnotesize
	\begin{align}
		&	E = \frac{V_a^2}{2g} + z_u\\
		&   \dot{E} = \frac{V_a\dot{V_a^2}}{g} + \dot{z_u}\\
		&   \ddot{E} = \frac{\dot{V_a^2} + V_a\ddot{V_a}}{g} + \ddot{z_u}
	\end{align}
	\label{eq_energy}
\end{subequations}

A feedback control law similar to that of \cite{dobrokhodov2014cooperative} can be considered for issuing the turn rate commands $\dot{\psi}_{\textit{cmd}}$ for the agent's autopilot (see Fig. \ref{fig_DLnT} for further details on the employed control strategy). $\mathcal{K}_\psi$ in Eq. \ref{eq_turn_cmd} is the feedback gain and $\dot{\psi}_{\textit{ss}} = \frac{V_a}{R_{\textit{cmd}}}$ is the steady state turn rate for sustaining a constant climb rate and $R_{\textit{cmd}}$ is the desired flight radius, which depends on the updraft's size.

\begin{equation}
	\small
	\dot{\psi}_{\textit{cmd}} = 	\dot{\psi}_{\textit{ss}} - \mathcal{K}_\psi\ddot{E}
	\label{eq_turn_cmd}
\end{equation}

Still, agents have another possibility to exploit updrafts and namely if they desperately need to regain altitude and no strong lift is found. In this case, the agents may not follow a circular orbiting pattern but they will adapt their flights to track the strongest regions of lift within an updraft while triggering the "\textit{chase lift}" state. Decisions regarding the acceptance of weak lift sources may vary accordingly with the level of knowledge that an agent can acquire. The more interactions with the environment the more effective the learning and the more accurate decisions are made.

A second level of coordination between UAVs can be called forth when an agent $A_x$ starts exploiting an updraft. The global manager ensures that the sub-area assigned to the soaring agent would not be left without any surveillance or exploration being performed.\\
The operating principle of the global manager can be mathematically summarized as: $\forall\; i,j \in [1, N_u]$; \; $\exists\; X_{ij}^* \in \mathfrak{S}_G = \mathfrak{S}_{L_i} \cup \mathfrak{S}_{L_j}$ such that the probability $\mathsf{P}$ is maximized:

\begin{equation}
	\scriptsize
	\max\; \mathsf{P}\left(\mathfrak{R}^+_{G}(X_{ij}^*) > \delta_\chi\mathfrak{R}^-_{G}(X_{ij}^*) \Bigg| \sum_{i=1}^{N_u}\left[\mathfrak{R}^{t+1}_{L_i}(X_i^*) - \mathfrak{R}^{t}_{L_i}(X_i) \geq 0 \right] \geq \varsigma \right)
\end{equation}

where $\mathfrak{S}_{L_i}$ is a set of available local actions of an agent and $X_{ij}^*$ is the set of actions recommended by the manager for each agent. $\mathfrak{R}^{t}_{L_i}(X_i)$ is the reward value of an action carried out by an agent at an instant $t$. $\varsigma$ represents the majoritarian number of agents to approve the set $X_{ij}^*$, whereas $\mathfrak{R}^+_{G}$ and $\mathfrak{R}^-_{G}$ are respectively the global reward value for the mission and the penalty value for the plausible risk on the mission and $\delta_\chi$ being a risk tolerance threshold.\\
The objective is to maximize the global benefit of the whole mission (i.e., the global reward should be greater than the tolerable risk, which is likely to frequently occur). This global strategy is built from the actions of the UAVs, and so the recommended action for each UAV by the manager should also be beneficial to all the UAVs (i.e., not worse than what they gained at the instant $t-1$). Since it is unlikely to satisfy this constraint for all the UAVs, the global strategy should be accepted if is profitable for a certain number of UAVs.

And so, the manager will evaluate the set of available and feasible actions using several metrics like distances, current state of other agents (altitude, battery level, flight mode, sub-mission). It then decides if an agent $A_y$ should switch to an appropriate mode of surveillance or exploration, or take over the mission carried out the soaring agent $A_x$ before engaging in such maneuver. This agent $A_y$ will get a new set of temporary waypoints for its new sub-mission and once the agent $A_x$ completes its lift exploitation, agent $A_y$ should automatically resume its initial sub-mission. The objective is to maintain the persistency of surveillance and information gathering.\\

A final level of coordination is launched when an agent or more complete their assigned sub-missions. The global manager may command these agents to either substitute their assigned sub-areas, proceed with the same previous sub-mission or engage in a new different one. This decision is made by virtue of additional metrics depending on the current status of agents, their performance (monitored targets, etc.) and the information they gathered (e.g., how many updrafts have detected /exploited).\\

One last point to be discussed regarding the evolution of the lift map $\mathcal{M}_{\textit{lift}}$, whose cells incorporate memories of where updrafts have been encountered. The map is continuously updated using the information collected by agents through their surveillance /exploration and new data entry is inserted whenever a new updraft is evaluated. A decaying weight $f_{\textit{decay}}(x)$ is associated with the content of a cell in the map, where a lift has been detected for the first time at an instant $\Delta^{t_0}_{\textit{lift}}$. The goal is to let all the agents accessing the map would be aware and avoid chasing a lift that may have become weak if not dissipated. $c_{\max}$ in Eq. \ref{eq_decay} represents an estimated value of the updrafts' lifecycle in the adopted model.

\begin{equation}
	\small
	f_{\textit{decay}}(x) = \frac{-1}{c_{\max}}x + 1; \; x = |\Delta^{t_0}_{\textit{lift}}-t| \in [0, c_{\max}]
	\label{eq_decay}
\end{equation}

\begin{algorithm}
	\scriptsize
	\caption{ Cooperative persistent surveillance  }
	\label{alg_soaring}
	\begin{algorithmic}[1]
		\STATE $\textit{FLT}_{\textit{track}} \gets [\;]; \; t \gets T_0$
		\STATE $[\textit{FLT}_{\textit{flag}}, \textit{FLT}_{\textit{data}}, \textit{FLT}_{\textit{mode}}] \gets \textbf{\texttt{INITIALIZE\_FLIGHT\_CONDITIONS}}(N_u,\textit{UAV}_{\textit{data}})$
		
		\STATE $ \mathcal{A} \gets \textbf{\texttt{ASSIGN\_AREAS}}(\textit{RoI}_\rho,N_u,LB_{\textit{RoI}}, UB_{\textit{RoI}})$
		\STATE $ \mathcal{T} \gets$ \textbf{\texttt{ASSIGN\_TASKS}}($N_u$) \hfill\COMMENT{\textcolor{gray}{Algorithm \ref{alg_waypoint}}} 
		\STATE $ [\mathcal{W_P}, \textit{FLT}_{\textit{data}}] \gets \textbf{\texttt{COORDINATE\_UAVs\_LEVEL1}}( \mathcal{A}, \mathcal{T},\textit{FLT}_{\textit{data}} )$
		\STATE $[\mathcal{M}_{\textit{lift}},\mathcal{M}_{\textit{lift\_prob}},\mathcal{M}_{\textit{mission}}] \gets \textbf{\texttt{INITIALIZE\_MAPS}}(LB_{\textit{RoI}}, UB_{\textit{RoI}})$
		\STATE $ t \gets t + \Delta_t$
		
		\WHILE{$t \leq T_s$}
		\FORALL{$u \in [1, N_u]$}
		\IF{$\textit{FLT}_{\textit{flag}}[u] = \textit{true}$}
		\STATE $\mathcal{ID}_{\textit{target}} \gets (x_t-x_u)^2 + (y_t - y_u)^2 \leq z_u^2\tan^2(\frac{\vartheta}{2})$
		\IF{$\mathcal{ID}_{\textit{target}} \neq \emptyset\; \land \; |\Delta^t_{\textit{target}}-t| \geq \delta_t \; \land \; |{\Delta^{t_0}}_{\textit{target}}-t| \leq \tau_t$}
		\STATE $\mathcal{M}_{\textit{mission}} \gets \textbf{\texttt{UPDATE\_MAP}}(\mathcal{M}_{\textit{mission}},(x_t,y_t),\mathcal{ID}_{\textit{target}})$
		\ENDIF
		\STATE $ D_{\textit{lift}} \gets \textbf{\texttt{CHECK\_MAPPED\_LIFT}}(\mathcal{M}_{\textit{lift}},(x_u,y_u))$
		
		\IF{$z_u \leq \mathcal{Z}$}
		\STATE $\textit{LIFT}_{\textit{needed}} \gets \textit{true}$
		\STATE $ \mathcal{W_P}[u] \gets \textbf{\texttt{GOTO\_MAPPED\_LIFT}}(\mathcal{W_P}[u],  \mathcal{M}_{\textit{lift}})$
		\ENDIF
		
		\STATE $\eta_{\textit{lift}} \gets |z^{t-1}_u-z^t_u| \neq \sigma_u \; \land\; \mathcal{ID}_{\textit{lift}} \notin \mathcal{M}_{\textit{lift}} $
		\IF{$\eta_{\textit{lift}}\; \lor\; (\exists\; D_{\textit{lift}} \leq \Delta_{\textit{map}}\; \land\; {V_w}_{_{\textit{map}}} \neq 0 \; \land\; |\nabla^t_{\textit{lift}}-t| \geq \delta_l ) $}
		\STATE $\textit{SOAR}_{\textit{flag}}[u] \gets \langle\texttt{first\_turn} \rangle$
		\ELSIF{$ \neg\; \eta_{\textit{lift}} \land (\exists\; D_{\textit{lift}} \leq \Delta_{\textit{map}} \; \land  {V_w}_{_{\textit{map}}} = 0 \;  \land  |\nabla^t_{\textit{lift}}-t| \geq \delta_l)  $}
		\STATE $\textit{SOAR}_{\textit{flag}}[u] \gets \langle\texttt{searching\_orbit} \rangle$
		\ENDIF
		
		\IF{$\textit{SOAR}_{\textit{flag}}[u] = \langle \texttt{eval\_completed} \rangle $}
		\STATE $[\mathcal{M}_{\textit{lift}},\mathcal{M}_{\textit{lift\_prob}}] \gets \textbf{\texttt{UPDATE\_MAP}}(\mathcal{M}_{\textit{lift}},\mathcal{M}_{\textit{lift\_prob}}, \textit{METEO}_{\textit{data}})$
		\IF{$ w_c \geq {w_c}_{{_{\scaleto{\min}{2.5pt}}}}\; \land\; \neg\; \textit{LIFT}_{\textit{rejected}}$}
		\STATE $\textit{SOAR}_{\textit{flag}}[u] \gets \langle\texttt{exploit} \rangle$
		\ELSIF{$w_c < {w_c}_{{_{\scaleto{\min}{2.5pt}}}}\; \land\; \textit{LIFT}_{\textit{needed}} \; \land\; \neg\; \textit{LIFT}_{\textit{rejected}}$}
		\STATE $\textit{SOAR}_{\textit{flag}}[u] \gets \langle\texttt{chase\_lift} \rangle$
		\ENDIF
		\STATE $[\mathcal{W_P},\textit{FLT}_{\textit{data}}] \gets \textbf{\texttt{COORDINATE\_UAVs\_LEVEL2}}(\mathcal{W_P},\mathcal{A},\mathcal{T},\textit{FLT}_{\textit{data}})$
		\ENDIF

		\IF{$\textit{SOAR}_{\textit{flag}}[u] \neq \emptyset$}
		\STATE $ \textbf{\texttt{UPDATE\_STATE}}((x_u, y_u, z_u),\textit{SOAR}_{\textit{flag}}[u],\textit{FLT}_{\textit{data}}[u])$ \hfill\COMMENT{\textcolor{gray}{Algorithm \ref{alg_path}}} 
		\ELSE
		
		\IF{$z_u \geq \mathcal{Z}$}
		\STATE $\textit{FLT}^{\;t}_{\textit{mode}}[u] \gets \langle \texttt{glide} \rangle  $
		\STATE $ \textbf{\texttt{UPDATE\_STATE}}((x_u, y_u, z_u),\textit{FLT}^{\;t}_{\textit{mode}}[u],\textit{FLT}_{\textit{data}}[u])$
		
		\ELSIF{$(\textit{FLT}^{\textit{ t--}1}_{\textit{mode}}[u] = \langle engine \rangle \land z_u \leq \mathcal{Z})\; \lor\; (\textit{FLT}^{\textit{ t--}1}_{\textit{mode}}[u] = \langle glide \rangle \; \land \; z_u \leq \mathcal{Z}_{\min})$}
		\STATE $\textit{FLT}^{\;t}_{\textit{mode}}[u] \gets \langle \texttt{engine} \rangle  $
		\STATE $ \textbf{\texttt{UPDATE\_STATE}}((x_u, y_u, z_u),\textit{FLT}^{\;t}_{\textit{mode}}[u],\textit{FLT}_{\textit{data}}[u])$
		\STATE $\textit{FLT}_{\textit{data}}[u].\textit{battery\_capacity} \gets \textit{FLT}_{\textit{data}}[u].\textit{battery\_capacity} - b_{c}$
		\ENDIF
		\ENDIF
		
		\IF{$\textit{FLT}_{\textit{data}}[u].\textit{battery\_capacity} < \mathcal{B}_d$}
		\STATE $\textit{FLT}_{\textit{flag}}[u] \gets \textit{false}$
		\ENDIF
		
		\IF{$\mathcal{W_P}[u] = \emptyset$}
		\STATE $\mathcal{T}[u].{\textit{progress}} \gets \langle \texttt{completed} \rangle$
		\ENDIF
		
		\ENDIF
		
		\STATE $\textit{FLT}_{\textit{track}}[u] \gets  \textbf{\texttt{APPEND}}((x_u,y_u,z_u),\textit{FLT}_{\textit{data}}[u],\mathcal{ID}_{\textit{target}},\mathcal{ID}_{\textit{lift}} )$
		\ENDFOR
		
		\STATE $[\mathcal{W_P},\mathcal{A},\mathcal{T},\textit{FLT}_{\textit{data}}] \gets \textbf{\texttt{COORDINATE\_UAVs\_LEVEL3}}(\mathcal{W_P},\mathcal{A},\mathcal{T},\textit{FLT}_{\textit{data}})$
		
		\STATE $\mathcal{M}_{\textit{lift}} \gets \textbf{\texttt{DECAY\_MAP}}(\mathcal{M}_{\textit{lift}},t,\Delta^{t_0}_{\textit{lift}})$
		\STATE $t \gets t + \Delta_t$
		\ENDWHILE
		\RETURN $ \textit{FLT}_{\textit{track}} $
	\end{algorithmic}
\end{algorithm}

\subsection{Decision-making policy}
\label{sec4.2}

Considering a set of $N_a$ available and $N_f$ feasible (predicted) actions $\mathfrak{S} = \bigg\{ \underset{i \leq N_a}{X_i} \bigcup  \underset{j \leq N_f}{\widehat{X}_j}\bigg\}$ forming a decision matrix $DM\times(-1)^{\Lambda}$. Each row (i.e., action) is described by its $N_r$ reward or penalty scores (a penalty can be viewed as a negative reward). $\Lambda = \{0,1\}$ here reflects the nature of the reward score either to be minimized or maximized. The objective is to identify the great trade-off leading to an action that maximizes the benefit of an agent or the multiple agents. This logic is implemented within the two decision-making levels in the framework, in other words, the behavioral local and distributed global sub-modules (illustrated in Fig. \ref{fig_components}).\\
By leveraging the concepts of prospect theory and rational choice theory, decisions will be made according to a set of rules described as follows:
\begin{itemize}
	\itemsep0em
	\item [-] The outcomes of a decision are relatively ranked with respect to a "neutral" reference point $\mathcal{N}_\mathfrak{F}$. Then by labeling those, which are inferior as losses and the greater ones as equivalent gains, the outcome that has a potential to avoid probable losses is preferred. The metrics expressed in Eq. \ref{eq_DM} help avert the risks and prioritize the more profitable outcomes:
	
	\begin{subequations}
		\footnotesize
		\begin{align}
			{\Xi_1}_i & = -\sum_{j = 1}^{N_r}\Big| DM_{ij} - {\mathcal{N}_\mathfrak{F}}_j \Big|; \; \forall\; i \in [1, N_a+N_f] \label{eq_DM1} \\
			{\Xi_2}_i & = -\sum_{j = 1}^{N_r} \begin{cases}
				1  & \text{if} \; \nexists\;{{\mathcal{N}_\mathfrak{F}}_{\widetilde{j}}}  < {DM_i}_j\; \land\; \exists\; {DM_i}_j < {{\mathcal{N}_\mathfrak{F}}_{\widetilde{j}}} \\
				0  & \text{otherwise} \; \forall\; \widetilde{j} \in [1,N_r]
			\end{cases} 
			\label{eq_DM2}
		\end{align}	
		\label{eq_DM}
	\end{subequations}
	
	\item [-] Under uncertainty, an agent subjectively tends to seek an outcome with maximum expected utility. But without complete information, the agent may experience regret when discovering later that a different choice would have leaded to a better outcome.\\ In order to quantify the difference in value between what would have been the optimal and the already made decision, the following metrics are respectively used to gauge the the closeness to the ideal utility, diversity and similarity between the outcomes of the possible actions:
	
	\begin{subequations}
		\footnotesize
		\begin{align}
			{\Xi_3}_i & = \sum_{j = 1}^{N_r}\Big| DM_{ij} - \min\left( DM_{j}\right) \Big| \label{eq_DM3} \\
			{\Xi_4}_i & = \sum_{j = 1}^{N_r} \sqrt{(j-\widehat{j})^2 + (DM_{ij}-DM_{i\widehat{j}})^2}; \; \widehat{j} = (j \bmod N_r)+1  \label{eq_DM4} \\
			{\Xi_5}_i & = \frac{1}{2}  \underset{j = 0, 1, \ldots, N_r-1}{\Big| (\widehat{j}\times DM_{i\widetilde{j}}) - (DM_{i\widehat{j}}\times\widetilde{j}) \Big|}; \; \begin{bmatrix}
				\widehat{j} \\
				\widetilde{j}
			\end{bmatrix} = (j \bmod N_r) + \begin{bmatrix}
				1 \\
				2
			\end{bmatrix}  \label{eq_DM5}
		\end{align}
	\end{subequations}
	
	\item [-] For the next step, a new decision matrix made up with previously discussed metrics is needed. From which the strictly dominated combinations will be discarded and only those which are non-dominated and considered. \\
	Following this logic, the action selected to be executed is none other than the one with the best aggregated $\Xi$ scores, i.e.,\\ $\footnotesize\mathfrak{S}\left(\texttt{Argwhere}\left(\mathfrak{S} == \min\left(\sum_{j = 1}^{5}DM_{kj}^{\text{new}}\right)\right)\right)$,\\ $k\in[1,N_a+N_f]\; \land \; \left({DM_{k_i}^{\text{new}}} \leq {DM_{k_j}^{\text{new}}} \right)\; \land\; \exists\; {DM_{k_i}^{\text{new}}} < {DM_{k_j}^{\text{new}}}$ \; $\forall\; i,j\neq i \in [1,5]$.
	
	\begin{equation}
		\footnotesize
		DM^{\text{new}} = \begin{bmatrix}
			{\Xi_1}_1  &  {\Xi_2}_1  &  {\Xi_3}_1  &  {\Xi_4}_1  &   {\Xi_5}_1 \\
			{\Xi_1}_2  &  {\Xi_2}_2  &  {\Xi_3}_2  & {\Xi_4}_2   &   {\Xi_5}_2 \\
			\vdots  &  &  \ddots &   &  \vdots \\
			{\Xi_1}_{N_a+N_f}  &  &  \ldots  &   &   {\Xi_5}_{N_a+N_f}
		\end{bmatrix}
	\end{equation}
	
\end{itemize}

\subsection{Path planning}
\label{sec4.3}

To navigate across the assigned area of operation, a waypoint generation module (Algorithm \ref{alg_waypoint}) is invoked to supply the agents with a set of destinations to be visited in accordance with their ongoing sub-mission.
For a $\langle \texttt{sweep} \rangle$ search surveillance, the sub-area assigned to an agent defined by its $N_v$ vertices $V_{\textit{RoI}}$ and $N_e$ edges $E_{\textit{RoI}}$ should be rearranged to start from the edge whose perpendicular distance form all vertices is the largest (i.e., $\max(\Vert \overrightarrow{V_{\textit{RoI}}[j]E^{\perp}_{\textit{RoI}}} \Vert); \; \forall\; j \in [1, N_v]$).  Then, a set of coverage waypoints $(x_p, y_p)$ is obtained by: 1) generating a group of parallel lines inclined with a sweeping degree of $\xi_{{_{\scaleto{A}{2.5pt}}}}$ and spaced by a coverage width $\xi_{{_{\scaleto{W}{2.5pt}}}}\geq 2\mathcal{Z}_{\min}\tan(\frac{\vartheta}{2})$. Here, $\vartheta$ is the UAV's field-of-view angle and $\mathcal{Z}_{\min}$ is the minimum MSL (mean sea level) altitude. 2) intersecting these lines with the edges of the new arranged polygon starting from the points located in the direction $\xi_{{_{\scaleto{D}{2.5pt}}}}=\{ 1, -1 \}$ either clock-wise or anti-clock-wise.

Similarly, the $\langle \texttt{expand\_sweep} \rangle$ surveillance mode works in such a way to increase the scanning frequency by dividing the assigned sub-area into concentric expanding polygons. The orientation of these polygons is governed by $\xi_{{_{\scaleto{D}{2.5pt}}}}$ while their number depends on: $\xi_{{_{\scaleto{W}{2.5pt}}}}$ and the maximum distance between the polygon's centroid and all its vertices. The starting vertex is chosen based on two criteria: 1) small deviation $|\psi_p-\psi_u|$ between the agent's heading and the angle between the centroid and such a vertex; 2) small distance towards the centroid.\\

Regarding the exploration sub-missions, a priority map $\mathcal{M}_{\textit{priority}}$ is required to guide the agents as they seek lift or search for targets. This map depends on the probability of detecting updrafts either by the cooperating agents or weather forecasts (wind, moisture, insolation, etc.) as well as the probability of detecting targets. Once constructed or updated, a set of the most promising and distinct waypoints will be selected from this map using a roulette wheel selection mechanism. It should be noted that $\mathcal{M}_{\textit{priority}}$ characterizes the entire region-of-interest. Also, waypoints should satisfy one of these two conditions: 1) they should strictly belong the sub-area assigned to each agent; 2) they should be within the proximity range $\mathcal{Q}_u$ of an agent performing local exploration.\\

In this context, Algorithm \ref{alg_path} outlines the steps that an agent located at $\mathcal{P}_u = [x_u, y_u, z_u]$ may follow to reach a destination $\mathcal{P}_d$. The $H{{_{\scaleto{w}{4pt}}}}H$ path planning algorithm is strongly connected with the "\textit{maintain safety}" state (Fig. \ref{fig_states}). As it names implies, it enables the agent to navigate hop by hop between waypoints and within horizons (delimited uniform space-time with length $\mathcal{H_{{_{\scaleto{\mathcal{L}}{3pt}}}}}$ and duration $\mathcal{H_{{_{\scaleto{\mathcal{T}}{3pt}}}}}$). In the $H{{_{\scaleto{w}{4pt}}}}H$ planning scheme, the agent is assumed to be surrounded by a safety space of radius $R_{\textit{safe}}$ that should give it enough time for evasive maneuvers to escape collisions in urgent cases.

Moreover, two occupancy concepts are employed in the approach: recorded and predicted occupancy spaces. The former is meant for visible obstacles where the agent keeps track of the current and previous movements of obstacles (velocity, heading and flight path) within a definite time window in case the agent needs to follow a backtracking path. The latter is destined for the obstacles that might be encountered in the future. An agent can predict the likely trajectories of other moving obstacles (agents in this work) on the basis of communicated data indicating their current state, thus helping to improve its environmental awareness.\\
Putting it together, the agent should have access to a variety of feasible and safe trajectories, from which the optimal one will be chosen.\\

Again, as stated in Algorithm \ref{alg_path}, the agent first checks if the destination is within its visibility range $\mathcal{V}_{\textit{free}}$ and no obstacle is expected to be encountered for a time interval $\Delta_t \leq \mathcal{H_{{_{\scaleto{\mathcal{T}}{3pt}}}}}$. When these conditions are met the agent straightly "hops" with a distance $T_D$ and a flight angle $ \gamma_d = \arccos{\frac{\overrightarrow{\mathcal{R}ef} \cdot \overrightarrow{\mathcal{P}_u \mathcal{P}_d}}{\Vert \overrightarrow{\mathcal{R}ef} \Vert \Vert \overrightarrow{\mathcal{P}_u \mathcal{P}_d} \Vert}}$ and repeatedly keeps this behavior until reaching the destination. Here, $\overrightarrow{\mathcal{R}ef}$ is the reference vector and ($\cdot$) denotes the dot product.

However, if any of these two criteria are present: either an obstacle is predicted to be encountered $(\mathcal{V}_{\mathcal{P}} \neq \emptyset)$ or a visible obstacle is nearby $(\mathcal{V}_{\mathcal{O}} \neq \emptyset)$, then the agent switches to another behavior allowing it to construct a search space $\mathcal{S_P}$, and create a tree of candidate safe trajectories $\mathcal{L}_{\textit{paths}}$.
To this end, the following path planning functions should be considered:

\begin{subequations}
	\footnotesize
	\begin{align}
		&  \min \;   \frac{1}{1 + \exp\left(\Vert \overrightarrow{\mathcal{P}_i(t)\mathcal{P}_o(t)} \Vert-\varrho R_{\textit{safe}}\right)};  \forall\; i \in [1, N_u]; \forall\; o \in [1, N_o]  \label{eq_path1}\\
		&  \min \;  \Vert \mathcal{P}_i(t) - \mathcal{P}_d(t) \Vert =  \begin{cases}
			\Vert x_{i}(t) - x_{d_i}(t) \Vert  \\
			\Vert y_{i}(t) - y_{d_i}(t) \Vert \\
			\Vert z_{i}(t) - z_{d_i}(t) \Vert
		\end{cases} \; \forall\; i \in [1, N_u]  \label{eq_path2} \\
		& \mathcal{P}_i(t) = \begin{bmatrix}
			x_{i}(t)   \\
			y_{i}(t) \\
			z_{i}(t) 
		\end{bmatrix} = \begin{bmatrix}
			\int_{t}^{\mathcal{H_{{_{\scaleto{\mathcal{T}}{3pt}}}}}} V_{a_i}(t)\cos\gamma_i(t)\cos\psi_i(t)dt  \\
			\int_{t}^{\mathcal{H_{{_{\scaleto{\mathcal{T}}{3pt}}}}}} V_{a_i}(t)\cos\gamma_i(t)\sin\psi_i(t)dt  \\
			\int_{t}^{\mathcal{H_{{_{\scaleto{\mathcal{T}}{3pt}}}}}} \left(V_w(t) + V_{a_i}(t)\sin\gamma_i(t)\right)dt
		\end{bmatrix} \nonumber
	\end{align}
\end{subequations}

where $\mathcal{P}_o$ being the coordinates of the $N_o$ obstacles and $\varrho$ a safety threshold.

Consecutive sub-destinations leading to $\mathcal{P}_d$ should be optimally selected from $\mathcal{L}_{\textit{paths}}$ along with choosing the appropriate airspeed $V_d$ that enables avoiding immediate or future collisions, i.e., $V_d = \{ \exists\; V_a \in [V_{a_{\min}}, V_{a_{\max}}] : V_a\Delta_t \;\cap \;  (\mathcal{S_P} \subset (\mathcal{V}_{\mathcal{P}} \cup \mathcal{V}_{\mathcal{O}})) = \emptyset \}$.

\begin{algorithm}
	\scriptsize
	\caption{  Waypoints generation  }
	\label{alg_waypoint}
	\begin{algorithmic}[1]
		\STATE $\mathcal{W_P} \gets [\;]$
		\STATE Let the wrapping function $\varpi(x,n) = ((x-1) \bmod n)+1$
		\IF{$\mathcal{T} = \langle \texttt{surveillance} \rangle$}
		\STATE $\psi_p \gets [\;]; \; D_p \gets [\;]$
		\STATE $\mathcal{C} = [x_c,y_c] \gets \textbf{\texttt{GET\_AREA\_CENTROID}}(V_{\textit{RoI}},E_{\textit{RoI}})$
		\STATE $\xi_{{_{\scaleto{W}{2.5pt}}}} \gets \textbf{\texttt{COMPUTE\_COVERAGE\_WIDTH}}(\vartheta, \Delta_z)$
		
		\IF{$\mathcal{T}.\textit{mode} = \langle \texttt{sweep} \rangle$}
		\FOR{$i \gets [1, N_e]$}
		\FORALL{$j \gets [1, N_v]$}
		\STATE $D_p \gets \textbf{\texttt{APPEND}}(\max(\Vert \overrightarrow{V_{\textit{RoI}}[j]E^{\perp}_{\textit{RoI}}[i]} \Vert))$
		\ENDFOR
		\ENDFOR
		\STATE $[\xi_{{_{\scaleto{A}{2.5pt}}}},V_{\textit{RoI}},E_{\textit{RoI}}] \gets \textbf{\texttt{COMPUTE\_SWEEP\_ANGLE}}(D_p, V_{\textit{RoI}}, E_{\textit{RoI}})$
		\STATE $[x_p, y_p] \gets\textbf{\texttt{SET\_COVERAGE\_PATTERN}}((V_{\textit{RoI}},E_{\textit{RoI}}), \xi_{{_{\scaleto{A}{2.5pt}}}},\xi_{{_{\scaleto{W}{2.5pt}}}},\xi_{{_{\scaleto{D}{2.5pt}}}})$ 
		\STATE $\mathcal{W_P} \gets \textbf{\texttt{APPEND}}(\texttt{STACK}([x_p, y_p], \mathcal{C}))$
		
		\ELSIF{$\mathcal{T}.\textit{mode} = \langle \texttt{expand\_sweep} \rangle$}
		\FOR{$i \gets [1, N_v]$}
		\STATE $\psi_p \gets \textbf{\texttt{APPEND}}(\textbf{\texttt{COMPUTE\_HEADING}}(\mathcal{C}, V_{\textit{RoI}}[i]))$
		\STATE $D_p \gets \textbf{\texttt{APPEND}}(\Vert \overrightarrow{\mathcal{C}V_{\textit{RoI}}[i]} \Vert)$
		\ENDFOR
		\STATE $n_p \gets \nint{\frac{1}{\xi_{{_{\scaleto{W}{2.5pt}}}}}\min(D_p)}$
		\STATE $v_{\textit{index}} \gets \textbf{\texttt{SELECT\_VERTEX}}(V_{\textit{RoI}},|\psi_p-\psi_u|,D_p)$
		\STATE $\psi_p \gets \psi_p(\varpi\left([v_{\textit{index}},v_{\textit{index}}+\xi_{{_{\scaleto{D}{2.5pt}}}},\ldots,v_{\textit{index}}+\xi_{{_{\scaleto{D}{2.5pt}}}}(N_v-1)], N_v\right))$
		\FOR{$j \gets [1, n_p]$}
		\FOR{$k \gets [1, N_v]$}
		
		\STATE $\mathcal{W_P} \gets \textbf{\texttt{APPEND}}([x_c +j\xi_{{_{\scaleto{W}{2.5pt}}}}\cos\psi_p[k], y_c +j\xi_{{_{\scaleto{W}{2.5pt}}}}\sin\psi_p[k]])$
		\ENDFOR
		\ENDFOR
		\STATE $\mathcal{W_P} \gets \textbf{\texttt{APPEND}}(\mathcal{C}))$
		\ENDIF
		
		\ELSIF{$\mathcal{T} = \langle \texttt{exploration} \rangle$}
		\STATE $\mathcal{M}_{\textit{priority}} \gets \textbf{\texttt{GET\_PRIORITY\_MAP}}(\mathcal{M}_{\textit{lift\_prob}},\mathcal{M}_{\textit{mission}})$
		\STATE $ n_p \gets \textbf{\texttt{GET\_DESTINATIONS\_NUMBER}}(\mathcal{A}_u, \Delta_z, \vartheta)$
		\STATE $\textit{FLAG}_\textit{region} \gets \mathcal{M}_{\textit{priority}} \in \mathcal{A}_u $
		\STATE $\textit{FLAG}_\textit{vicinity} \gets \textit{FLAG}_\textit{region} \; \land\; (\mathcal{M}_{\textit{priority}} \in \mathcal{Q}_u$)
		
		\FOR{$i \gets [1, n_p]$}
		\IF{$\mathcal{T}.\textit{mode} = \langle \texttt{global} \rangle$}
		\STATE $\mathcal{W_P} \gets \textbf{\texttt{APPEND}}({\texttt{ROULETTE\_WHEEL}}(\mathcal{M}_{\textit{priority}} \otimes \textit{FLAG}_\textit{region}))$
		
		\ELSIF{$\mathcal{T}.\textit{mode} = \langle \texttt{local} \rangle$}
		\STATE $\mathcal{W_P} \gets \textbf{\texttt{APPEND}}({\texttt{ROULETTE\_WHEEL}}(\mathcal{M}_{\textit{priority}} \otimes \textit{FLAG}_\textit{vicinity}))$
		\ENDIF
		\STATE $\mathcal{M}_{\textit{priority}} \gets \textbf{\texttt{UPDATE\_MAP}}(\mathcal{M}_{\textit{priority}},\mathcal{W_P})$
		\ENDFOR
		\ENDIF
		\RETURN $\mathcal{W_P}$
	\end{algorithmic}
\end{algorithm}

\begin{algorithm}
	\scriptsize
	\caption{  $H{{_{\scaleto{w}{4pt}}}}H$ path planning  }
	\label{alg_path}
	\begin{algorithmic}[1]
		
		\STATE $\textit{Path} \gets [\;]$
		\STATE $D \gets\Vert \overrightarrow{\mathcal{P}_u\mathcal{P}_d} \Vert$
		\WHILE{$D > 0$}
		\STATE $\mathcal{V}_{\textit{free}}  \gets \textbf{\texttt{COMPUTE\_VISIBILITY}}(\mathcal{P}_u, \mathcal{P}_d, (\mathcal{H_{{_{\scaleto{\mathcal{L}}{3pt}}}}}, \mathcal{H_{{_{\scaleto{\mathcal{T}}{3pt}}}}})) $
		\STATE $\mathcal{V}_{\mathcal{P}} \gets \textbf{\texttt{PREDICT\_OBSTACLES}}(\mathcal{P}, V_{a}, \gamma,\psi,(\mathcal{H_{{_{\scaleto{\mathcal{L}}{3pt}}}}}, \mathcal{H_{{_{\scaleto{\mathcal{T}}{3pt}}}}})) $
		
		\IF{$\mathcal{V}_{\mathcal{P}} = \emptyset \; \land\; \mathcal{V}_{\textit{free}} = \textit{true}$}
		\STATE $T_D \gets V_{a_u}\times\Delta_t$
		\IF{$\mathcal{P}_d \in \mathcal{V}_{\textit{free}} \; \land\ D \leq T_D$}
		\STATE $\textit{Path} \gets \textbf{\texttt{APPEND}}(\mathcal{P}_d) $
		\ELSE
		\STATE $\gamma_d \gets \textbf{\texttt{COMPUTE\_DESTINATION\_ANGLE}}(\mathcal{P}_u, \mathcal{P}_d)$
		\STATE $\textit{Path} \gets \textbf{\texttt{APPEND}}( [x_u + T_D\cos\gamma_d, y_u + T_D\sin\gamma_d, z_u+\Delta_z]) $
		\STATE $\mathcal{P}_u \gets \textit{Path}[\textit{end}] $
		\STATE $ D \gets D - T_D$
		\ENDIF
		
		\ELSE
		\STATE $\mathcal{V}_{\mathcal{O}}  \gets \textbf{\texttt{OBSERVE\_OBSTACLES}}(\mathcal{P}, V_a, \gamma,\psi,(\mathcal{H_{{_{\scaleto{\mathcal{L}}{3pt}}}}}, \mathcal{H_{{_{\scaleto{\mathcal{T}}{3pt}}}}}))$
		\STATE $ \mathcal{S_P} \gets \textbf{\texttt{BUILD\_SEARCH\_SPACE}}(\mathcal{P}_u,\mathcal{V}_{\mathcal{P}},\mathcal{V}_{\mathcal{O}},(\mathcal{H_{{_{\scaleto{\mathcal{L}}{3pt}}}}}, \mathcal{H_{{_{\scaleto{\mathcal{T}}{3pt}}}}}))$ 
		\STATE $ \mathcal{L}_{\textit{paths}} \gets \textbf{\texttt{SOLVE\_CONFLICT}}(\mathcal{P}, \mathcal{P}_d, \mathcal{S_P})$ 
		\WHILE{$\mathcal{V}_{\mathcal{P}} \neq \emptyset \; \lor \; \mathcal{V}_{\mathcal{O}} \neq \emptyset$}
		\STATE $[\chi_d, V_d] \gets \textbf{\texttt{SELECT\_DESTINATION}}(\mathcal{L}_{\textit{paths}}, \mathcal{P}_d)$ 
		\STATE $D \gets D - (V_d\times \Delta_t)$
		\STATE $\textit{Path} \gets \textbf{\texttt{APPEND}}(\chi_d) $
		
		\STATE $\mathcal{P}_u \gets \textit{Path}[\textit{end}] $
		
		\ENDWHILE
		
		\ENDIF
		
		\ENDWHILE
		
		\RETURN $\textit{Path}$

	\end{algorithmic}
\end{algorithm}

\subsection{Control strategy}
\label{sec4.4}

When a given UAV is not in a soaring flight mode, it navigates between waypoints through two distinct modes. As outlined in Algorithm \ref{alg_soaring}, the typical one is to glide towards a next practical destination point from its glide footprint whose shape is affected by wind and environment features as long as the agent has enough altitude. The agent is free to choose the optimal airspeed that leads to a minimum sink rate and therefore slower potential energy loss. The difference in altitude between two consecutive gliding points can be approximated as: $\Delta_z = -V_a\frac{ C_{{_{\scaleto{D}{4pt}}}}}{ C_{{_{\scaleto{L}{4pt}}}}}$.

Nonetheless, the agent can be forced to use its engine power to regain altitude in the expense of consuming its battery capacity if the altitude loss rate is too high and no lift is found or expected be reached in time.
A PID controller (Eq. \ref{eq_pid}) is used to enable the agents to track the commanded path regardless of disturbances based on the 6-DOF defined in section \ref{sec3}).

\begin{equation}
	\small
	u(t) = \mathcal{K}_pe(t) + \mathcal{K}_i\int_{0}^{t}e(t)dt + \mathcal{K}_d\frac{d(e(t))}{dt}
	\label{eq_pid}
\end{equation}

Five command signals (${\phi}_{\textit{cmd}},$ ${\theta}_{\textit{cmd}},$ ${\psi}_{\textit{cmd}},$ ${z}_{\textit{cmd}},$ ${V_a}_{\textit{cmd}}$) are considered for this powered-climb optimization process: roll, pitch, yaw, altitude and airspeed.

\begin{figure*}[!h]
	\centering
	\hspace*{-0.15in}
	\includegraphics[scale=0.37]{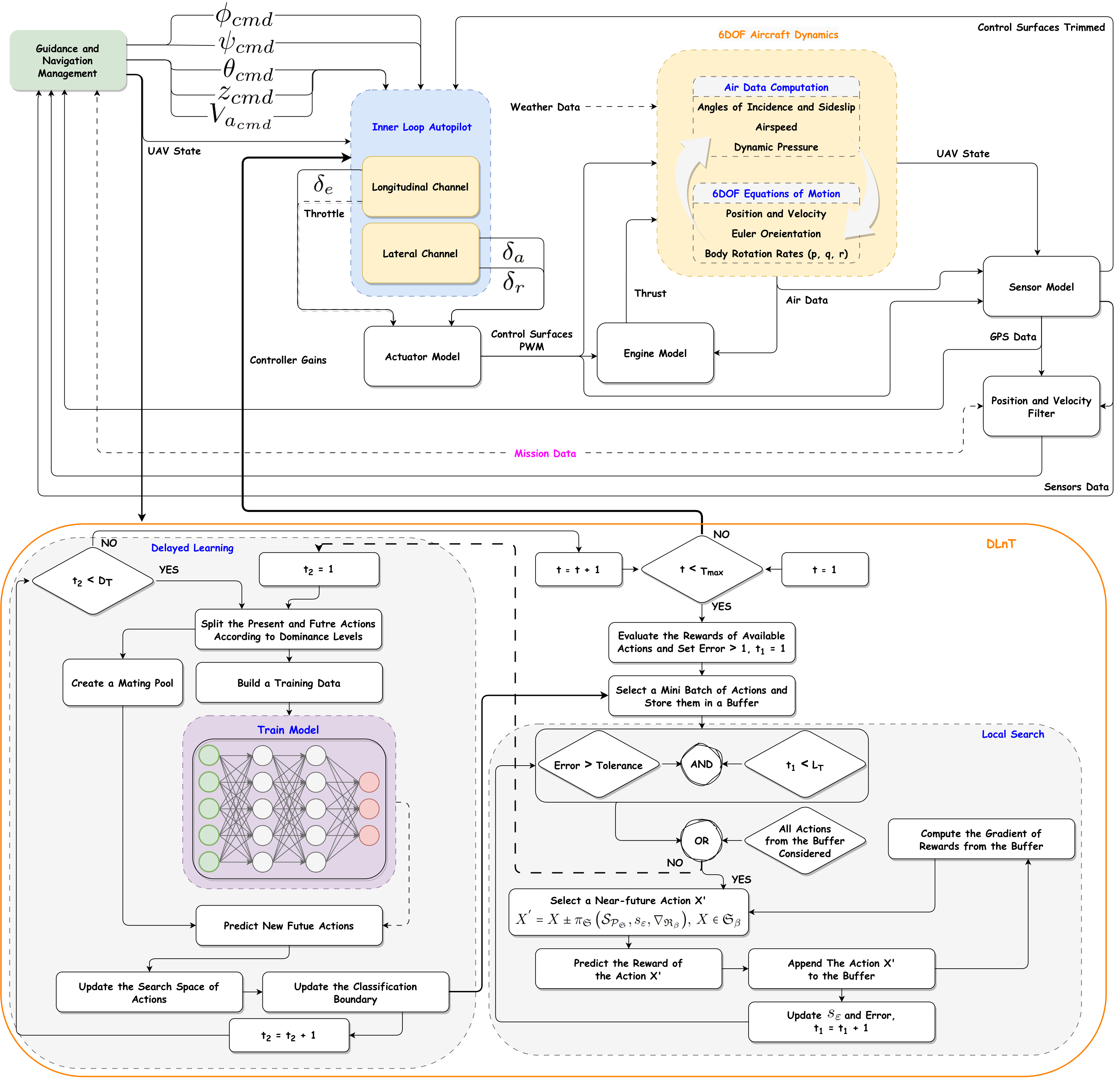}
	\caption{ Step-wise overview of the delayed learning and tuning (DLnT) algorithm  }
	\label{fig_DLnT}
\end{figure*}

The controller's gains are periodically (within time horizons) tuned using the proposed DLnT (Delayed Learning and Tuning) algorithm. As illustrated in Fig. \ref{fig_DLnT} and upon supplying the algorithm with the actual state of a UAV, the first step consists in delimiting the scope of the search space ${\mathcal{S}_{\mathcal{P}_\mathfrak{S}}}$ where the values of the controllers' gains should be set. After kicking-off with an initial set of actions each representing a possible combination of gains values, each action is evaluated on the basis of some reward scores. The goal is to minimize the tracking errors ($e_{\phi}(t)$, $e_{\theta}(t)$, $e_{\psi}(t)$, $e_{z}(t)$, $e_{{{_{\scaleto{V_a}{4pt}}}}}(t)$) between the commanded signals (Eq. \ref{eq_error}) and the actual longitudinal/ lateral responses in addition to the difference $e_{{{_{\scaleto{path}{4pt}}}}}(t)$ between the actual path pursued and the reference one.

\begin{equation}
	\small
	\lim\limits_{t \to \mathcal{H_{{_{\scaleto{\mathcal{T}}{3pt}}}}}}  \begin{Vmatrix}
		e_{\phi}(t) \approx\; {\phi_{{\textit{cmd}}}}(t) - {\phi}(t) \\
		e_{\theta}(t) \approx\; {\theta_{{\textit{cmd}}}}(t) - {\theta}(t) \\
		e_{\psi}(t) \approx\; {\psi_{{\textit{cmd}}}}(t) - {\psi}(t) \\
		e_{z}(t) \approx\; {z_{{\textit{cmd}}}}(t) - {z}(t) \\
		e_{{{_{\scaleto{V_a}{4pt}}}}}(t) \approx\; {V_{a_{\textit{cmd}}}}(t) - {V_a}(t)
	\end{Vmatrix}  = 0
	\label{eq_error}
\end{equation}\\

Next, an initially empty buffer $\mathfrak{S}_\beta$ will be filled with a small set of elite actions whose reward scores are not dominated by any other action. This selection is achieved with respect to a reference labeled as a classification boundary. In the subsequent steps as long as the ending criteria indicated in Fig. \ref{fig_DLnT} are not met, the algorithms enters a local search phase. The objective is to improve the the quality of the buffer, which in turn would help increasing the speed of learning and allow to make the most of the past experiences.\\
By observing the reward scores stored in the buffer, the gradient $\nabla_{\mathfrak{R}_\beta}$ can be computed. It will be used alongside an adaptive step size $s_\varepsilon$ to create new possible actions within the proximity of each action from the buffer for a finite time frame $L_T$. These new actions can be mapped with the previous one using a random walk function $\pi_{\mathfrak{S}}$ as shown in Fig. \ref{fig_DLnT}.\\

Once predicting the reward scores of the new actions and storing them within the buffer is completed, the DLnT algorithm switches to the delayed learning phase. The agent in this phase makes use of the buffer and learns from its history. The objective of invoking this phase is to keep the classification boundary up-to-date with the evolution of the search space and the aftermaths of carried-out actions.\\
The process begins by dividing the buffer into two balanced subsets each containing the Pareto non-dominated and the dominated actions with respect to the current classification boundary. The two subsets are used to build a training dataset  $\{\langle \mathfrak{C}_i, \mathfrak{C}_j \rangle, \mathfrak{L}_{{_{\scaleto{\langle \mathfrak{C}_i, \mathfrak{C}_j \rangle}{7pt}}}}\}$ with $\mathfrak{C}_i, \mathfrak{C}_j$ being the rewards scores of two actions from the buffer. A classifier with three label categories is adopted for the training process:

\begin{equation}
	\small
	\mathfrak{L}_{{_{\scaleto{\langle \mathfrak{C}_i, \mathfrak{C}_j \rangle}{7pt}}}} = \begin{cases}
		-\infty  & \text{if} \; \texttt{all}(\mathfrak{C}_i <= \mathfrak{C}_j) \;  \land \; \texttt{any}(\mathfrak{C}_i < \mathfrak{C}_j) \\
		0 &  (\text{if} \; \texttt{any}(\mathfrak{C}_i <= \mathfrak{C}_j) \;  \land \; \texttt{any}(\mathfrak{C}_i < \mathfrak{C}_j))\; \lor \; (\texttt{all}(\mathfrak{C}_i == \mathfrak{C}_j)) \\
		+\infty  &  \text{if} \; \neg\texttt{all}(\mathfrak{C}_i <= \mathfrak{C}_j) \;  \land \; \neg\texttt{any}(\mathfrak{C}_i < \mathfrak{C}_j)
	\end{cases}
\end{equation}

A feedforward neural network (FNN) is then used for to recognize and learn the dominance relationship between the pairs of actions. The goal is to select random pairs from the buffer to anticipate new offspring actions from which the feasible and favorable ones will be stored according to their predicted dominance relationships.\\
The learning delay should be set in such a way to avoid quickly saturating buffer and losing track of the promising solutions. Hence, frequently updating classification boundary each $L_T+D_T$ periods grantees to minimize the variance and improve the convergence towards the optimal action yielding the finest tuned gains for the controller.

\section{Results and analysis}
\label{sec5}

All the analyses and comparative evaluations reported in this section are carried out using SIMULINK and MATLAB 2025a on an Intel(R) Core(TM) i9-13900H PC with a 64 GB of RAM.\\
The region-of-interest is a flat area of $6\times6$ Km$^2$ above which there would be present a number of thermals having lifecycles that vary from $[6,10]$ minutes, radii and strengths respectively within the ranges $[50, 100]$ m and $[0.1, 4]$ m/s and also a number of targets each with a duration ranging from $[6,30]$ minutes. The operational limits of altitude are: $\mathcal{Z}_{\min} = 200, \mathcal{Z}_{\max}=1000 $ m. The deployed gliders are of model "phoenix 2400" with 50 Wh as initial battery capacities for a mission lasting $T_{\mathcal{M}} = 6$ hours.\\
For the comparative analysis, three baselines are chosen: 1)  a semi-cooperative baseline, in which the deployed UAVs only share the bare minimum information, that is the location of where lift has been encountered. 2) a non-cooperative baseline, in which each UAV has its own soaring maps and moves independently of other UAVs. 3) a zero-knowledge baseline, where the UAVs do not have access to any soaring map and they rely on the chance to seek lift. Similarly, 15 multi-objective heuristics (see Table \ref{tab_benchmarking}) and 4 multi-agent variants of RRT algorithm (Lazy-RRT, RRT-connect, Extended-RRT, RRT$^\thinstar$) are chosen to validate the proposed path planning and tracking strategy:

\begin{table*}[h!]
	\centering \tiny
	\caption{  List of the selected heuristics for the comparative analyses }
	\begin{tabular}{ c  c   c  c    p{2.5cm}  p{2.5cm} }
		\hline
		Abbreviation& Full heuristic name   &    Abbreviation & Full heuristic name   \\
		\hline
		\textit{FLEA}& Fast sampling based evolutionary algorithm   &    \textit{MOPSO} & Multi-objective particle swarm optimization   \\
		\textit{MODA} & Multi-objective dragonfly algorithm  &     \textit{MSFLA} & Multi-objective shuffled frog leaping algorithm   \\
		\textit{MOEA/D} & Evolutionary algorithm based on decomposition &     \textit{NSGA-III} & Non-dominated sorting genetic algorithm III	\\
		\textit{MOFPA} & Multi-objective flower pollination algorithm  &     \textit{PREA} & promising-region based evolutionary algorithm  \\
		\textit{MOGOA} & Multi-objective grasshopper optimization &     \textit{REMO} & Expensive optimization by relation learning and prediction  \\
		\textit{MOGWO} & Multi-objective grey wolf optimization  &     \textit{RSEA} & Radial space division based evolutionary algorithm  \\
		\textit{MOLSA} &  Multi-objective lightning search algorithm &     \textit{RVEA} & Reference vector guided evolutionary algorithm  \\
		&  &   \textit{SMEA} & Self-organizing multi-objective evolutionary algorithm  \\ 
		
		\hline

	\end{tabular}
	\label{tab_benchmarking}
\end{table*}

The analyses are then regrouped into six categories as discussed in the next subsections:

\subsection{Soaring flight and duration analysis}

\begin{figure}[!h]
	\centering
	\hspace*{-0.2in}
	\includegraphics[scale=0.33]{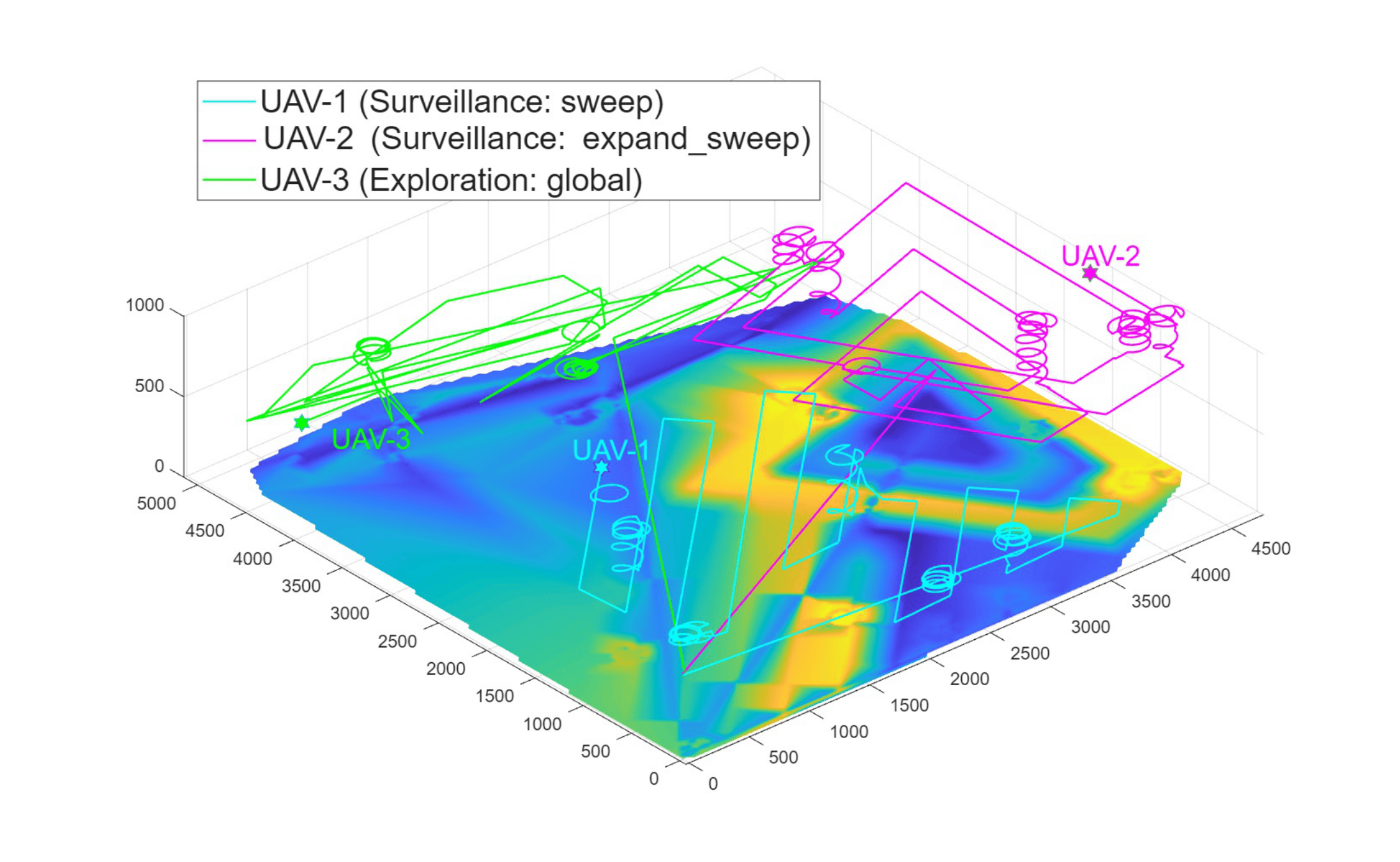}
	\caption{ 3D view of flight paths for three UAVs after 1 hour }
	\label{fig_flight_path}
\end{figure}

The first obvious metric for evaluating the endurance of UAVs is to examine how long they have flown and ideally without resorting to use their engines.\\
Fig. \ref{fig_flight_path} represents the 3D trajectories that three out of four UAVs followed once the mission had started (the fourth one was omitted for clarity). It is clear that the UAVs coordinated themselves and each one was assigned to a sub-area where it operated temporarily. As indicated by the heat map, two UAVs are performing exploration and the remaining two (UAV 1 and UAV 2) are sweeping their sub-areas and have scanned a great part of them. It can also be perceived that the UAVs have exploited and rejected some updrafts on their ways.\\

Table \ref{tab_flight_time} highlights two important phases (unpowered and powered climbs) in the soaring flight for a number of UAVs during the whole mission. Two values are given in brackets. The first one represents the total flight time the UAVs took while exploiting the updrafts and the second one is the total flight time with engine on. \\
Table \ref{tab_flight_time} further shows two variants of the proposed approach. The one with label "(\textit{RoI}$_\rho=1$)" means that the UAVs were free to divide the RoI into sub-areas whereas for the one with label "(\textit{RoI}$_\rho=0$)", they were forced by the GCS to operate on the mission area as it is.

It can be observed that all the UAVs in two variants have spent more than 17.8\% of the mission time gaining altitude by exploiting updrafts and almost only 1.7\% of the mission time using their engines for powered climbs. This clearly implies that the UAVs have been gliding without consuming their batteries for much a longer period.\\ Moreover, the variant labeled "(\textit{RoI}$_\rho=0$)" is slightly better than the other variant in terms of unpowered climbs, which is due to being on the same area, alleviates the restriction on staying on separate sub-areas until competing their sub-missions or urgently needing a lift and so, the UAVs have equitable chances to exploit the same number of lift sources known to all of them.\\
On the other hand, the benchmarking baselines fell behind as the UAVs take only 1.3\% to almost 9.7\% of the mission time for unpowered climbs and plainly more time using their engines (up to 5\%).

\begin{table*}[h!]
	\centering \tiny
	\caption{ Flight durations for unpowered and powered climbs (minutes)    }
	\begin{center}
		\begin{tabular}{c  c  c  c  c  c  c   p{2.5cm}   p{2.5cm}   p{2.5cm}  p{2.5cm}  p{2.5cm}   p{2.5cm}  p{2.5cm}   } 
			\hline
			\multicolumn{2}{c}{\backslashbox{\textit{Number of UAVs}}{\textit{Algorithms}}} &  \textit{Proposed (\textit{RoI}$_\rho=1$)}   &   \textit{Proposed (\textit{RoI}$_\rho=0$)}  &  \textit{Semi-cooperative}   &   \textit{Non-cooperative} &  \textit{Zero-knowledge}   \\
			
			\hline
			
			\multirow{2}{*}{ 2 UAVs }  &  UAV ID--1:  & (75.23, 5.3333) &(83.8026, 1.4333) &(11.5205, 16.5013) &(13.4337, 15.7667) &(41.2399, 12.9) \\
			&   UAV ID--2:  &  (65.5037, 5.0167) &(81.4016, 3.6) &(4.8384, 17.9466) &(29.1326, 12.799) &(19.2954, 14.3333) \\
			
			\hline
			
			\multirow{3}{*}{ 3 UAVs }  &  UAV ID--1:  & (39.9614, 8.4833) &(66.3691, 6.8) &(22.9901, 13.6273) &(27.7015, 14.3333) &(33.1109, 13.7289) \\
			&   UAV ID--2:  & (60.2945, 9.5667) &(91.3199, 4.2) &(28.1223, 12.2088) &(17.7937, 13.6167) &(32.6115, 13.4) \\
			&  UAV ID--3:  &  (72.7052, 8.6) &(71.8077, 4.85) &(24.8437, 12.9468) &(18.854, 14.35) &(8.8162, 16.5) \\
			
			\hline
			
			\multirow{4}{*}{ 4 UAVs }  &  UAV ID--1:  & (55.5027, 6.45) &(48.4362, 10.75) &(24.5505, 14.3333) &(13.7897, 16.4921) &(33.618, 10.75) \\
			&   UAV ID--2:  & (76.5592, 5.7333) &(76.338, 5.1333) &(15.3624, 15.7705) &(27.77, 12.1833) &(8.3927, 16.4867) \\
			&  UAV ID--3:  & (36.8315, 4.55) &(79.4741, 6.9) &(30.6892, 13.6385) &(16.4958, 15.0732) &(5.445, 17.2167) \\
			&  UAV ID--4:  & (55.4728, 9.0167) &(58.7267, 7.1667) &(16.9092, 17.2183) &(15.4395, 15.7667) &(35.2471, 12.2082) \\
			
			\hline
			
			\multirow{5}{*}{ 5 UAVs }  &  UAV ID--1:  & (69.927, 5.9167) &(60.3152, 5.7333) &(31.1555, 11.9913) &(29.5094, 12.7368) &(46.974, 10.0333) \\
			&   UAV ID--2:  &  (48.8539, 9.2833) &(80.4297, 4.4333) &(30.7216, 12.25) &(17.0799, 15.8059) &(15.4412, 15.7926) \\
			&  UAV ID--3:  &  (66.7622, 6.9333) &(64.5631, 7.8833) &(18.8351, 13.3833) &(27.3749, 13.639) &(11.3873, 17.202) \\
			&  UAV ID--4:  &  (62.346, 6.5833) &(70.9288, 7.8833) &(13.7066, 16.4871) &(21.0531, 13.6667) &(14.9917, 15.4297) \\
			&  UAV ID--5:  & (81.5028, 7.2) &(75.1384, 10.5167) &(24.4104, 15.2428) &(29.0173, 13.6167) &(36.1567, 12.15) \\
			
			\hline
			
			\multirow{6}{*}{ 6 UAVs }  &  UAV ID--1:  &  (73.4514, 6.4167) &(50.5939, 1.45) &(35.8492, 11.2143) &(16.5962, 17.2) &(28.2213, 13.6284) \\
			&   UAV ID--2:  &  (50.1825, 8.7539) &(75.8158, 7.9) &(15.9558, 16.5) &(30.403, 12.9) &(34.5392, 12.9043) \\
			&  UAV ID--3:  &  (73.5229, 5.5) &(64.3053, 5.25) &(22.7449, 16.4938) &(18.781, 12.9) &(13.7938, 16.489) \\
			&  UAV ID--4:  & (71.5497, 4.5333) &(49.0779, 8.7) &(22.9552, 14.0167) &(10.9675, 16.7629) &(22.1606, 15.1901) \\
			&  UAV ID--5:  &  (66.5851, 4.1833) &(59.8643, 5.8667) &(25.508, 14.3333) &(14.7554, 17.2261) &(21.8871, 15.1119) \\
			&  UAV ID--6:  & (58.0371, 6.45) &(69.6138, 9.4515) &(38.9174, 10.75) &(24.5051, 13.0441) &(24.4051, 13.6301) \\
			
			\hline
			
		\end{tabular}
		\label{tab_flight_time}
	\end{center}
\end{table*}

\subsection{Updrafts detection and exploitation analysis}

\begin{figure}[!h]
	\centering
	\includegraphics[scale=0.3]{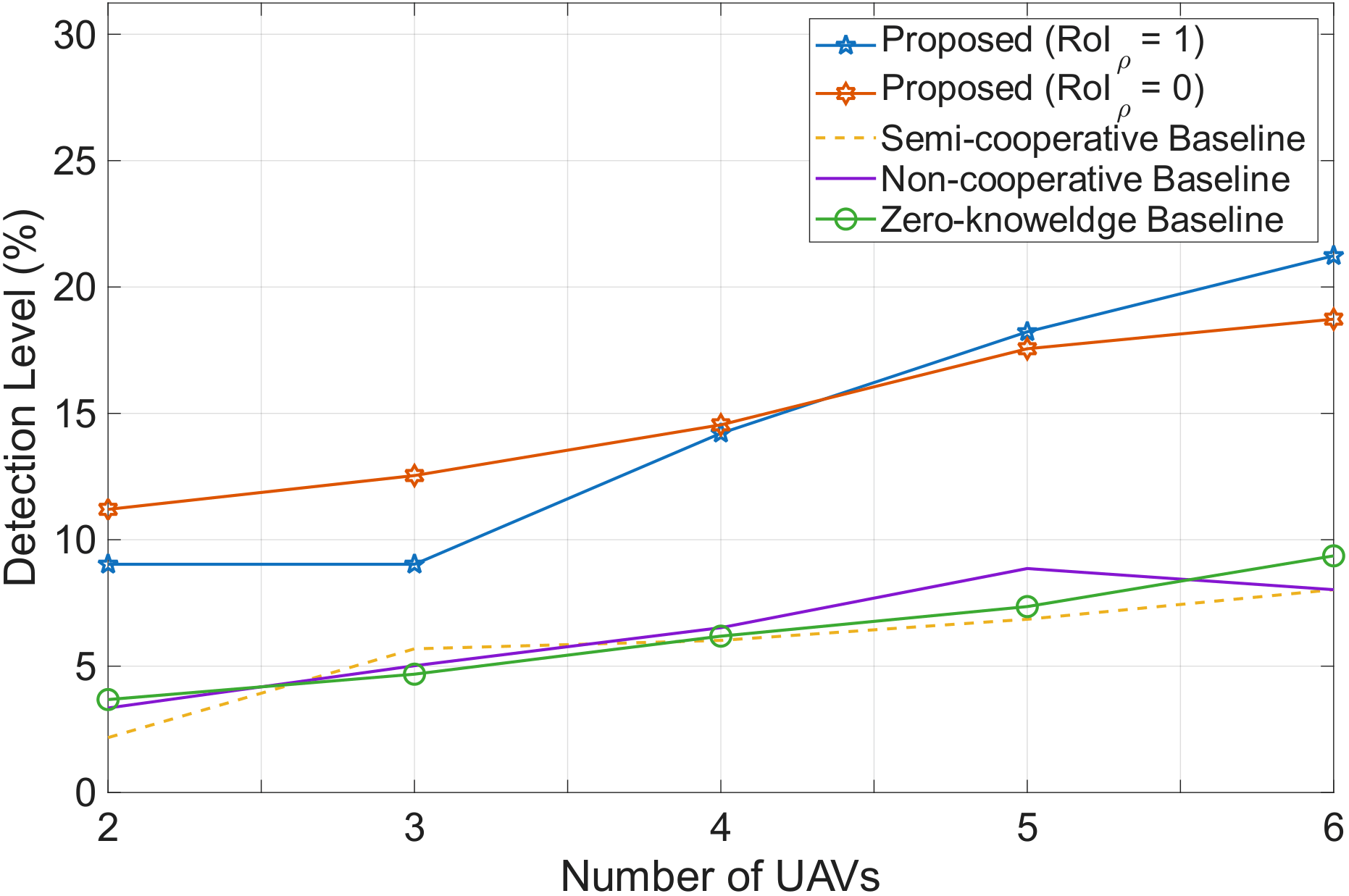}
	\caption{ Total level of newly detected and non-mapped updrafts }
	\label{fig_wnd_detect_all}
\end{figure}

\begin{figure*}[!h]
	\centering
\hspace*{-0.7in}
	\subcaptionbox{\label{fig_wnd_detect_5u}}{\includegraphics[scale=0.3]{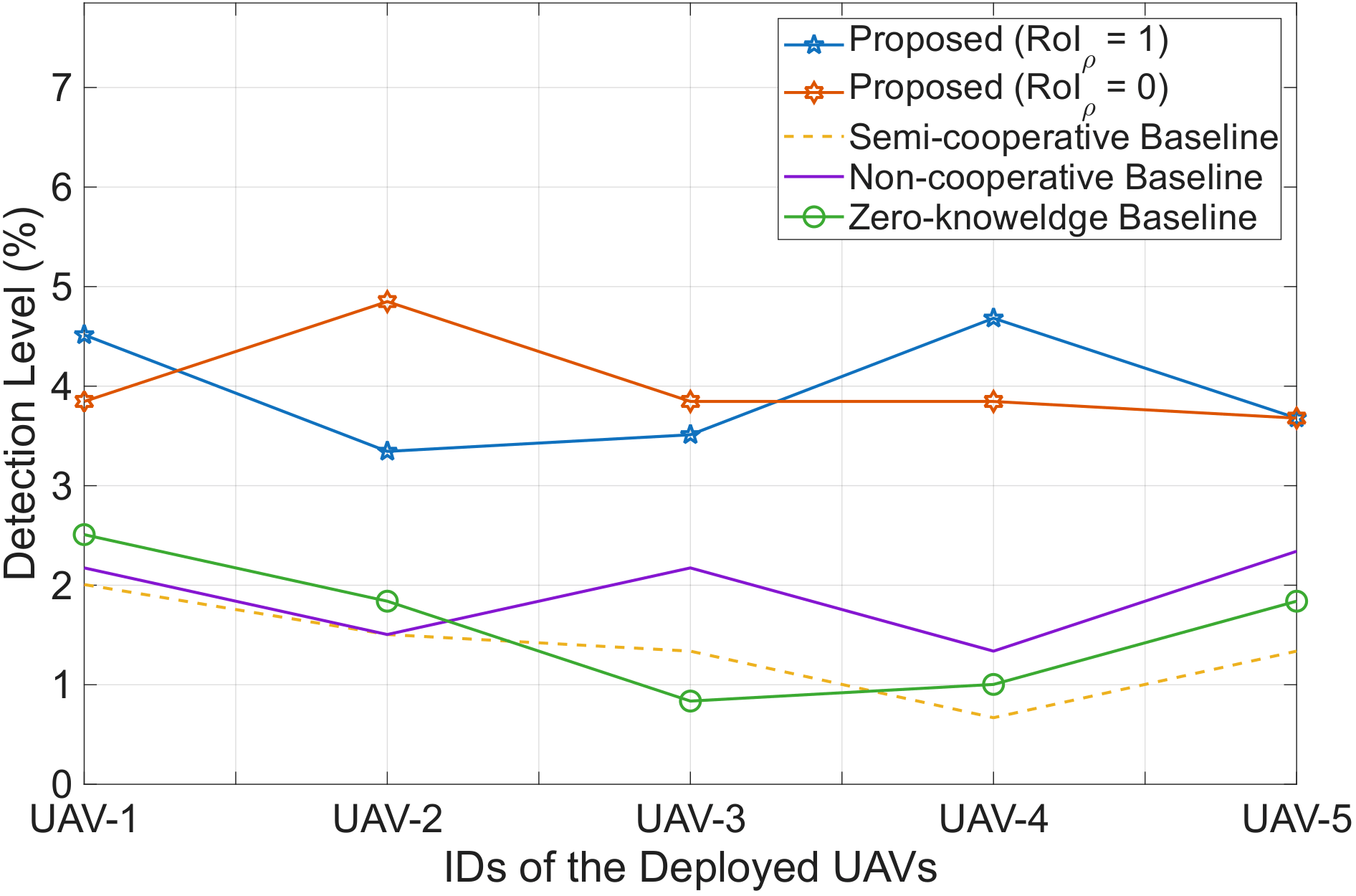}}
	\subcaptionbox{\label{fig_wnd_detect_6u}}{\includegraphics[scale=0.3]{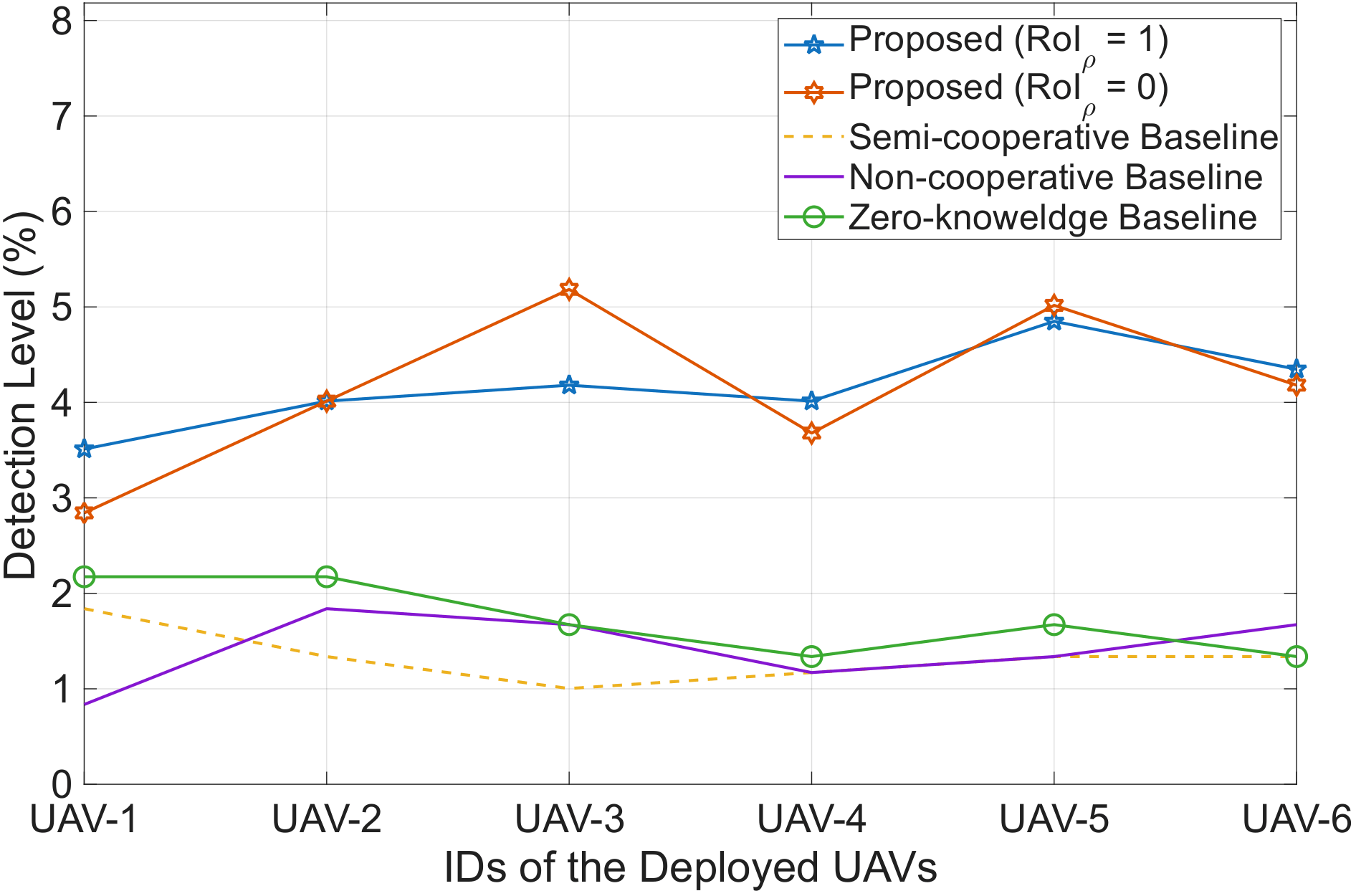}}
	\caption{  Contribution of each UAV to the detection of new updrafts: (a) use case of 5 UAVs, (b) use case of 6 UAVs  }
	\label{fig_wnd_detect_u}
\end{figure*}

The purpose of this subsection is to provide numerical proofs on the efficiency of the proposed cooperative soaring strategy in terms of making the most of updrafts to regain lost energy and stay aloft without landing.
In this context, Fig. \ref{fig_wnd_detect_all} portrays the total rate of detecting new updrafts with respect to the maximum number of updrafts that have been formed during the mission (a total number of 598 updrafts have been simulated through 6 hours).\\
As can be seen, the proposed approach is noticeably outperforming the benchmarking baselines. Fig. \ref{fig_wnd_detect_u} for its part, it additionally goes deeper to precise the exact number of detected updrafts by each UAVs when deploying 5 and 6 UAVs.

\begin{figure}[!h]
	\centering
	\includegraphics[scale=0.3]{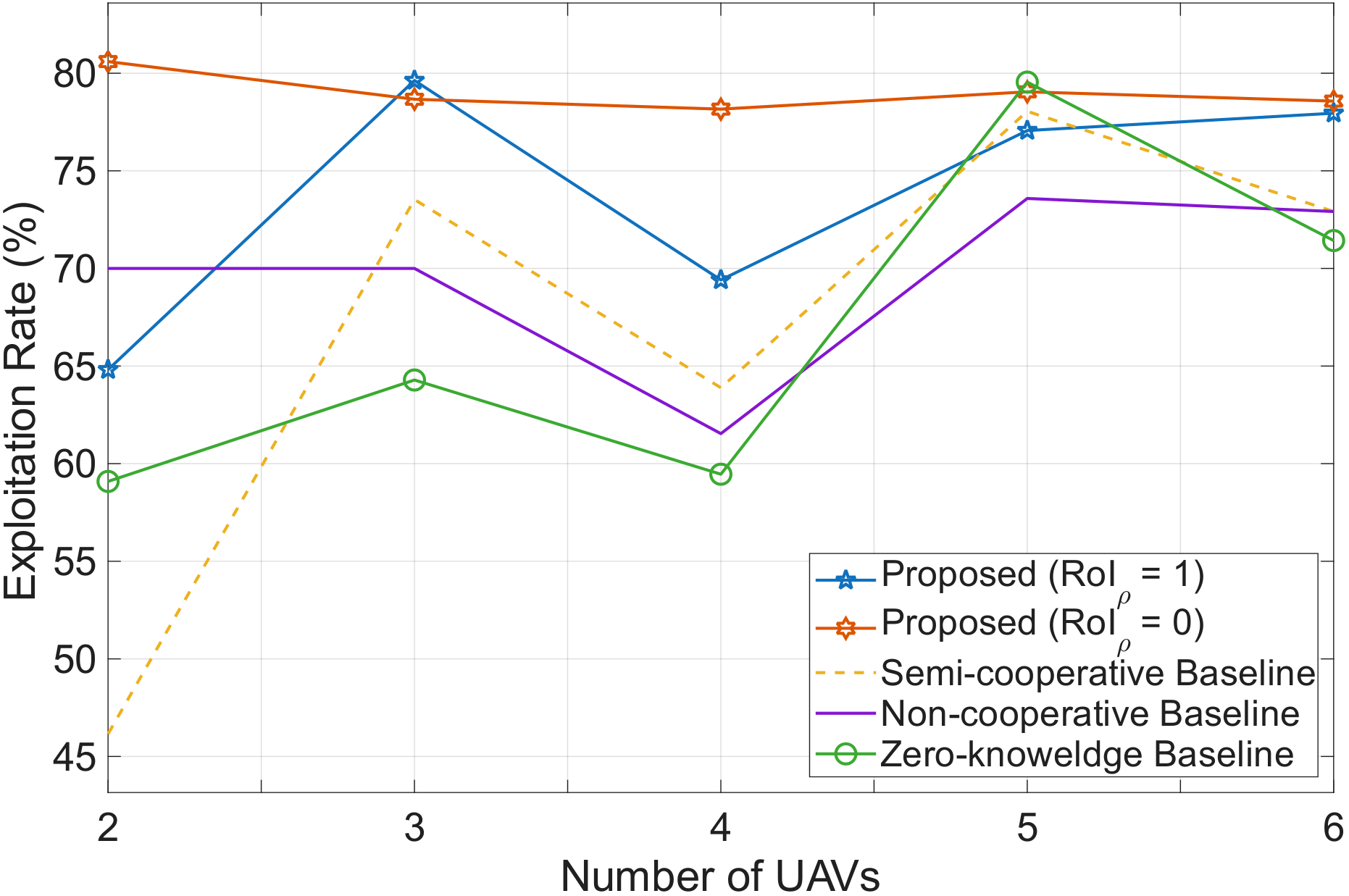}
	\caption{ Total rates of exploiting the detected and mapped updrafts  }
	\label{fig_wnd_exploit}
\end{figure}

Again, it can be observed that the proposed approach with the variant "(\textit{RoI}$_\rho=0$)" is slightly better that the variant "(\textit{RoI}$_\rho=1$)" for up to four UAVs. The benchmarking baselines could not compete with the proposed approach as there is always a difference of at least 10 updrafts in the best case.\\
Similarly, Fig. \ref{fig_wnd_exploit} illustrates the ratio of how many updrafts have been exploited with respect to what have been detected. A difference in appearance with that of Fig. \ref{fig_wnd_detect_all} is perceived and this is due to one of the following reasons:
\begin{itemize}
	\itemsep0em
	\item [-]  not all of the detected updrafts were exploitable because of their weaknesses (either being still in growing or in fading phases).
	\item [-] a detected updraft was exploited multiple times by the same UAV detecting it.
	\item [-] a UAV has exploited updrafts that were detected by the other cooperating UAVs.\\
\end{itemize}

Fig. \ref{fig_wnd_exploit} also shows that the semi-cooperative baseline in the case of 2 UAVs has the weakest aptitude in exploiting updrafts unlike the other baselines and this can be explained by a misuse of the shared lift map since the UAVs were restricted to exploit mapped instead of exploring more promising regions where strong updrafts may be present.

\begin{figure*}[!h]
	\centering
\hspace*{-0.7in}
	\subcaptionbox{\label{fig_wnd_num5}}{\includegraphics[scale=0.3]{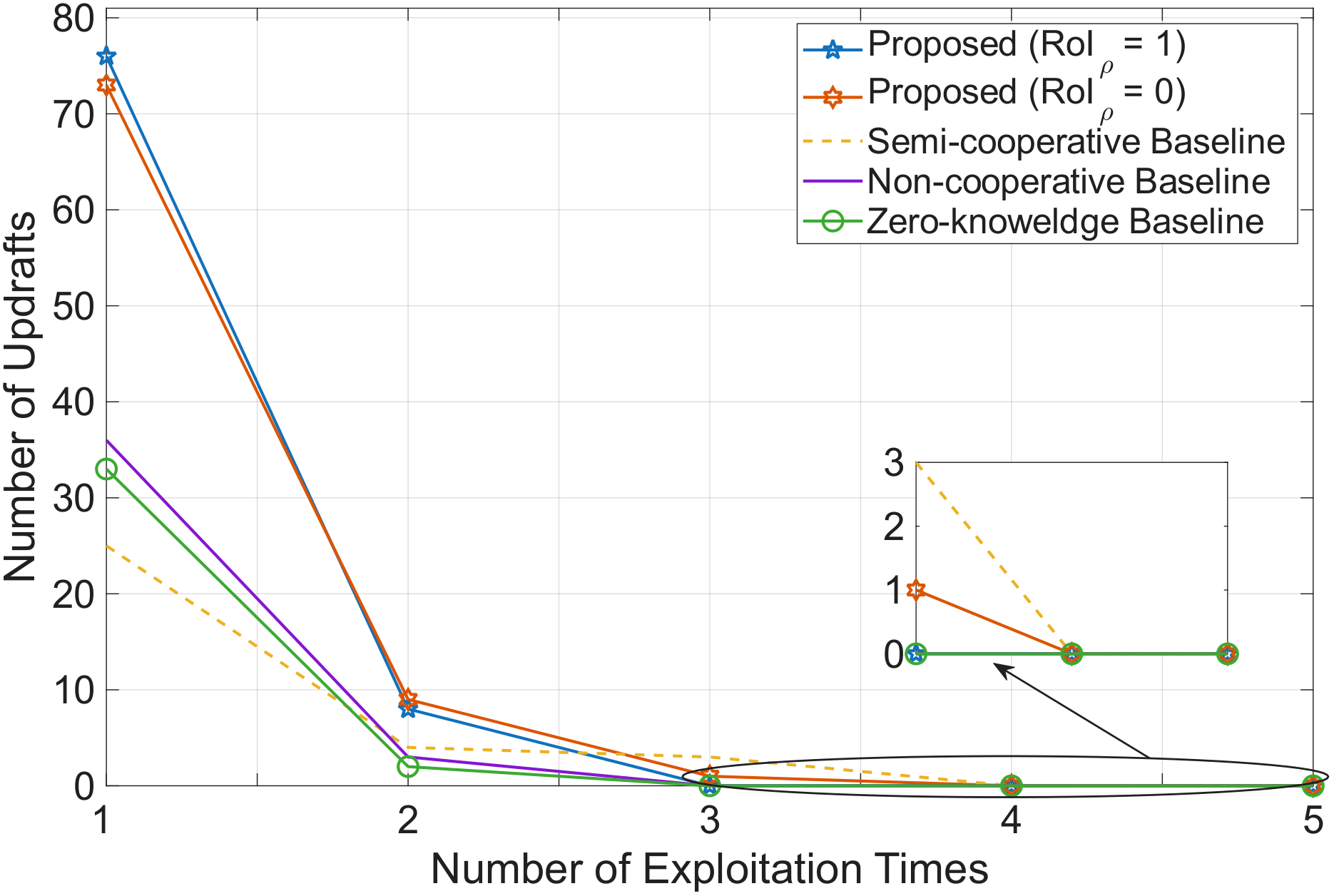}}
	\subcaptionbox{\label{fig_wnd_num6}}{\includegraphics[scale=0.3]{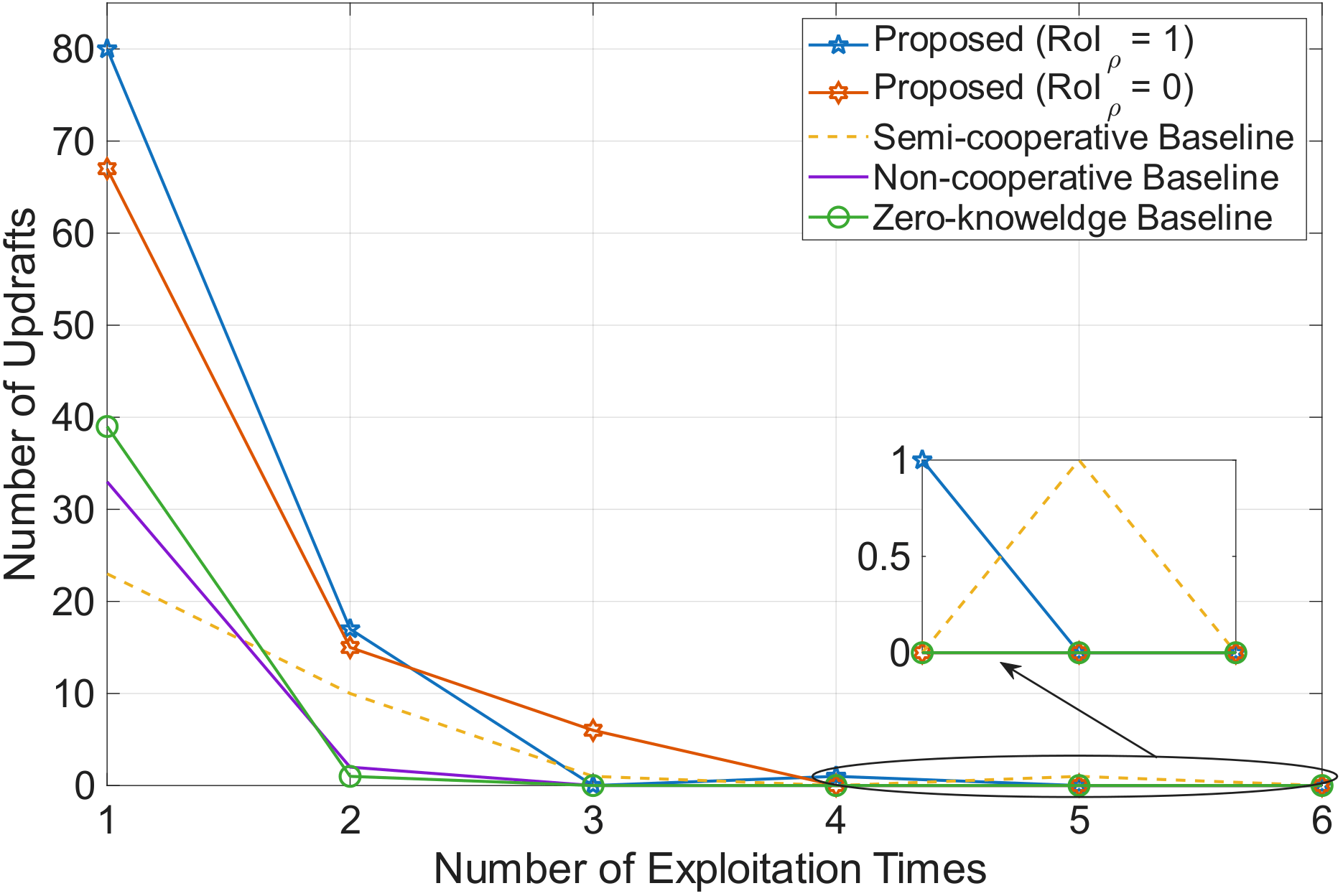}}
	\caption{  Number of exploitation occurrences of detected updrafts: (a) use case of 5 UAVs, (b) use case of 6 UAVs }
	\label{fig_wnd_num}
\end{figure*}

An auxiliary metric consists in evaluating the number of times an updraft has been exploited either by the same UAV or by another one is shown in Fig. \ref{fig_wnd_num}. The number of updrafts exploited once and twice in the proposed approach are respectively two times greater than those of the benchmarking baselines. Hence demonstrating the efficiency of the employed cooperative strategy, where the probability of exploiting an updraft many times increases with the number of deployed UAVs.\\
However, the zero-knowledge has slightly shown a better result in exploiting un updraft than that of the semi-cooperative baseline due to pure chance as the UAVs fly without keeping track of where lift has been previously encountered.

\subsection{Targets detection analysis}

\begin{figure}[!h]
	\centering
	\includegraphics[scale=0.3]{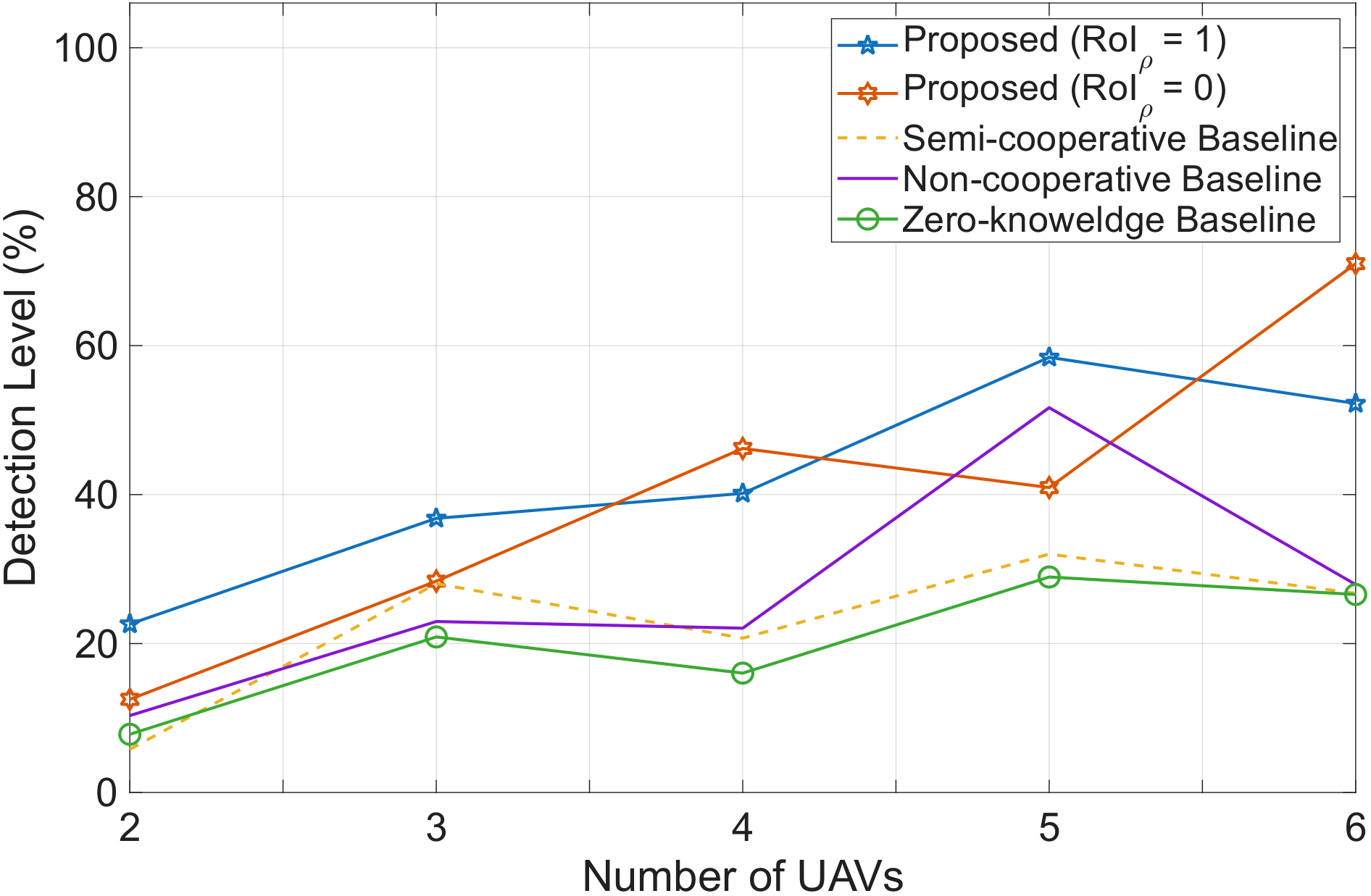}
	\caption{ Total level of detected non-duplicate targets }
	\label{fig_tar_detect_all}
\end{figure}

The principal objective of persistent surveillance is to monitor and collect data from ground targets. The more targets communicated to the GCS the more reliable the adopted strategy is. To this end, Fig. \ref{fig_tar_detect_all} depicts the total rate of detecting non-duplicate targets with respect to the maximum number of targets that have popped up during the mission (a total number of 1790 targets have been simulated through 6 hours).\\
Fig. \ref{fig_tar_detect_all} shows that the proposed variant "(\textit{RoI}$_\rho=1$)" has the upper hand with an augmenting slope as the number of UAVs grows.
This is due to fact of distributing the UAVs over different sub-areas increases the likelihood of observing more targets by diversifying the navigation methods between surveillance and exploration and reciprocally updating the mission map $\mathcal{M}_{\textit{mission}}$ (discussed in section \ref{sec4}). This will in turn bias the exploration waypoints towards more promising locations; bearing in mind that the targets have a short duration before disappearing.

\begin{figure*}[!h]
	\centering
\hspace*{-0.7in}
	\subcaptionbox{\label{fig_tar_detect_5u}}{\includegraphics[scale=0.3]{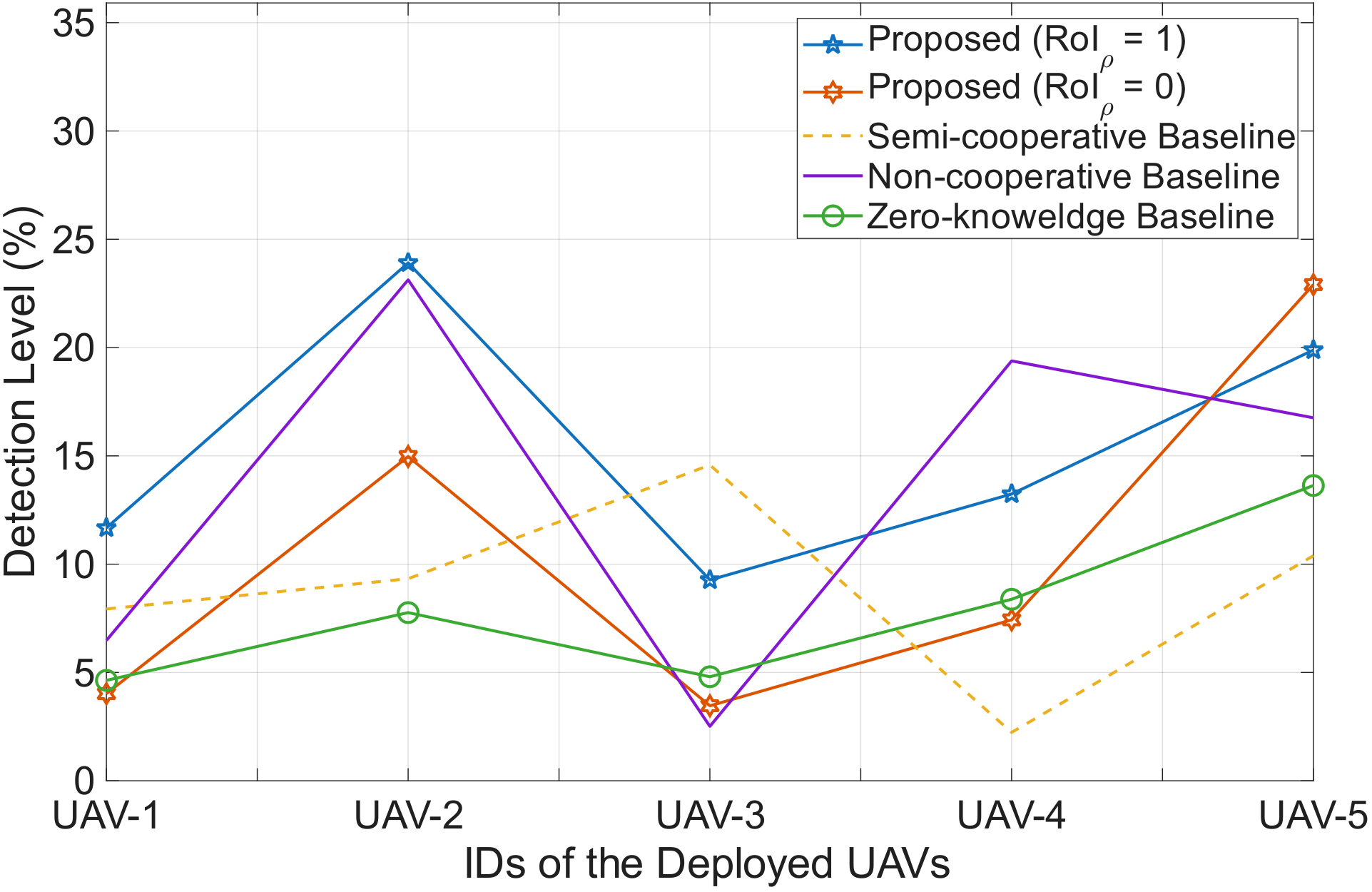}}
	\subcaptionbox{\label{fig_tar_detect_6u}}{\includegraphics[scale=0.3]{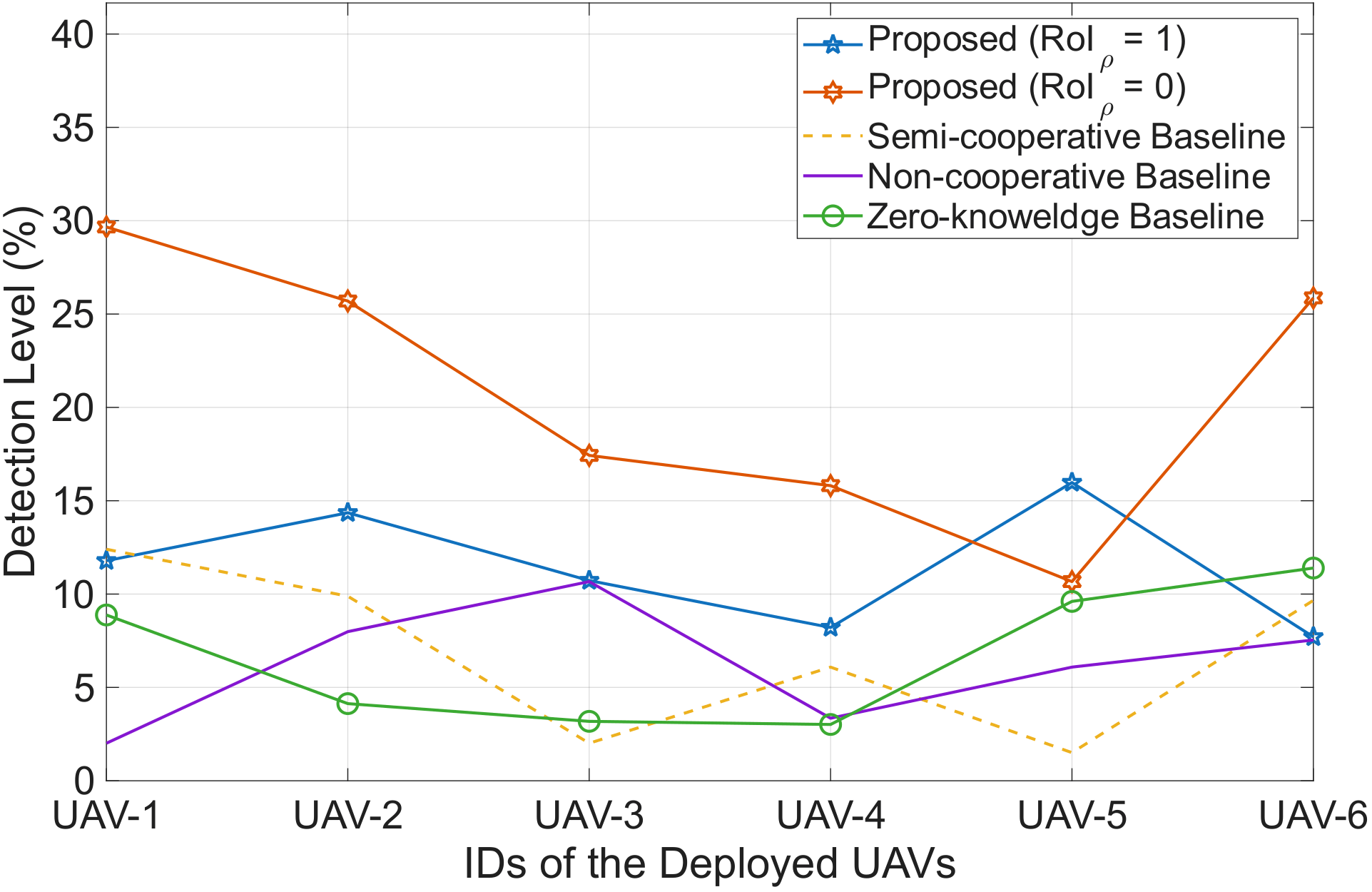}}
	\caption{  Contribution of each UAV to the detection of non-duplicate targets: (a) use case of 5 UAVs, (b) use case of 6 UAVs   }
	\label{fig_tar_detect_u}
\end{figure*}

Analogously to the previous subsection, Fig. \ref{fig_tar_detect_u} gives details on how many unique targets have been detected by each UAV.\\
The proposed variant "(\textit{RoI}$_\rho=0$)" exhibited in overall satisfying results for the case of 6 UAVs (Fig. \ref{fig_tar_detect_6u}) except for UAV 5, which was marginally outclassed.
However, in contrast to UAV 1 from the proposed variant "(\textit{RoI}$_\rho=1$)" having the highest rate scores, and UAV 1 of the non-cooperative baseline have not detected much. This can be explained by either:
\begin{itemize}
	\itemsep0em
	\item [-] no targets popped up in the regions where the UAVs were operating or have dissipated/moved somewhere else by the time of UAVs' arrival.
	\item [-] the UAVs were doing surveillance and due to the sweeping patterns, more duplicated copies of the same targets is observed many times and so the rate of unique ones was low.
\end{itemize}

\begin{figure*}[!h]
	\centering
	\hspace*{-0.7in}
	\subcaptionbox{\label{fig_tar_num5}}{\includegraphics[scale=0.3]{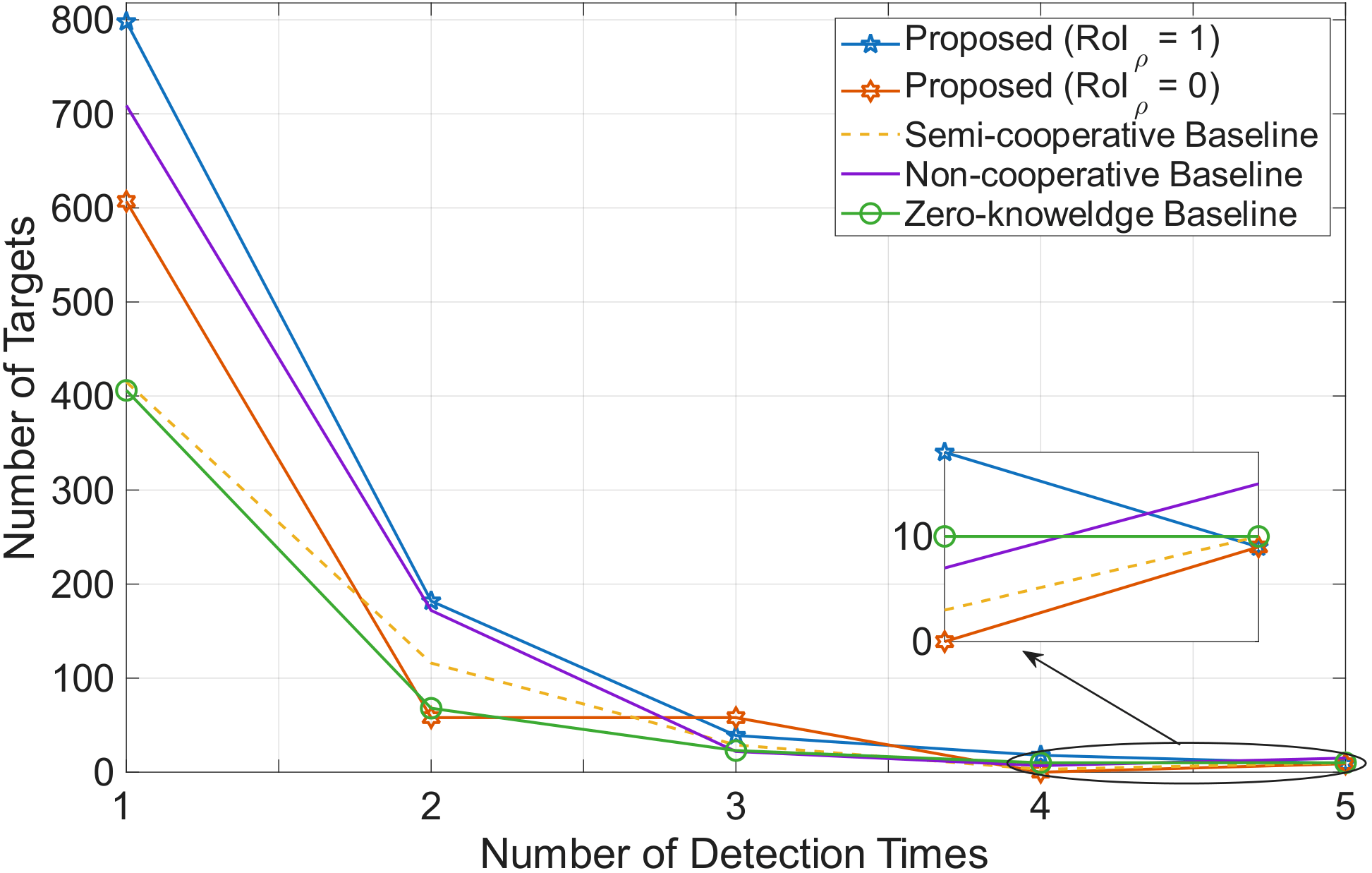}}
	\subcaptionbox{\label{fig_tar_num6}}{\includegraphics[scale=0.3]{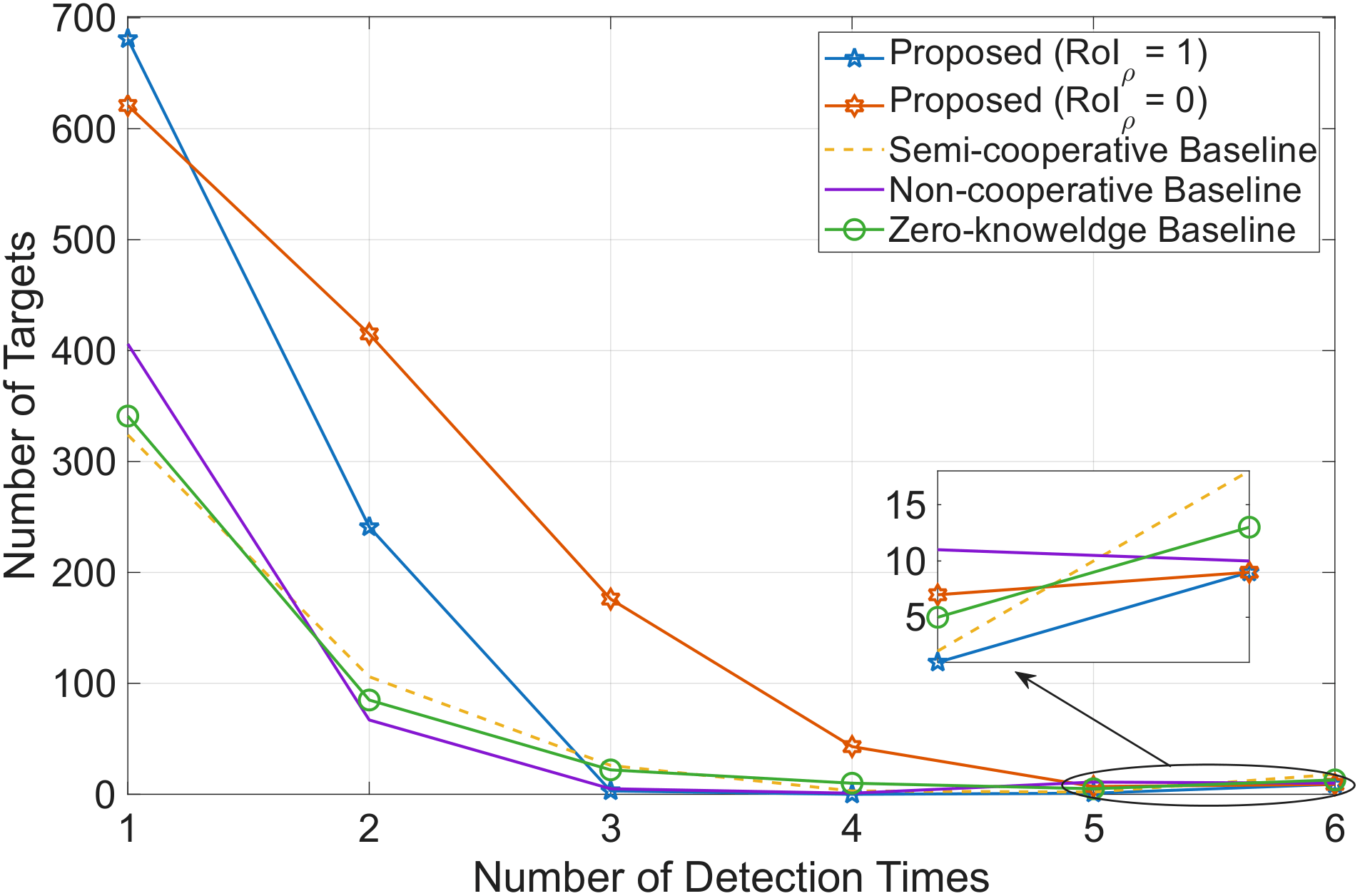}}
	\caption{ Number of targets detected by multiple UAVs: (a) use case of 5 UAVs, (b) use case of 6 UAVs   }
	\label{fig_tar_num}
\end{figure*}

Furthermore, Fig. \ref{fig_tar_num} displays how may targets have been caught in the coverage field-of-view of multiple UAVs. Once more, the proposed approach goes far beyond the benchmarking baselines even for unlikely and difficult case where more than 3 UAVs have monitored the same target (not in the same time due to the constraint of Eq. \ref{eq_const4}. Only the non-cooperative baseline was showing fair results in comparison with the remaining two baselines and this is because the UAVs are not restricted by cooperating rules and don't waste time sharing and updating a joint lift map.

\subsection{Power consumption analysis}

\begin{figure}[!h]
	\centering
	\includegraphics[scale=0.3]{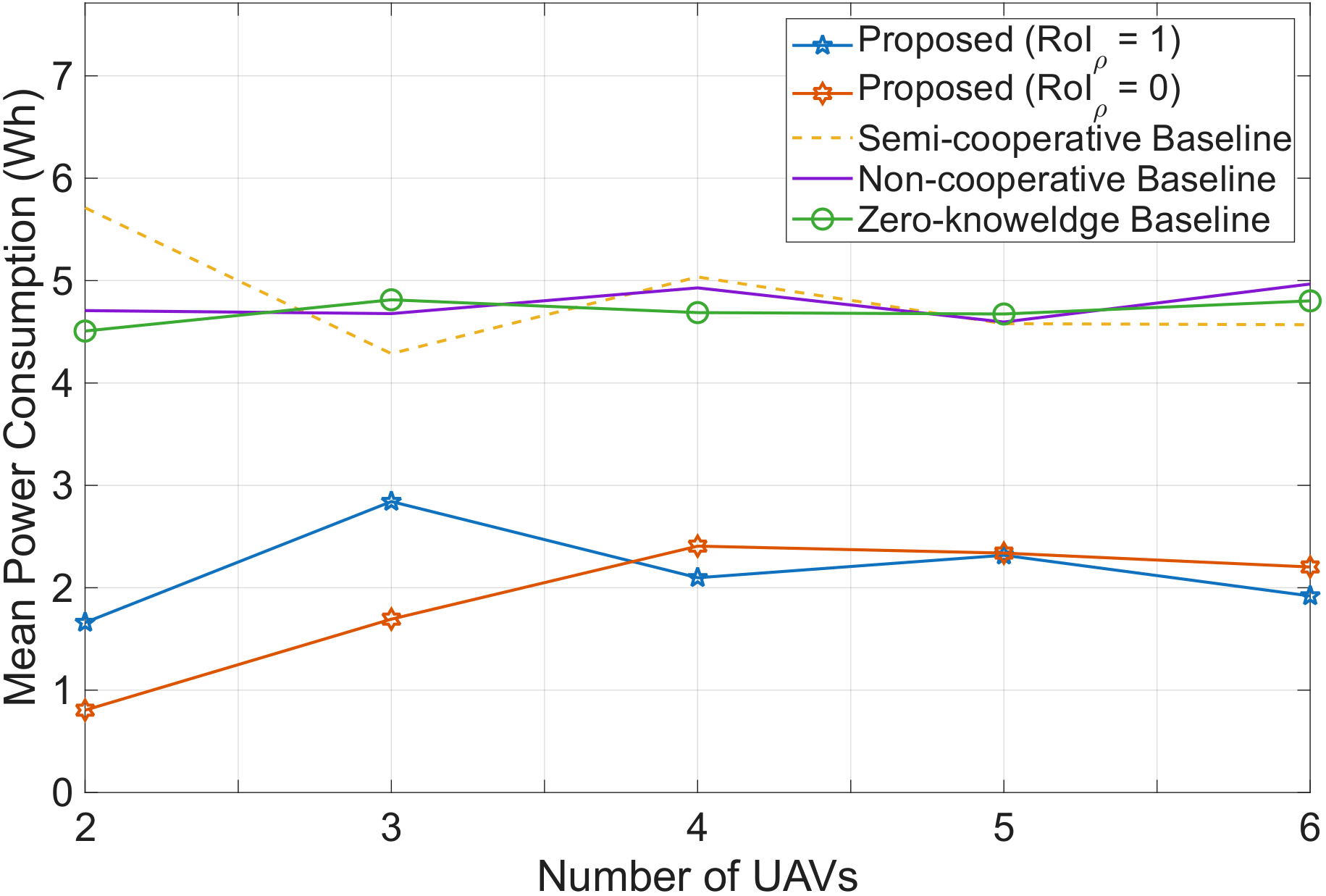}
	\caption{ Total mean power consumption during powered climbs }
	\label{fig_pwr_all}
\end{figure}

In this subsection, the levels of battery capacity at the end of the mission are reviewed.
Fig. \ref{fig_pwr_all} illustrates the average power consumption in Watt-hour of all deployed UAVs by the end of the mission. As can be remarked namely for the proposed approach and the semi-cooperative baseline, the more UAVs are utilized the less total power is consumed since the chances of exploiting more updrafts will be higher (as revealed in Fig. \ref{fig_wnd_exploit}).

\begin{figure*}[!h]
	\centering
	\hspace*{-0.8in}
	\subcaptionbox{\label{fig_batt_5u}}{\includegraphics[scale=0.3]{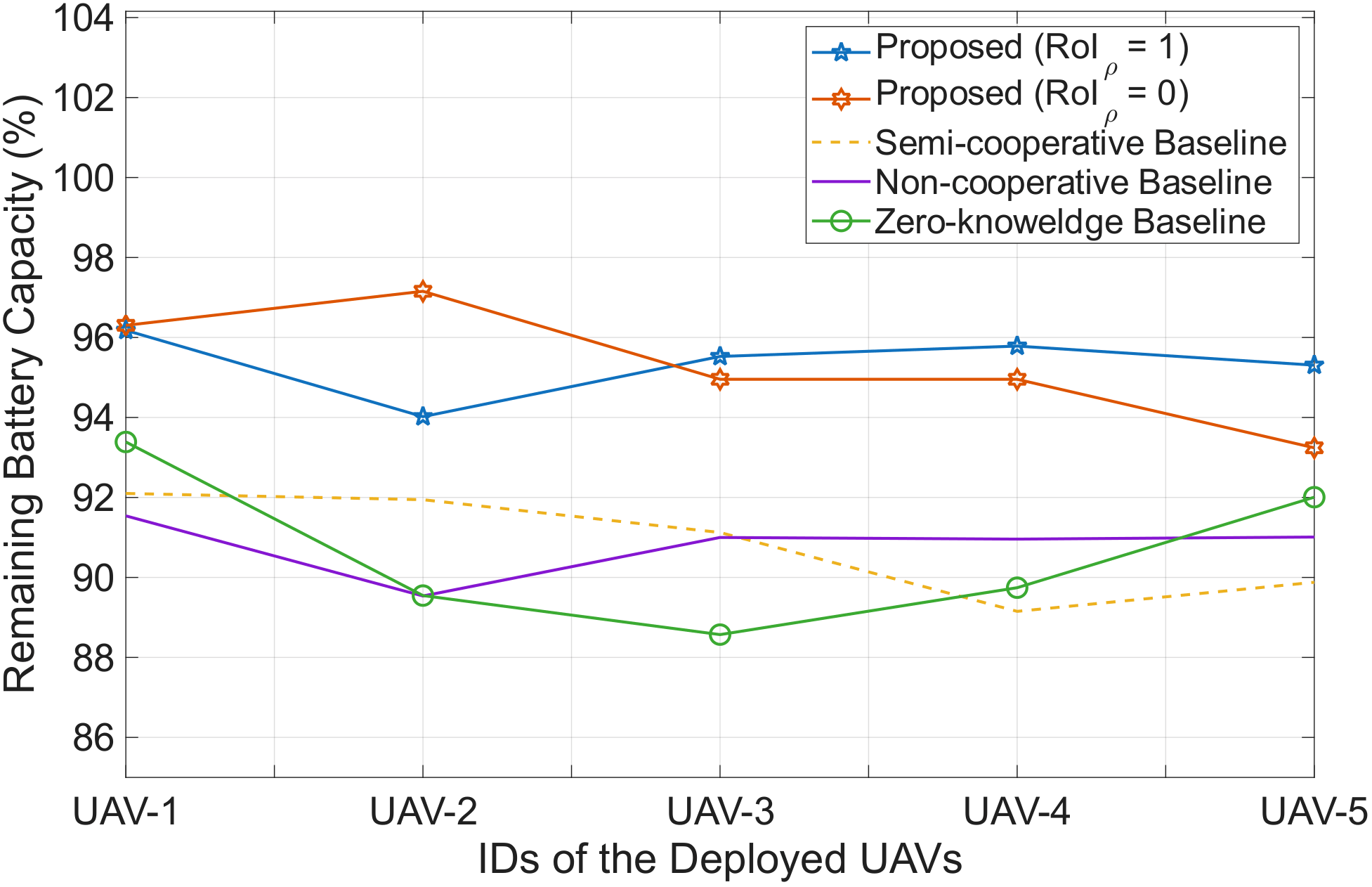}}
	\subcaptionbox{\label{fig_batt_6u}}{\includegraphics[scale=0.3]{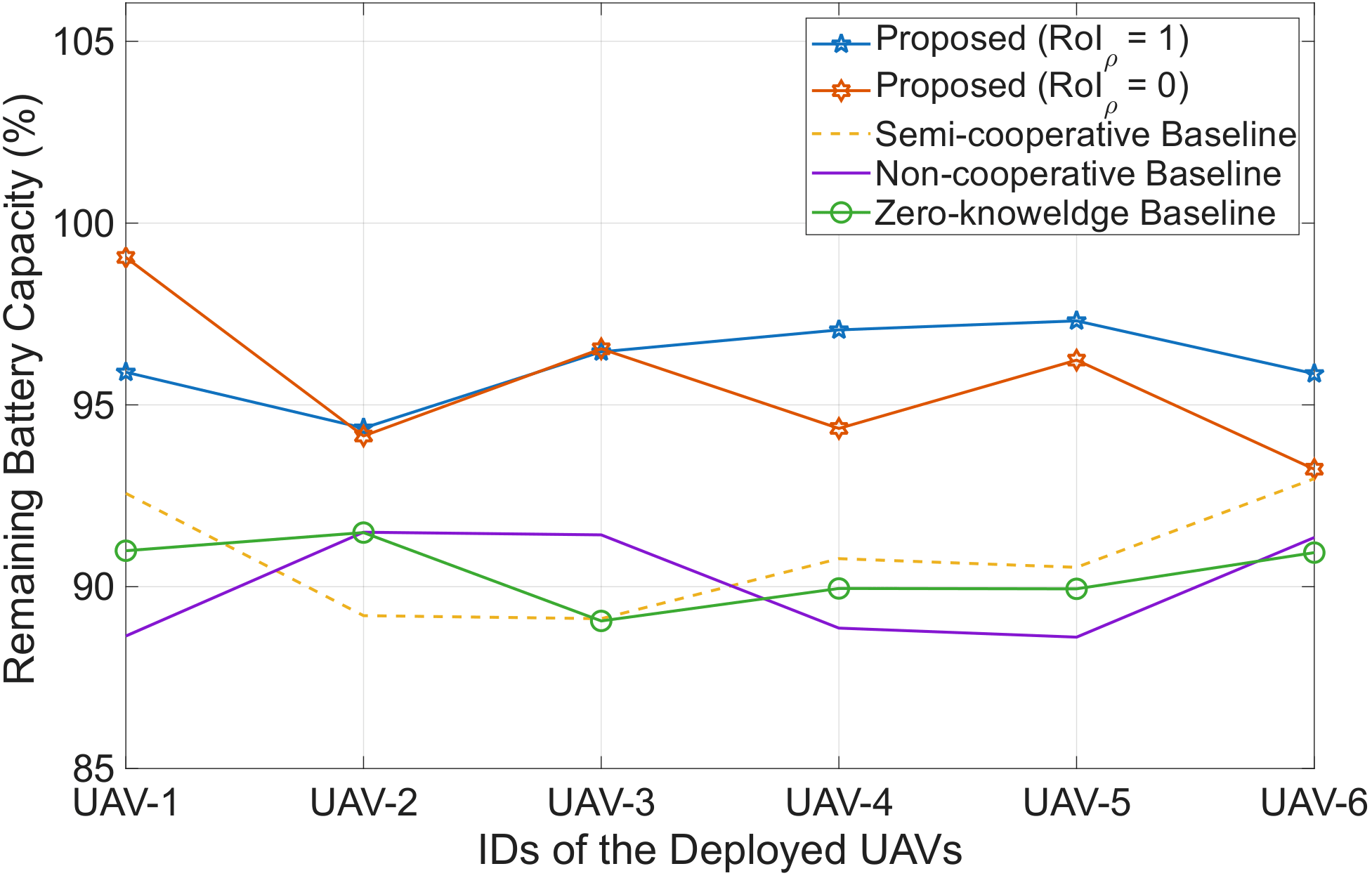}}
	\caption{  Mean percentage of remaining battery capacity for each UAV: (a) 5 UAVs, (b) 6 UAVs  }
	\label{fig_batt_u}
\end{figure*}

Diving into details, Fig. \ref{fig_batt_u} highlights the exact total amount of residual battery capacity per each UAV (the case of 4 UAVs).

The semi-cooperative baseline shows high remaining battery capacity for the majority of UAVs in comparison with the other baselines. However, it remains far bellow the results obtained by the proposed approach as can be observed from Fig. \ref{fig_batt_u}, which also shows that there exists UAVs whose battery capacity has almost not been consumed (e.g., UAV 1 in the proposed variant (\textit{RoI}$_\rho=0$) from Fig. \ref{fig_batt_6u}). This is because making use of a common lift map and sharing the locations of encountered updrafts does not guarantee that the UAVs will choose the optimal actions without more appropriate planning and coordination mechanism as implemented in the proposed approach.

\subsection{Path planning analysis}

\begin{figure*}
	\centering
	\hspace*{-0.3in}
	\subcaptionbox{\label{fig_path1}}{\includegraphics[scale=0.3]{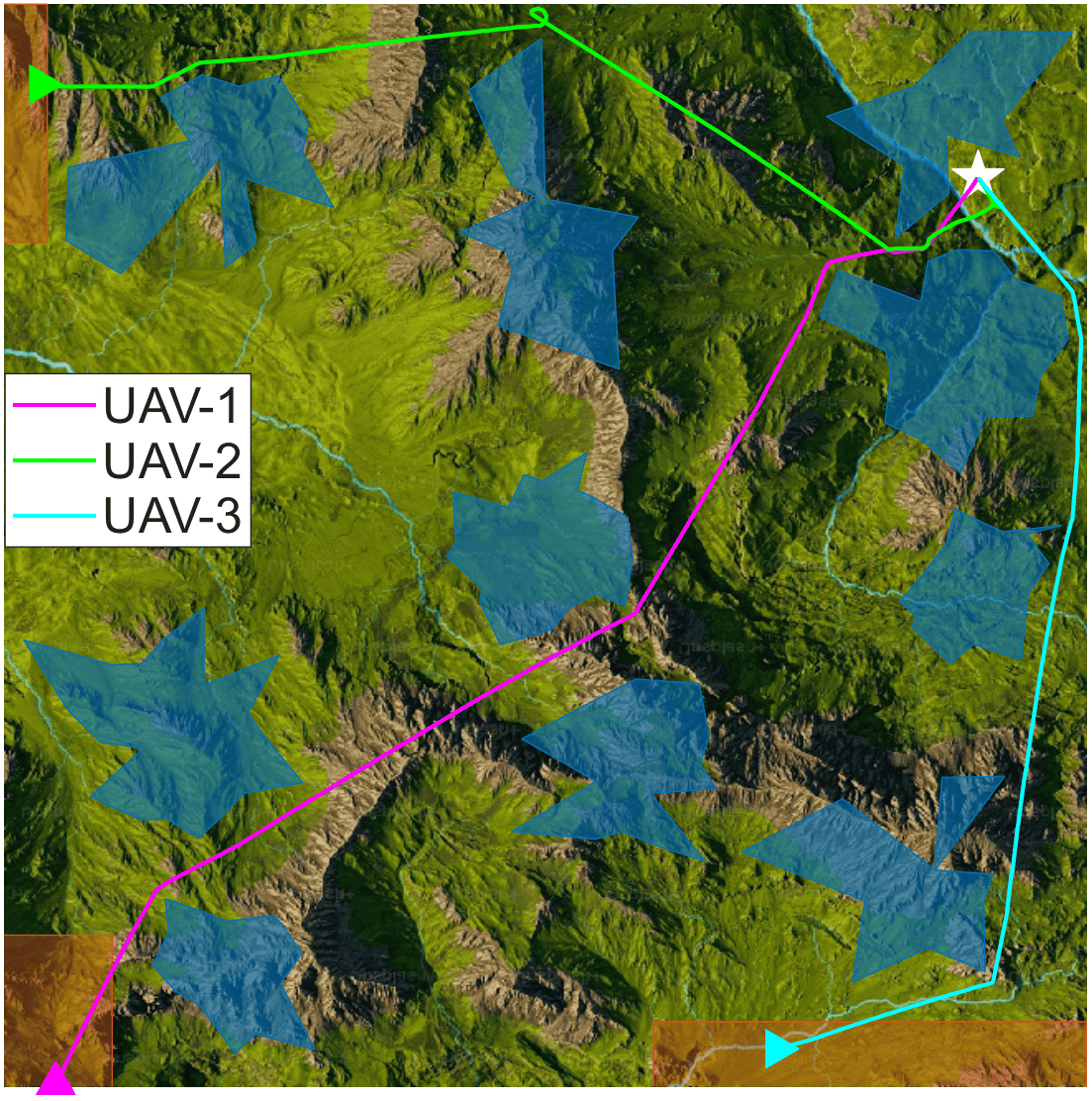}}
	\subcaptionbox{\label{fig_path3}}{\includegraphics[scale=0.3]{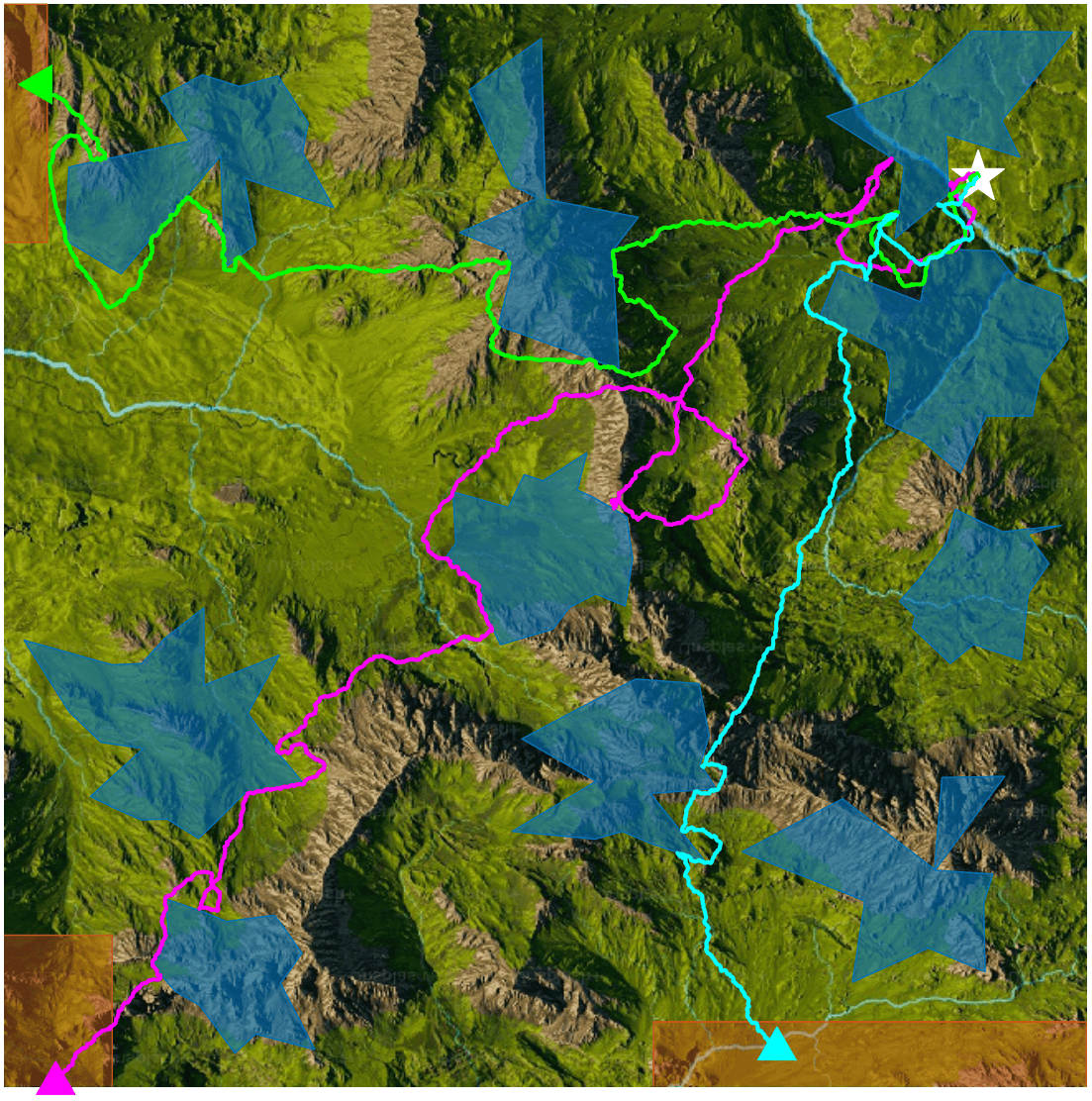}}
	\subcaptionbox{\label{fig_path4}}{\includegraphics[scale=0.3]{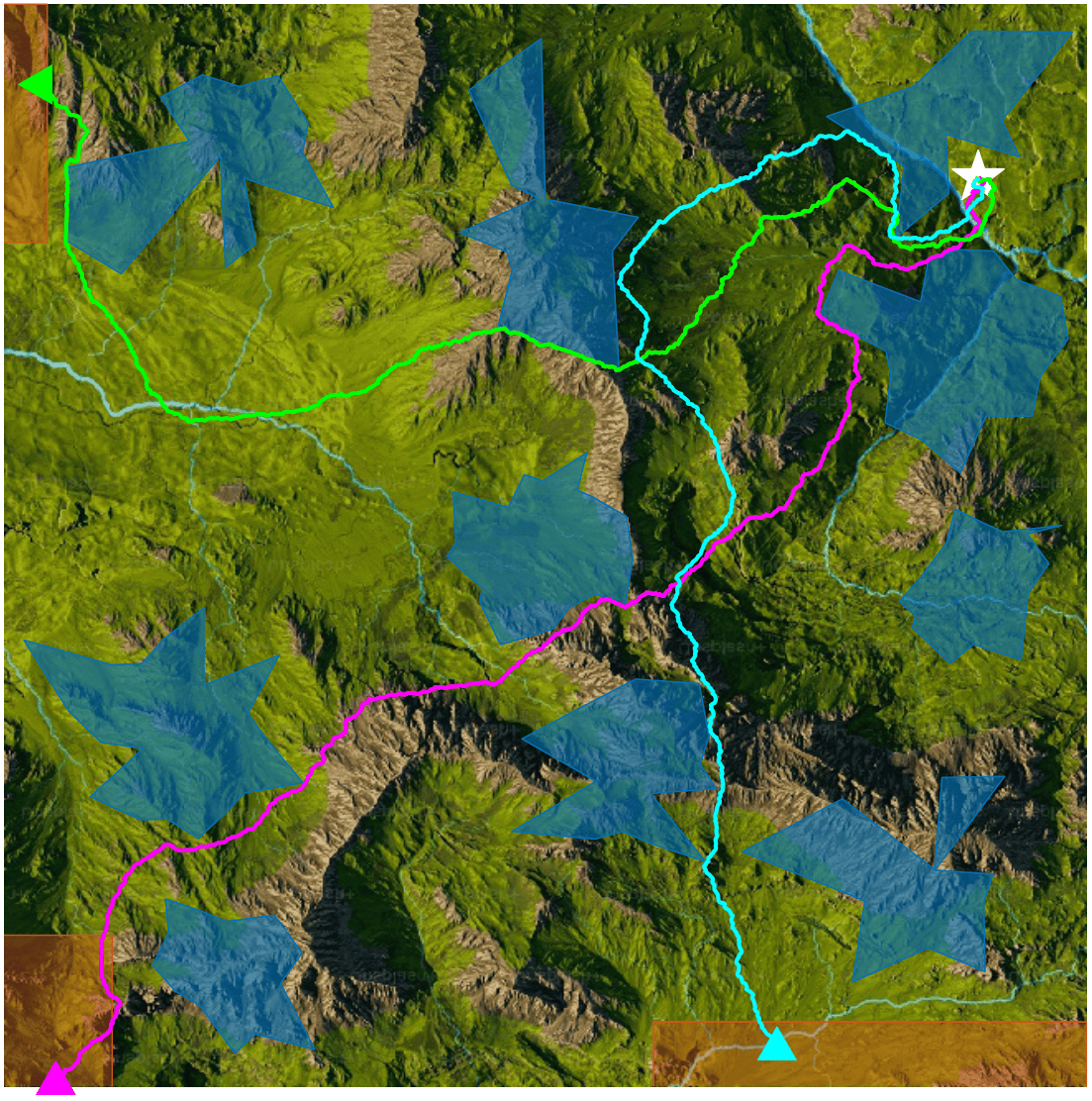}}
	\subcaptionbox{\label{fig_path2}}{\includegraphics[scale=0.3]{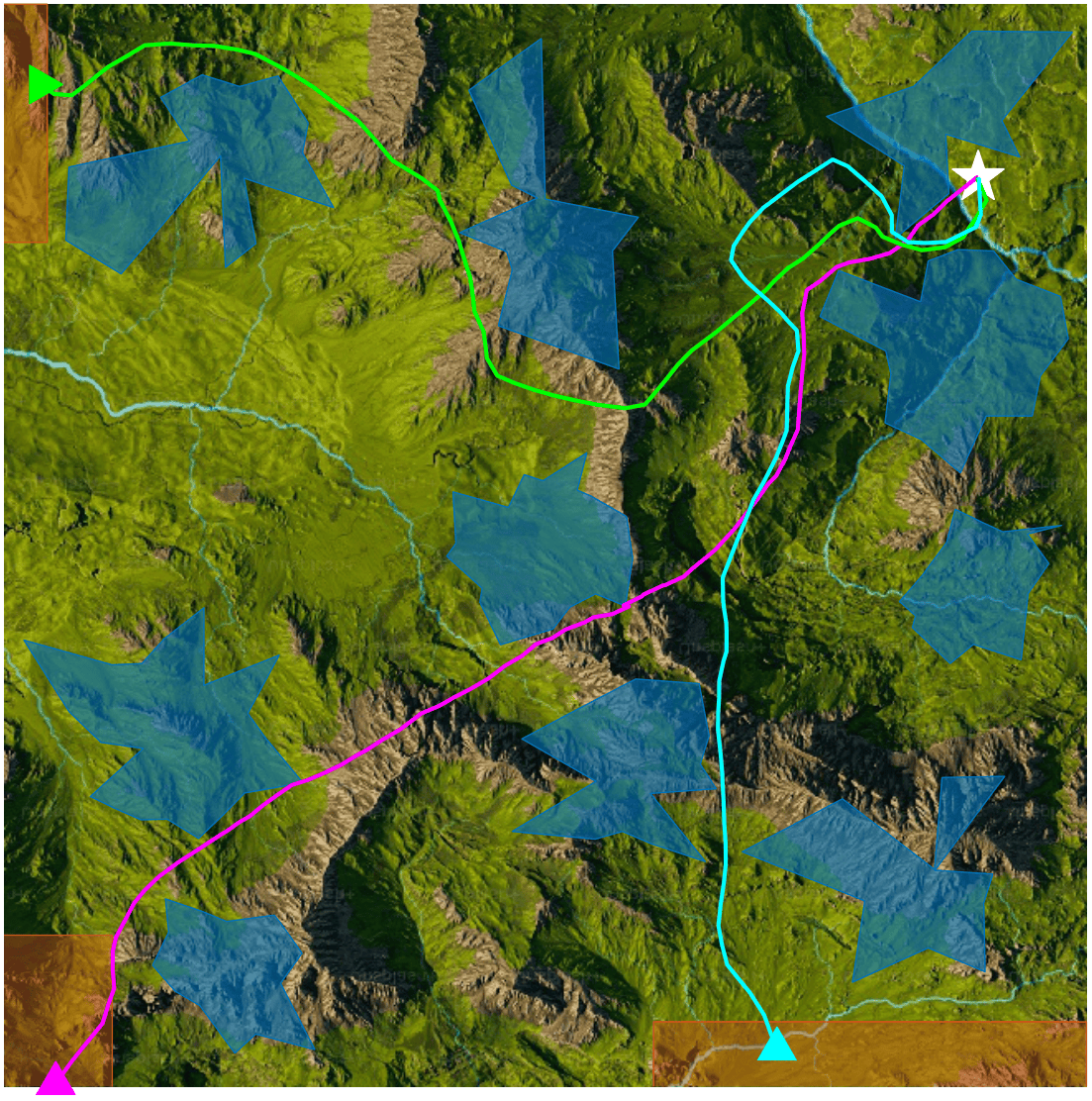}}
	\caption{  Multi-UAV path planning results: (a)  $H{{_{\scaleto{w}{4pt}}}}H$, (b) Lazy-RRT, (c) Extended-RRT, (d) RRT$^\thinstar$. The destination is symbolized by a white star while the blue polygons represent no-flight zones and orange rectangles are for the take-off zones  }
	\label{fig_path_3UAV}
\end{figure*}

The following class of tests aims at analyzing the performance of the proposed $H{{_{\scaleto{w}{4pt}}}}H$ path planning algorithm in terms of generating short and safe trajectories. To this end, a scenario of three UAVs taking-off from different locations and flying within a sub-area containing 10 irregular no-flight zones to reach a certain destination is considered.\\
Fig. \ref{fig_path_3UAV} illustrates the 2D view of the generated trajectories. It is clear that the proposed algorithm was able to produce the shortest and steadiest trajectories with less number of turns in comparison with the \textit{Lazy-RRT}. It is also observed that  \textit{RRT}$^\thinstar$ generated a trajectory for UAV 1 similar to that of the proposed approach.

Table. \ref{tab_path} portrays the average results of all the generated paths that can be tracked to reach the destination. It can be seen that the proposed algorithm clearly managed to generate the shortest and safest paths. Among the benchmarking algorithms, only the heuristics \textit{MOEA/D} and \textit{MSFLA} managed to get good results in terms of path length but at the detriment of path safety as their paths are respectively feasible by only 85.7\% and 94.5\%. Unlike the RRT variants, which succeeded to produce feasible paths, the remaining algorithms failed to do so.

\begin{table*}
	\caption{ Mean optimality and feasibility scores of planned trajectories for a UAV }
	\begin{minipage}{0.45\textwidth}
		\centering
		\tiny
		\begin{tabular}{c  c  c  c  c  c  c  c  c   c   c  c   p{2.5cm} p{2.5cm}   p{2.5cm} p{2.5cm}   p{2.5cm}   p{2.5cm}  p{2.5cm} p{2.5cm}  p{2.5cm}  p{2.5cm}  p{2.5cm}  p{2.5cm} } 
			\hline
			\multicolumn{2}{c}
			{\backslashbox{\textit{Metrics}}{\textit{Algorithms}}} &  $H{{_{\scaleto{w}{3pt}}}}H$   &   \textit{FLEA}  &  \textit{MODA}   &   \textit{MOEA/D} &  \textit{MOFPA}  &  \textit{MOGOA}   &  \textit{MOGWO}  &  \textit{MOLSA}  &  \textit{MOPSO}  &  \textit{MSFLA}    \\
			\hline
			\multicolumn{2}{c}{Path length (m)}  & 6025.08 &6292.80 &11766.02 &5905.49 &7855.96 &18139.54 &18663.87 &6977.19 &7848.83 &5969.79  \\
			\hline
			\multicolumn{2}{c}{Path safety violation (\%)} & 0 &3.53 &10.53 &14.3 &1.43 &15.54 &20.93 &1.27 &5.23 &5.5  \\
			\hline
		\end{tabular}
	\end{minipage}
	\vfill
	\begin{minipage}{0.45\textwidth}
		\centering
		\tiny
		\begin{tabular}{c  c  c  c  c  c  c  c    c   c   c  p{2.5cm}  p{2.5cm}  p{2.5cm}   p{2.5cm} p{2.5cm}   p{2.5cm} p{2.5cm}   p{2.5cm} p{2.5cm}   p{2.5cm}  p{2.5cm} } 
			\multicolumn{2}{c}{ \multirow{4}{*}{ -- Continued -- }}  &  \multirow{2}{*}{ \textit{NSGA-III}  }   &  \multirow{2}{*}{ \textit{PREA} }  &  \multirow{2}{*}{ \textit{RSEA}  }  &  \multirow{2}{*}{ \textit{RVEA} }   & \multirow{2}{*}{ \textit{SMEA} }    &  \multirow{2}{*}{ \textit{Lazy-RRT} }  & \multirow{2}{*}{ \textit{RRT-connect} }   & \multirow{2}{*}{ \textit{Extended-RRT} }    &  \multirow{2}{*}{ \textit{RRT$^\thinstar$} }   \\
			&  &  &  &   &  &   &  &  &  &  \\
			\cline{3-11}
			&  &  6069.82 & 6976.01 &6069.10 &15959.89 &15891.87 &9828.65 &9054.07 &8281.11 &6227.71 \\
			\cline{3-11}
			&  & 8.53  &  6.13 &9.17 &23.9 &10.83 &0 &0 &0 &0 \\
			\cline{3-11}
		\end{tabular}
	\end{minipage}
	\label{tab_path}
\end{table*}

\begin{figure*}
	\centering
	\subcaptionbox{\label{fig1_time1}}{\includegraphics[scale=0.33]{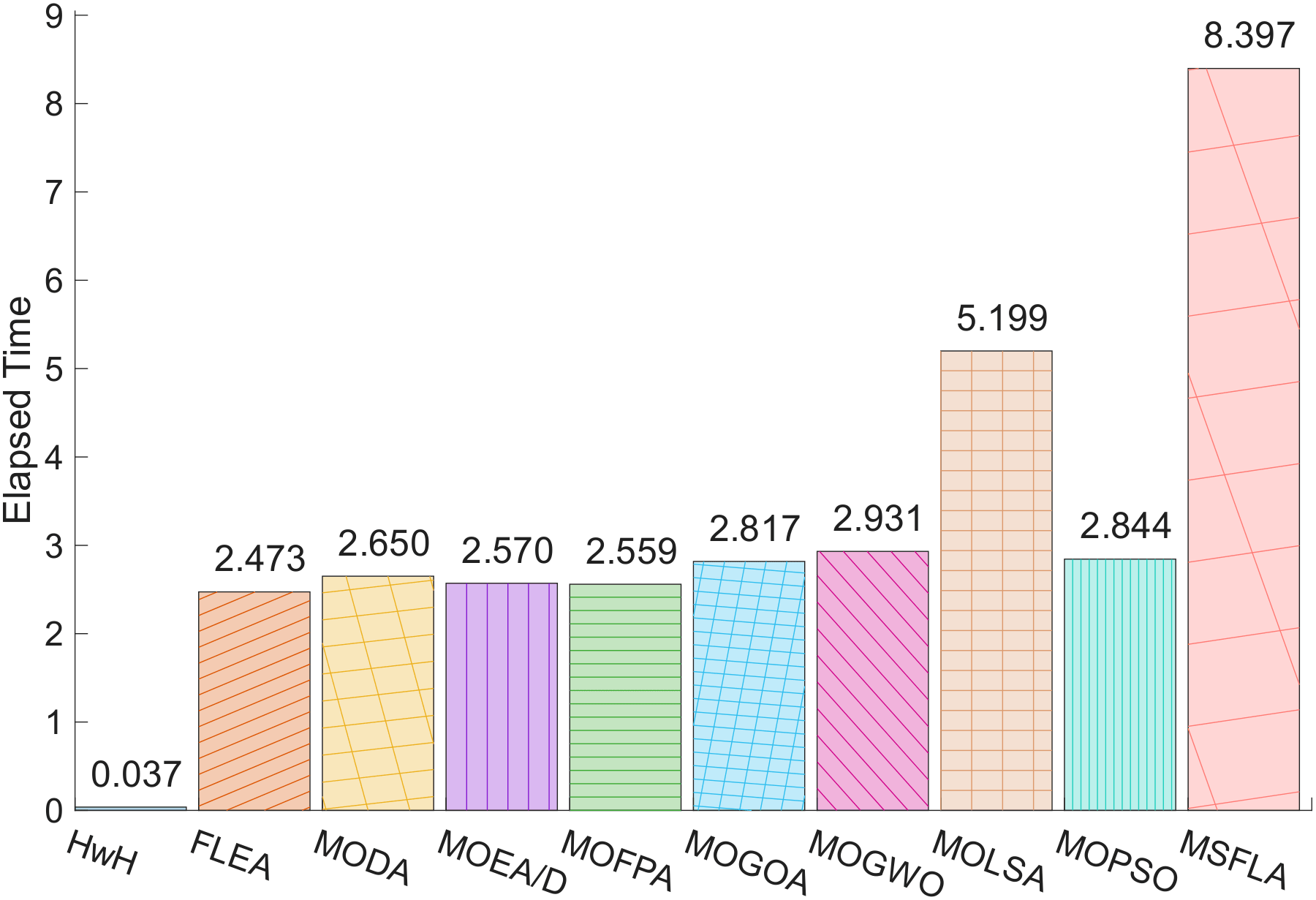}}
	\subcaptionbox{\label{fig1_time2}}{\includegraphics[scale=0.33]{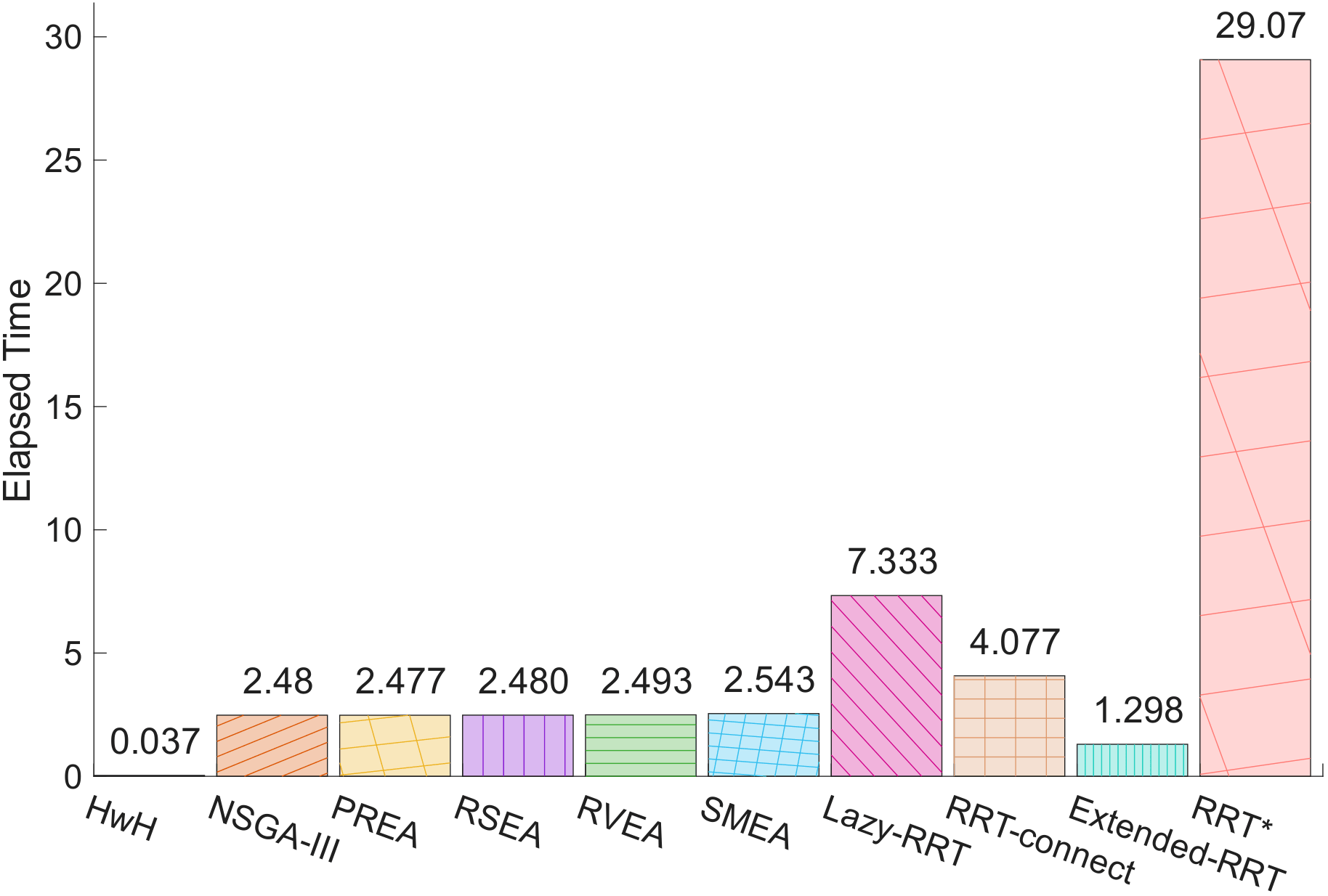}}
	\caption{   Mean path planning runtime results (in minutes)   }
	\label{fig_time}
\end{figure*}

An important metric in evaluating path planning algorithms and heuristics is their speed and amount of CPU runtime they take to create a solution. Fig. \ref{fig_time} portrays the average execution time of the proposed $H{{_{\scaleto{w}{4pt}}}}H$ versus the benchmarking path planning algorithms.
As can be clearly seen, the proposed algorithm has the upper hand in terms of speed in contrast to \textit{RRT}$^\thinstar$, which showed the slowest execution time than comes\textit{ Lazy-RRT} and \textit{MOLSA}.

\begin{figure*}
	\centering
	\subcaptionbox{\label{fig_success1}}{\includegraphics[scale=0.29]{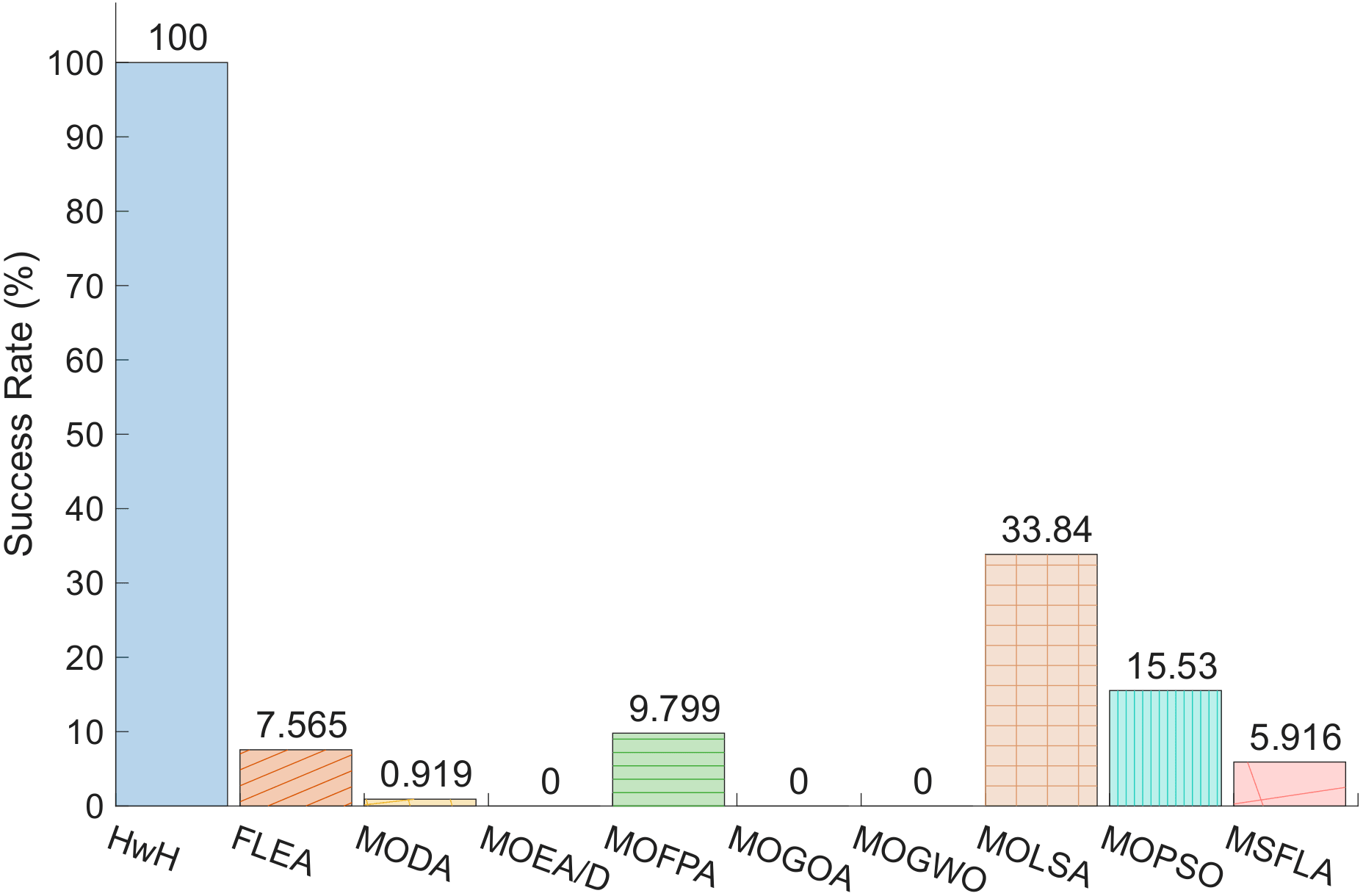}}
	\subcaptionbox{\label{fig_success2}}{\includegraphics[scale=0.29]{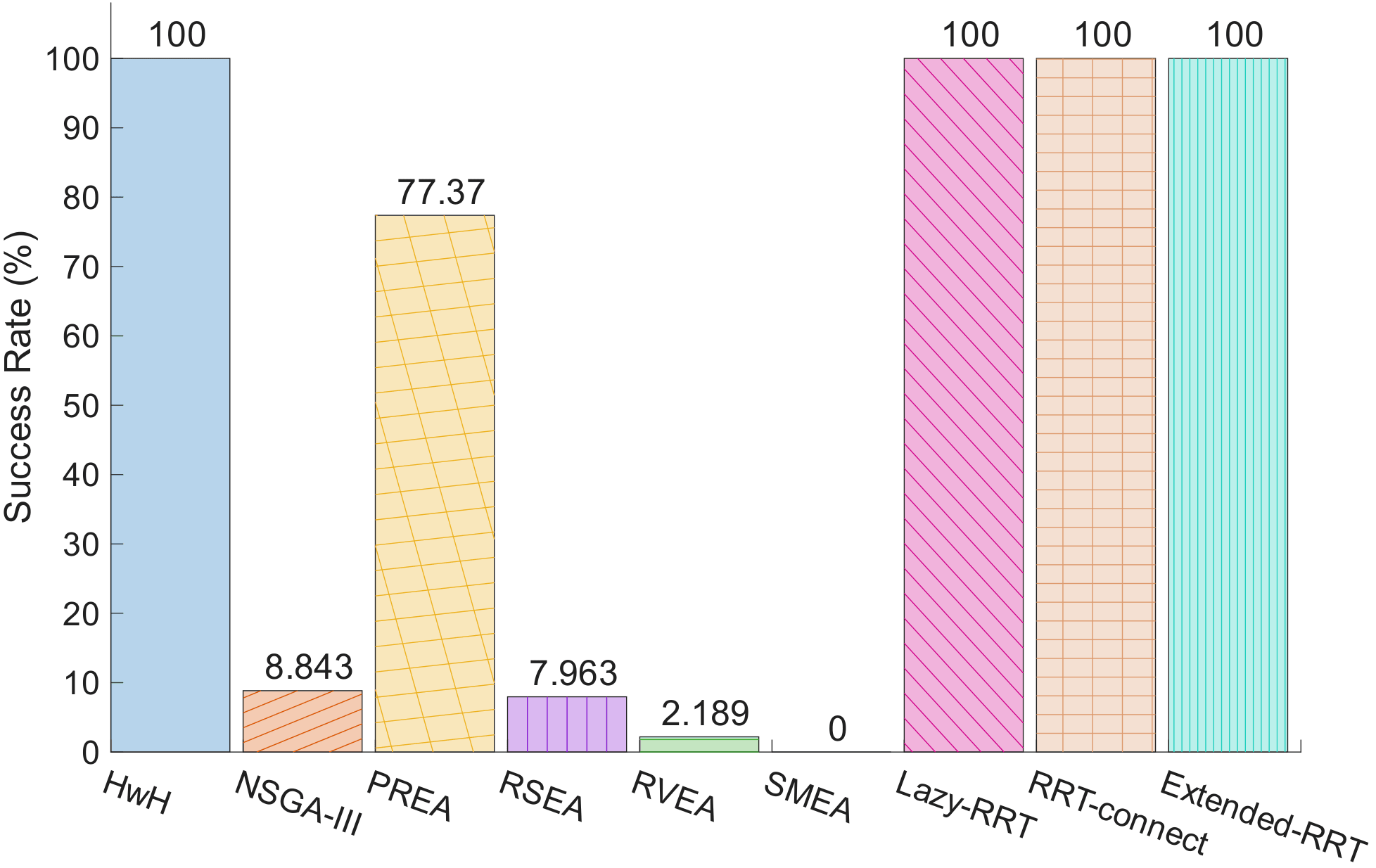}}
	\caption{  Mean success rates of generating feasible trajectories  }
	\label{fig_success}
\end{figure*}

Fig. \ref{fig_success} further demonstrates the success rates of how a given path planning algorithm can produce a set of feasible trajectories.\\

The results of Fig. \ref{fig_success} confirms that the proposed $H{{_{\scaleto{w}{4pt}}}}H$ algorithm and the RRT variants have an excellent stability in generating safe paths for different simulation trials. However, a major difference between $H{{_{\scaleto{w}{4pt}}}}H$ and RRT variants lies in the aptitude of obtaining the shortest paths in a short time and Fig. \ref{fig_time} with Table \ref{tab_path} proved the superiority of the proposed algorithm.
On the contrary, the benchmarking algorithms showed a mediocre performance as they exhibit very low success rates except for \textit{PREA}, which scored a success rate higher that 70\%.
Such results were likely due to the following issues:
\begin{itemize}
	\itemsep0em
	\item [-] vulnerability to several problems such as the curse of dimensionality when leaving the promising regions in the search space and fail to rediscover them again.
	\item [-] proneness of Pareto-based heuristics to fixate on isolated and remote non-dominated solutions in a population and by doing so they risk to overlook other equally good and well-distributed solutions.
	\item [-] there are some limitations emerging from the dependence on non-dominated sorting and crowding distance operators, which may restrict the convergence when solutions have akin fitness scores and fail to preserve the balance between convergence and diversity of solutions.
	\item [-] in addition to the conventional drawbacks of evolutionary algorithms such as: aptitude of easily falling in local optima, weak exploration, bad local searching, premature convergence or slow convergence rate.
\end{itemize}

\subsection{Controller stability analysis}

\begin{figure*}
	\centering
	\hspace*{-0.6in}
	\subcaptionbox{\label{fig_DLnT_PID1}}{\includegraphics[scale=0.32]{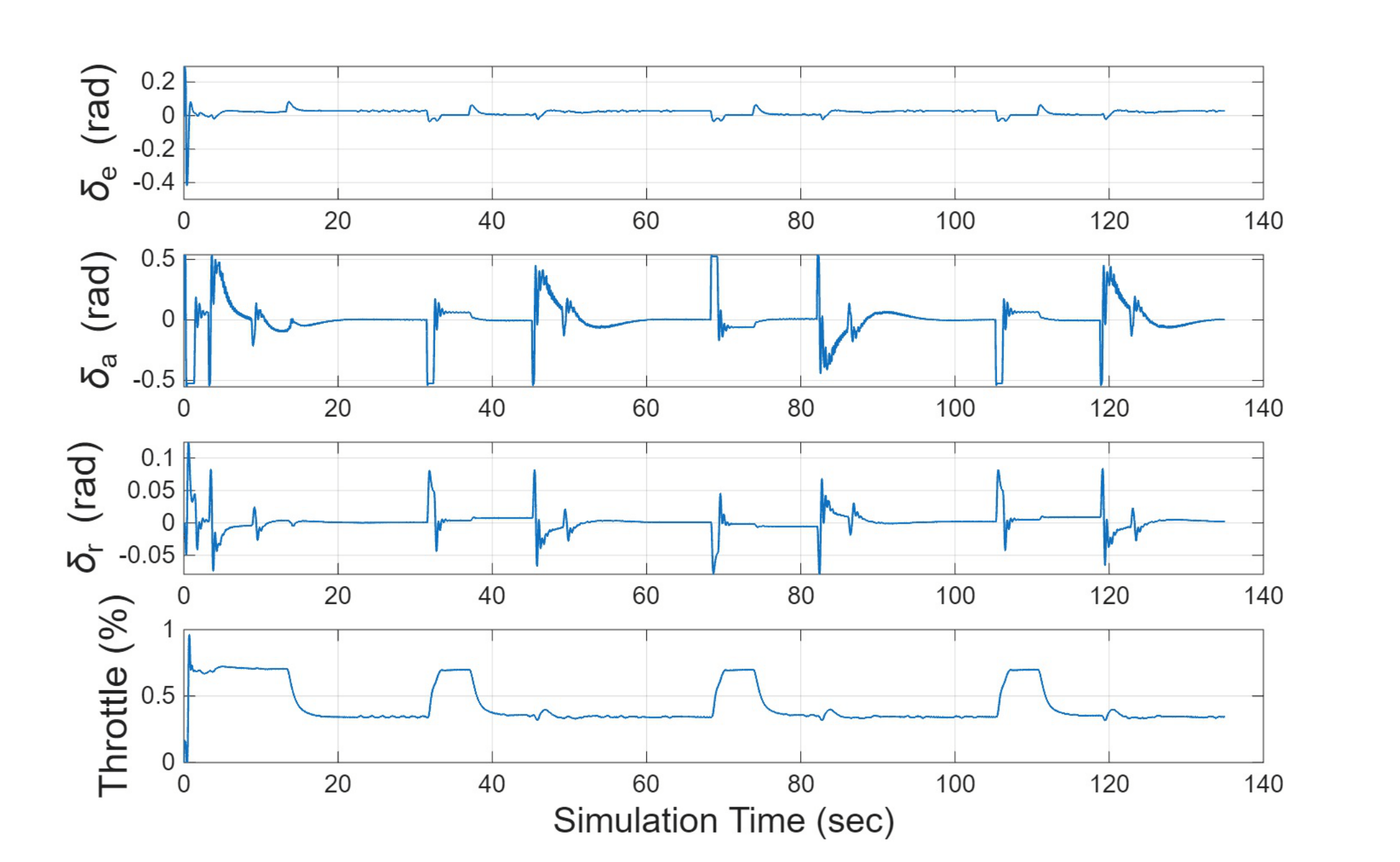}}
	\subcaptionbox{\label{fig_DLnT_PID2}}{\includegraphics[scale=0.32]{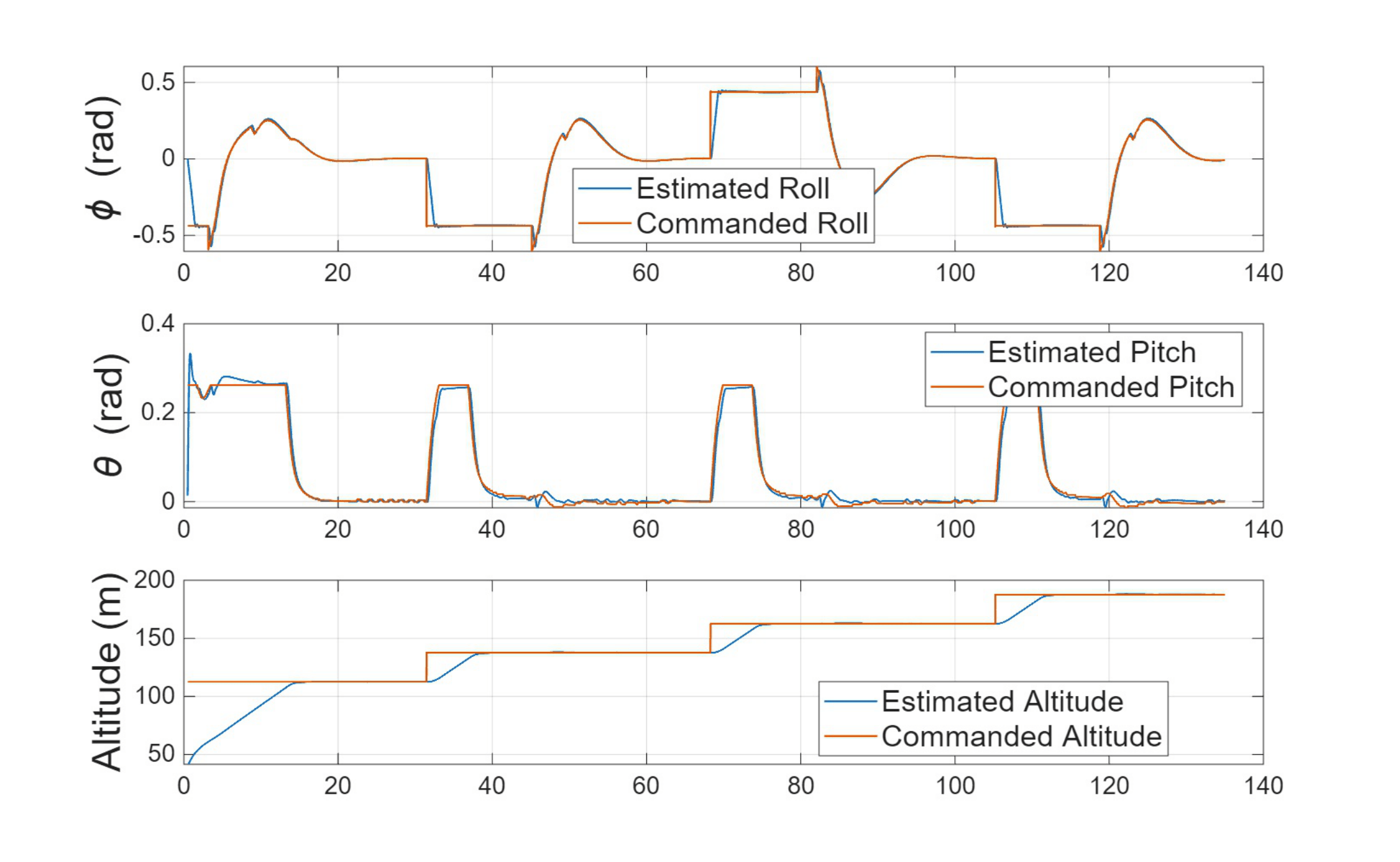}}
	\caption{ Results of using DLnT-assisted PID controller: (a) control surface deflections, (b) roll, pitch and altitude tracking performance }
	\label{fig_DLnT_PID}
\end{figure*}

The final class of tests aims at analyzing the performance of the PID controller's gains-tuning in terms of reducing the tracking errors.\\
Fig. \ref{fig_DLnT_PID1} demonstrates the efficiency of the proposed DLnT gain-tuning method on the control inputs $\delta_e, \delta_a, \delta_r$ (elevator, aileron and rudder deflections) and throttle. The viewed responses indicate less oscillatory behaviors leading to less energy consumption and reflecting its stability and adequacy to reduce the tracking errors. Likewise, Fig. \ref{fig_DLnT_PID2} portrays the tracking performance of roll, pitch and altitude between the commanded signals and measured UAV's responses. The resulted gains from the proposed algorithm manages to reduce the tracking errors and stabilizes the UAVs within the desired reference trajectory.

\begin{figure*}
	\centering
		\hspace*{-0.8in}
	\subcaptionbox{\label{fig_path_track1}}{\includegraphics[scale=0.3]{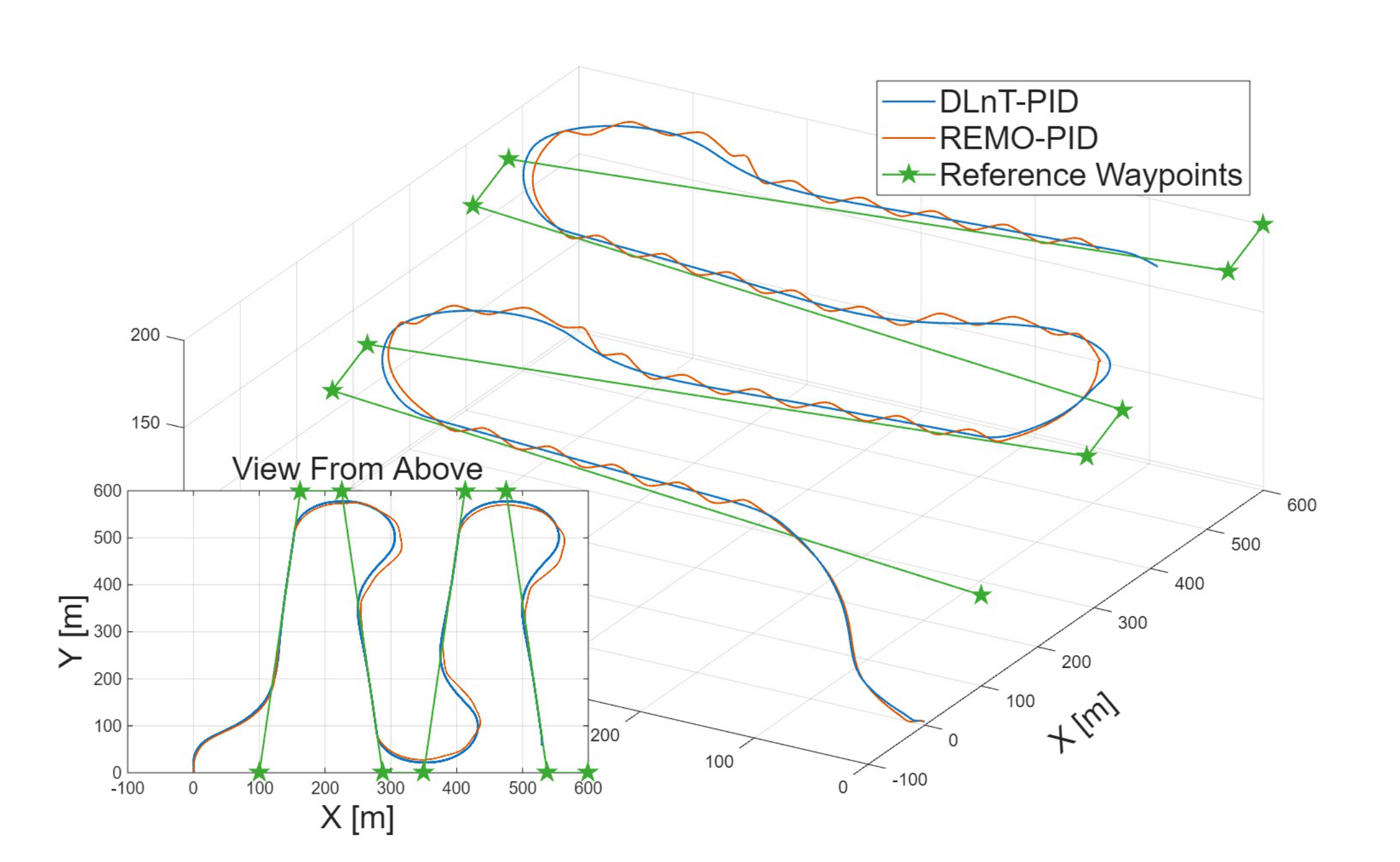}}
	\subcaptionbox{\label{fig_path_track2}}{\includegraphics[scale=0.3]{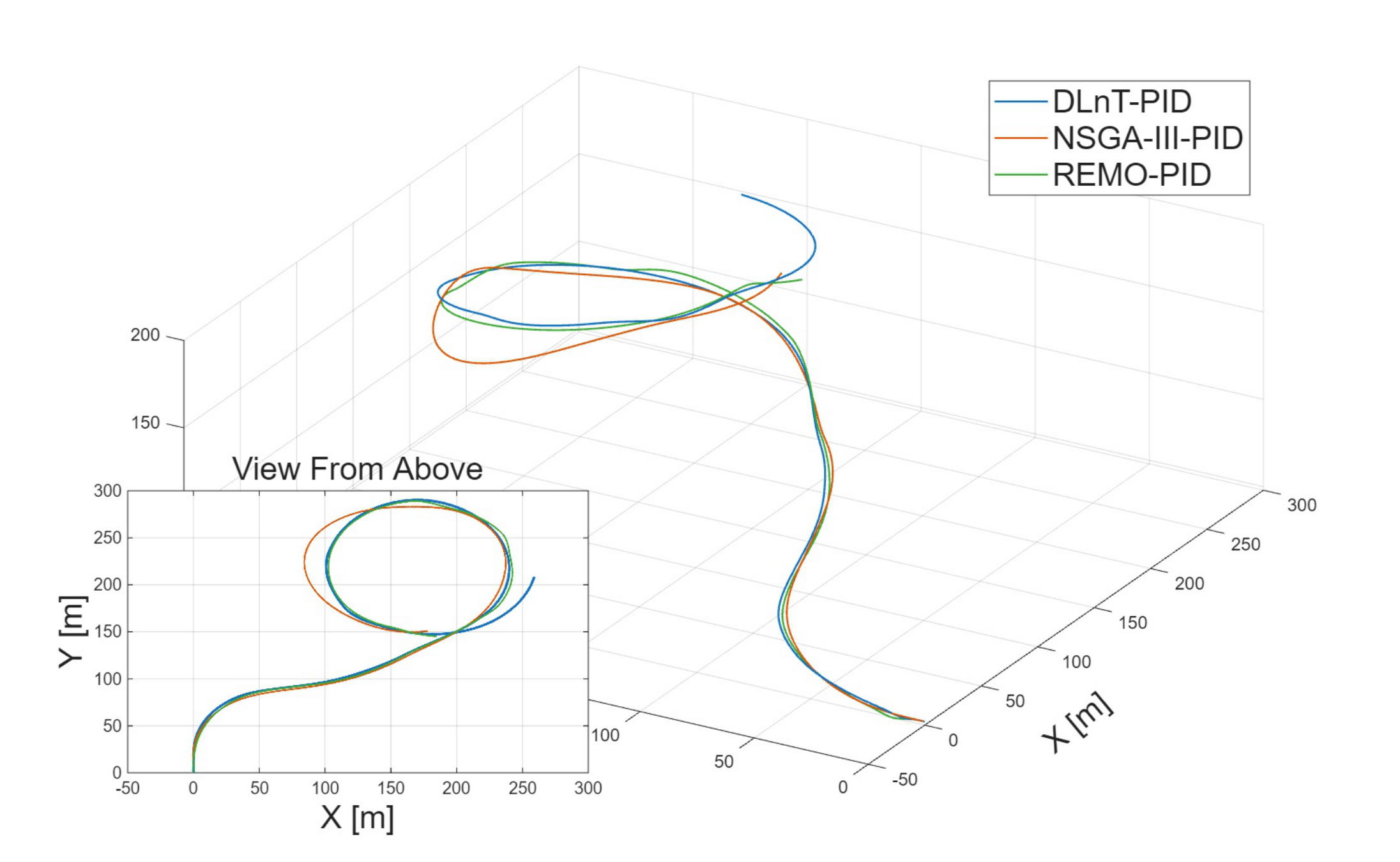}}
	\caption{  Path tracking results using DLnT and heuristic gain tuned PID during two different sub-missions  }
	\label{fig_path_track}
\end{figure*}

Finally Fig. \ref{fig_path_track} shows the 3D view of two path segments pursued by a UAV whose PID controller's gains were tuned using the DLnT algorithm. It is observed that the UAV is able to follow the reference trajectory but in accordance with its aerodynamic limits as it cannot achieve sudden brusque turns (as shown in Fig. \ref{fig_path_track1}). The \textit{REMO} gains tuned PID managed to follow the reference but the tracked trajectory is not smooth as it shows  countless oscillations. The use of \textit{NSGA-III} to adjust the PID gains (Fig. \ref{fig_path_track2}) did not yield satisfactory results as the UAV deviates considerably from the reference trajectory.

\section{Conclusion}
\label{sec6}

In this work, we addressed the problem of prolonging the endurance of soaring-capable and cooperative UAVs flock by harvesting potential energy from thermal updrafts to compensate their lost altitude due to gliding. The objectives were to stay airborne for as long as possible, keeping an eye on targets scattered over a geographical region-of-interest with less power consumption.
On top of defining the mathematical background of the problem, a self learning framework built upon modular components is proposed. It was examined via rigorous and diverse performance analyses, proving therefore its efficiency and adequacy for persistent surveillance applications. \\

The principle of the approach relies on modeling the UAVs as rational autonomous agents capable of not only interacting with themselves and with the environment but also capable of learning from such interactions. As such, the proposed solution involves several key aspects like encoding the actions and behaviors of agents as discrete switchable states with explicit roles, each can be triggered by certain criteria.\\ Also, the autonomy of agents is reinforced with a local behavioral decision-making module enabling them to choose the best available action. In the same way as the global manager works using a distributed decision-making module to coordinate and manage the behaviors of agents in favor of maximizing the mission objectives.

Moreover, a planning module is introduced to assign surveillance and exploration sub-missions for the agents and to provide them with appropriate navigation waypoints. And with the help of the $H{{_{\scaleto{w}{4pt}}}}H$ path planning algorithm employing the concepts of visible obstacles and predicted future ones, the agents can safely select the collision-free trajectories toward their destinations. Not forgetting the delayed learning and tuning strategy employed to optimize the gains of a PID controller and reduce the path tracking errors.\\

From a future work perspective, more guidance and soaring strategies will be investigated to exploit other energy and lift sources in the environment. This will in turn incorporate more challenging conditions, such as moving obstacles (e.g., birds) with non-uniform motion, obstruction and harsh meteorological phenomena.

\section*{Acknowledgments}
The research work addressed in this publication was funded by NSERC (Natural Sciences and Engineering Research Council of Canada), Mitacs and FRQ (Fonds de Recherche du Québec), two Canadian non-profit research organizations.


	%
	

\end{document}